%% file: main.tex
\documentclass[journal,onecolumn]{IEEEtran}

\usepackage{microtype}
\usepackage{graphicx}
\usepackage{subcaption}
\usepackage{titletoc}
\usepackage{booktabs} % for professional tables
\usepackage{amsmath}
\usepackage{amssymb}
\usepackage{mathtools}
\usepackage{amsthm}
\usepackage[inline]{enumitem}
\usepackage{float}
\usepackage{placeins}
\usepackage{array}
\usepackage{tcolorbox}
\usepackage{threeparttable}
\usepackage[table]{xcolor}
\usepackage{subcaption} 
\usepackage{adjustbox}
\usepackage{multirow}
\usepackage[ruled,vlined,linesnumbered,algo2e]{algorithm2e}
\tcbuselibrary{breakable} 
\usepackage[bookmarks=true,         colorlinks=true,linkcolor=green!60!black,urlcolor=blue,citecolor=green!60!black]{hyperref}
\usepackage{svg}
\usepackage[utf8]{inputenc} % allow utf-8 input
\usepackage[T1]{fontenc}    % use 8-bit T1 fonts
\usepackage{url}            % simple URL typesetting
\usepackage{amsfonts}       % blackboard math symbols
\usepackage{nicefrac}       % compact symbols for 1/2, etc.

\usepackage[nameinlink,capitalise]{cleveref}
\crefname{section}{\S}{\S\S}
\Crefname{section}{\S}{\S\S}
\crefname{lemma}{lemma}{lemmas}
\Crefname{lemma}{Lemma}{Lemmas}
\crefname{thm}{theorem}{theorems}
\Crefname{thm}{Theorem}{Theorems}

\usepackage[numbers]{natbib}

\newenvironment{inparaenum}
  {\begin{enumerate*}[label=(\roman*)]}
  {\end{enumerate*}}
\usepackage{hyperref}

\DeclareMathOperator*{\argmax}{arg\,max}
\DeclareMathOperator*{\argmin}{arg\,min}

\begin{document}

\title{ATLAS: Scaffold-Free Algorithm Synthesis by LLMs\\
via Embedding-Guided Quality-Diversity Search}
\author{Danial Yazdani, Mohammad Nabi Omidvar, Yuan Sun, Maksud Ibrahimov, and Xiaodong Li%
\thanks{Danial Yazdani and Xiaodong Li are with the School of Computing Technologies, RMIT University, Melbourne, VIC, Australia (e-mails: \mbox{danial.yazdani@gmail.com}; \mbox{xiaodong.li@rmit.edu.au}). 
Mohammad Nabi Omidvar is with the School of Computing and Leeds University Business School, University of Leeds, Leeds, UK (e-mail: \mbox{M.N.Omidvar@leeds.ac.uk}). 
Yuan Sun is with La Trobe Business School, La Trobe University, Melbourne, VIC, Australia (e-mail: \mbox{Yuan.Sun@latrobe.edu.au}). 
Maksud Ibrahimov is with Effective AI, Melbourne, VIC, Australia (e-mail: \mbox{maksud@effective-ai.com.au}).}%
\thanks{This work was supported by RMIT RACE through the provision of API access.}%
\thanks{Corresponding author: Xiaodong Li.}}
\maketitle

\begin{abstract}
Most LLM-based automated algorithm design methods optimize a designated component within a human-specified scaffold,
fixing overall organization and component interactions.
We present ATLAS, an embedding-guided quality-diversity framework for scaffold-free full-algorithm synthesis
in combinatorial optimization.
The problem specification supplies objectives and constraints; a minimal I/O interface fixes only instance
and solution formats; the LLM chooses and restructures components, interactions, and control flow.
This freedom enlarges the search space, risking invalid candidates and premature convergence to one design region.
ATLAS independently detects execution, interface, and feasibility failures, recomputes objectives,
and applies error-conditioned repair; similarity-based archive management preserves algorithms across embedding-space
regions to counter premature convergence.
Its three-layer search refines the best design, gives other regions dedicated refinement opportunities,
and performs cross-region synthesis to recombine components and their interactions.
Across four NP-hard problems, ATLAS outperforms several state-of-the-art component-synthesis methods
and a matched full-synthesis baseline while remaining competitive with strong human-designed algorithms.
One ATLAS run retains several algorithms with comparable performance from distinct embedding-space
regions rather than a single design.
Code inspection finds that these multi-component designs differ in their primary construction or global-search backbone.
Our results suggest that embedding-guided quality-diversity search can make the enlarged full-algorithm
design space practically searchable.
Source code and exact executable prompts are available at 
\href{https://github.com/Danial-Yazdani/ATLAS}{\nolinkurl{https://github.com/Danial-Yazdani/ATLAS}}.
\end{abstract}

\begin{IEEEkeywords}
Automated algorithm design, combinatorial optimization, large language models, multimodal optimization,
quality-diversity optimization.
\end{IEEEkeywords}

\section{Introduction}
\label{sec:introduction}

Optimization algorithms are fundamental tools for applications in logistics, scheduling,
and resource allocation~\cite{toth2014vehicle,papadimitriou1998combinatorial}.
Designing such algorithms typically requires extensive expert knowledge and problem-specific engineering effort.
Recent progress in automated algorithm design has shown that large language models (LLMs) can synthesize heuristic
components through iterative search~\cite{yang2024large,liu2024systematic}.
Methods such as \textsc{FunSearch} (Searching in the Function Space)~\cite{romera2024mathematical},
Evolution of Heuristics (\textsc{EoH})~\cite{liu2024evolution}, and Reflective Evolution
(\textsc{ReEvo})~\cite{ye2024reevo} evolve algorithmic functions (priority selectors, construction heuristics,
scoring functions) that operate as components within predefined scaffolds.
For these methods, the synthesized component occupies a designated role inside user-defined control flow;
the surrounding framework supplies the remaining end-to-end algorithmic machinery
and may perform solution-construction or feasibility-preserving operations.
Although this scaffolding makes synthesis and evaluation manageable, it imposes much of the algorithmic
architecture in advance and restricts the LLM from going beyond the designated interface to add, remove, reorder,
or jointly redesign components and their interactions.

Full-algorithm synthesis removes this restriction by placing complete algorithms, rather than only designated
components, under LLM-guided search.
This relaxation enables exploration of different component inventories and algorithmic organizations,
including construction heuristics, local search, metaheuristics, and hybrid pipelines.
Existing approaches instantiate it with different boundaries: \textsc{LLaMEA}~\cite{vanstein2025llamea}
generates complete metaheuristics under a box-constrained black-box interface,
whereas \textsc{AlphaEvolve}~\cite{novikov2025alphaevolve} evolves user-marked regions of supplied programs.
This greater design freedom expands the search space into a larger, more heterogeneous landscape
that is harder to navigate and exposes generated algorithms to potential failures in execution, interface compliance,
solution construction, and constraint handling.
This freedom can also increase synthesis cost because it admits more complex, multi-stage algorithms
whose execution and iterative refinement require greater computational effort.
However, the main advantage is the resulting design freedom: without a user-prescribed internal architecture, 
the LLM can explore alternative component inventories, component interactions, and control flows that a fixed scaffold 
would otherwise exclude.

The resulting algorithm landscape is larger and can also be multimodal:
competitive algorithms with different internal organizations may occupy distinct regions of the selected representation.
Niching methods in multimodal optimization seek to locate and preserve multiple optima or modes rather than collapsing
to a single solution~\cite{li2017seeking}.
This motivates full-algorithm search that preserves several competitive regions
while refining promising designs rather than committing too early to one incumbent.

In LLM-driven synthesis, archive diversity has an additional operational role.
Reference-conditioned operators can receive the complete source of algorithms from different regions
and reason about their components and interactions when constructing a new algorithm.
Coverage therefore supplies varied source material for cross-region synthesis,
not only insurance against premature convergence: the LLM can select, discard, adapt,
or combine mechanisms from different inputs into a new end-to-end organization.

To navigate this landscape, we introduce \textsc{ATLAS} (Archive-based Three-Layer Algorithm Synthesis),
an embedding-guided quality-diversity framework for full-algorithm synthesis in combinatorial optimization.
ATLAS maintains a coverage-preserving archive, or \emph{semantic repertoire},\footnote{Here and throughout,
\emph{semantic} denotes similarity in the pretrained embedding of an algorithm's name, description,
and preprocessed source code, not program semantics defined by execution behavior.
Embedding-space \emph{regions} or \emph{clusters} denote groups induced by this representation.
Mechanism-oriented family labels are used only for the author-annotated embedding benchmark or for illustrative
representatives examined separately through code inspection.} of executable algorithms organized in a pretrained
embedding space~\cite{cully2019autonomous,chatzilygeroudis2021quality,zhang2024mgte}.
Embedding distance organizes retrieval, clustering, redundancy control,
and reference selection without requiring hand-specified behavior descriptors.
The induced regions are recomputed as the archive develops rather than fixed in advance.
A three-layer strategy concentrates refinement around the current best algorithm
and gives representatives of other regions dedicated refinement opportunities.
Cross-region synthesis supplies the LLM with components and functions from multiple input algorithms.
The LLM can then reason about their interactions and reorganize them into new hybrid algorithms.

At its core, ATLAS is scaffold-free.
The problem definition supplies the objective and constraints as requirements common to any optimization algorithm,
while ATLAS fixes only a minimal problem-specific I/O interface and prescribes no internal component inventory,
function-role partition, or control flow.
The LLM therefore controls the complete algorithmic artifact and is responsible for end-to-end solution construction
and constraint handling.
ATLAS's external evaluator does not supply or complete this logic, and it does not make returned solutions feasible;
it executes candidates under the common resource protocol, verifies returned solutions,
and independently recomputes their objective values.
Because this responsibility widens the failure surface to malformed and infeasible outputs as well as execution errors,
error-conditioned \textsc{Repair} is treated as a search operator that regenerates the candidate algorithm from
classified failure evidence.

\begin{figure*}[t]
    \centering
    \includegraphics[width=\textwidth]{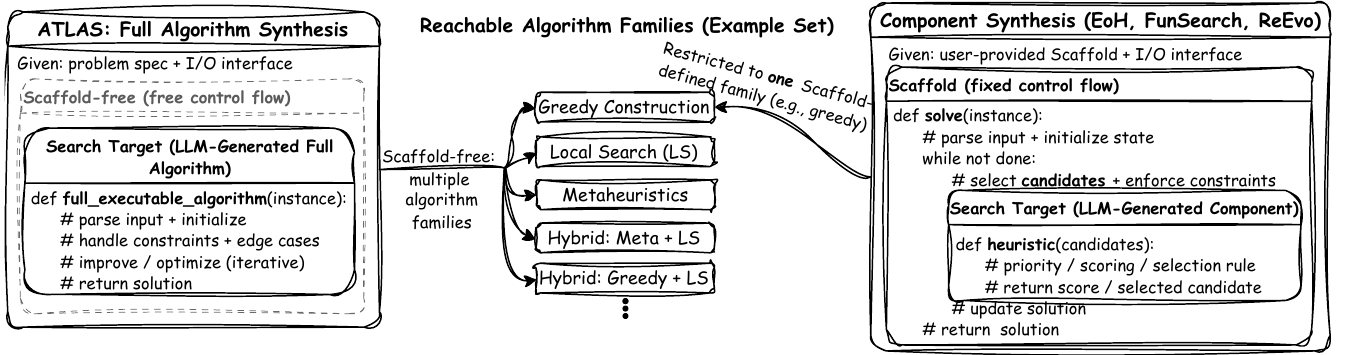}
\caption{\emph{Component-based vs. scaffold-free full-algorithm synthesis.}
ATLAS synthesizes a complete optimization algorithm that receives problem instances and returns solutions
through a fixed external I/O interface, without a user-provided internal algorithmic scaffold;
component methods synthesize a designated function within user-provided control flow.}
\label{fig:componentVSfull}
\end{figure*}

Figure~\ref{fig:componentVSfull} contrasts scaffold-constrained component synthesis
with ATLAS's full-algorithm search under its minimal external I/O interface.
ATLAS operationalizes this search through prompts that communicate the problem specification and I/O requirements,
together with LLM-driven full-algorithm operators that receive complete source code for refinement (\textsc{Improve},
\textsc{Tune}, \textsc{Simplify}), synthesis (\textsc{Combine}, \textsc{Diverge}),
and error-conditioned recovery (\textsc{Repair})~\cite{shinn2024reflexion}.
Similarity-based retrieval and redundancy-aware pruning preserve coverage across embedding-separated algorithm regions,
providing alternatives for targeted refinement and recombination.
The supporting execution system evaluates candidates under runtime and memory controls
and records code, lineage, outcomes, resource use, and failure diagnostics.

Our contributions are as follows.
\begin{itemize}[leftmargin=*,itemsep=0pt]

\item {\textbf{Scaffold-Free Full-Algorithm Synthesis:} We introduce ATLAS,
  a framework for scaffold-free full-algorithm synthesis in combinatorial optimization.
We provide the problem definition, including its objective and constraints,
  and specify the callable entry point and instance/solution I/O formats.
Beyond these necessary problem and interface requirements, the LLM is free to design the complete algorithm,
  choose its components, coordinate their interactions, and organize its end-to-end workflow without any user-prescribed
  restriction on its internal structure.}

\item \textbf{Embedding-Guided Quality-Diversity and Multimodal Search:} We propose a coverage-preserving semantic
  repertoire organized in pretrained embedding space.
Embedding-based retrieval, clustering, and redundancy control mitigate premature convergence
  and support multimodal search in the selected representation, retaining multiple near-best algorithms
  while search continues to improve the strongest designs.
The preserved alternatives also provide semantically varied references for cross-region synthesis,
  so coverage contributes to search progress as well as final repertoire quality.

\item \textbf{Hierarchical Three-Layer Search:} We develop a three-layer resource-allocation strategy over the clustered
  repertoire: Layer~1 targets the current best region, Layer~2 gives non-elite region representatives dedicated
  refinement opportunities, and Layer~3 recombines components and their interactions across regions to discover new
  organizations and hybrids.

\item \textbf{Independent Evaluation and Failure-Conditioned Recovery:}
  Because generated algorithms are responsible for solution construction and constraint handling,
  evaluation must detect a wider failure surface that includes execution, interface, malformed-output,
  and feasibility failures.
ATLAS independently verifies returned solutions and recomputes objectives,
  applies an all-or-nothing validity rule that records failures rather than converting them into finite penalties,
  and routes classified evidence to error-conditioned \textsc{Repair}.
The accompanying implementation provides problem- and interface-aware prompts,
  worker-process parallel evaluation with runtime and memory controls, and detailed lineage, resource,
  and failure records.

\item {\textbf{Empirical Design Insights for LLM-Based Full Synthesis:} Controlled studies examine
  artifact scope, representation, initialization, grouping, and search allocation.
They show that moving from component synthesis to full-algorithm synthesis through the ATLAS framework improves
  performance in most tested settings and provide practical guidance for the design of LLM-based full-algorithm
  synthesis systems.}
 
\end{itemize}

We validate these contributions across four NP-hard combinatorial optimization benchmarks.
ATLAS outperforms several state-of-the-art component-synthesis methods and the matched full-synthesis baseline,
remains competitive with strong human-designed, domain-specific algorithms,
and retains multiple competitive algorithms in separate archive regions within a run,
consistent with the intended multimodal search behavior.

The remainder of this paper is organized as follows.
Section~\ref{subsec:related_work} reviews LLM-based component and full-algorithm synthesis together
with relevant quality-diversity research; Section~\ref{sec:ATLASdesc} presents the ATLAS formulation
and search framework; Section~\ref{sec:experiments} describes the experimental protocol and results;
and Section~\ref{sec:conclusion} concludes the paper.
The supplementary material provides extended related work, implementation and evaluation details,
problem definitions and instance generation, pseudocode and evaluation protocols,
configurations and baseline specifications, complete protocols and analyses, additional experiments,
and descriptions of representative synthesized algorithms.

\section{Related Work}
\label{subsec:related_work}

This section reviews work related to the LLM-based automated algorithm design aspects of ATLAS.
We focus on component and full-algorithm synthesis, together with optimization methods
that act as meta-optimizers over the generated-algorithm space.
The concise discussion here highlights the distinctions needed for the present work;
a comprehensive taxonomic and method-by-method treatment is provided in Appendices~\ref{app:surveys}
and~\ref{app:sota_comparison}.

\paragraph{LLM-Driven Component Synthesis}
Recent methods use LLMs to synthesize \emph{algorithmic components} that occupy predefined roles within user-specified
algorithmic control flow.
Representative examples include \textsc{FunSearch}~\cite{romera2024mathematical}, \textsc{EoH}~\cite{liu2024evolution},
\textsc{ReEvo}~\cite{ye2024reevo}, HSEvo~\cite{dat2025hsevo}, MCTS-AHD~\cite{zheng2025monte}, CALM~\cite{huang2025calm},
\textsc{HeurAgenix}~\cite{yang2025heuragenix}, \textsc{MEoH}~\cite{yao2025multi}, and \textsc{EoH-S}~\cite{liu2025eoh}.
These methods differ in search, reflection, and diversity mechanisms, but their generated functions,
including priority rules, construction heuristics, and operator-selection policies,
do not replace the surrounding algorithmic scaffold. A callable interface alone does not imply component synthesis;
the distinction is whether the generated artifact owns the end-to-end algorithmic logic or fills a role within
predefined control flow.

\paragraph{Full Algorithm Synthesis}
\textsc{LLaMEA}~\cite{vanstein2025llamea} generates complete Python metaheuristics for box-constrained continuous
black-box optimization and refines one complete algorithm at a time using execution feedback.
\textsc{AlphaEvolve}~\cite{novikov2025alphaevolve} performs patch-based evolution over user-provided codebases
with annotated editable regions.
\textsc{A2DEPT}~\cite{chen2026a2dept}, developed concurrently with ATLAS,
evolves complete combinatorial-optimization algorithms using a program-lineage tree,
an explicit function-level representation, hierarchical micro/macro operators, and dependency-aware repair.
These methods all move beyond single-component synthesis but impose different domains, representations,
and structural priors.
ATLAS combines a problem specification and minimal combinatorial I/O interface
with scaffold-free full-algorithm operators that receive complete source code:
it prescribes neither a component inventory, function-role partition, nor user-provided algorithmic control flow,
and it organizes search to preserve coverage across embedding-space regions.

\paragraph{Meta-Optimization and Quality-Diversity Search}
The optimization method that searches over generated algorithms acts as a meta-optimizer
and must balance exploitation of strong designs against exploration and preservation of useful alternatives.
Prior LLM-based methods retain alternatives through mechanisms such as island models~\cite{romera2024mathematical},
tree search guided by Upper Confidence Bounds applied to Trees (UCT), which balances revisiting strong nodes
with exploring under-sampled branches~\cite{zheng2025monte}, program-lineage trees~\cite{chen2026a2dept},
and explicit objective- or instance-space diversity criteria~\cite{dat2025hsevo,yao2025multi,liu2025eoh}.
Quality-diversity methods preserve collections of high-performing candidates across a chosen descriptor
space~\cite{cully2019autonomous,chatzilygeroudis2021quality}.
ATLAS applies this principle through an embedding-based representation: distance organizes retrieval, clustering,
redundancy-aware pruning, and the selection of inputs for cross-region synthesis.
The resulting repertoire reduces premature convergence and supplies distinct regions for continued refinement.
It also allows the LLM to draw on components and their interactions from algorithms in different regions
when constructing hybrids.

\section{ATLAS}
\label{sec:ATLASdesc}
This section presents ATLAS and explains how it navigates the enlarged search space created by scaffold-free
full-algorithm synthesis.
We first clarify the responsibilities of the generated algorithm and external evaluator,
then describe the embedding-guided archive, semantic operators, initialization,
and three-layer search that jointly preserve and refine multiple algorithmic regions.

\subsection{Overview}
\label{subsec:overview}
\subsubsection{ATLAS Setting}

ATLAS synthesizes complete, executable optimization algorithms for a specified combinatorial optimization problem.
Each generated algorithm receives an instance through a fixed input interface and must return a structured solution.
Within these external boundaries, it is responsible for solution construction, constraint handling,
internal state management, and its own stopping logic within the resource limits.
ATLAS supplies no user-written algorithm skeleton, prescribed component inventory, function-role partition,
or internal control flow.
Consequently, candidate designs may vary their component organization, interactions,
and control flow instead of varying only a designated component within a fixed skeleton.

The synthesis setting fixes three external elements, none of which prescribes internal algorithm structure:
\begin{itemize}[leftmargin=*, itemsep=2pt, topsep=2pt]
    \item The \emph{problem specification} states the instance semantics, objective, constraints,
      and valid-solution requirements.
    These are properties of the optimization problem and apply to every algorithm regardless of how it was designed.
    \item A minimal algorithm-facing \emph{I/O interface} specifies the callable entry point
      and how problem instances and returned solutions are represented.
    \item The \emph{experimental protocol} specifies the execution environment
      and resource limits used uniformly within each comparison.
\end{itemize}

After executing a generated algorithm under this protocol, the external evaluator verifies the feasibility of the
returned solution and independently recomputes its objective; it does not complete, repair,
or make the solution feasible on behalf of the generated algorithm.
Scaffold freedom therefore concerns the absence of a prescribed algorithm decomposition,
not the absence of a problem specification, I/O interface, or common evaluation protocol.
We refer to this setting as \emph{scaffold-free full-algorithm synthesis under a fixed I/O interface}.
It differs from component synthesis~\cite{romera2024mathematical,liu2024evolution,ye2024reevo},
which optimizes functions inside fixed surrounding control flow, and from scaffolded full-algorithm evolution as
instantiated by \textsc{AlphaEvolve}~\cite{novikov2025alphaevolve}, where evolution targets user-marked blocks within an
externally supplied program.

An explicit function-role decomposition is not required to represent or revise multi-component designs in ATLAS.
Each reference-based operator receives the complete source of every selected algorithm,
including helper definitions and the calls linking them.
ATLAS evaluates and archives complete algorithms, while the operator suite spans different edit scopes:
\textsc{Tune} preserves structure, \textsc{Improve} preserves the core approach,
and \textsc{Combine} and \textsc{Diverge} can coordinate changes across functions or restructure the end-to-end
workflow.

Accordingly, \emph{full} refers to the scope of algorithm-design responsibility rather than to source-code length or to
treating an algorithm as a monolithic component.
Code inspection of four released illustrative algorithms shows coordinated multi-function pipelines.
In each inspected algorithm, construction or seeding feeds problem-specific improvement and diversification
mechanisms,
together with an explicit policy for accepting non-improving candidates.
These examples confirm that multi-component designs can arise in the displayed runs without a prescribed function-role
decomposition; they do not quantify how frequently such structures occur across the complete archives.

This evidence does not imply that explicit structural guidance can never improve reliability or search efficiency.
ATLAS can also use operator instructions that focus on a selected component or interaction without imposing a global
component decomposition; such policies are not compared here.

\subsubsection{Problem Formulation}
Given a problem $\mathcal{P}$ and training set $\mathcal{I}_{\text{train}} = \{x_1, \ldots, x_m\} \subset \mathcal{I}$,
the objective is to find an algorithm minimizing empirical cost:
\begin{equation}
\label{eq:objective}
    a_{\text{opt}} = \argmin_{a \in \mathcal{V}} \frac{1}{m} \sum_{i=1}^{m} \text{Cost}(a(x_i)),
\end{equation}
where $\mathcal{V}$ is the set of valid executable algorithms and $a(x_i) \in \mathcal{S}$ is the solution produced for
instance $x_i$.
Generalization is evaluated on separate test instances.

\subsubsection{ATLAS Approach}
ATLAS maintains an archive $\mathcal{A} \subseteq \mathcal{V}$ of valid algorithms
and iteratively refines it to approximate $a_{\text{opt}}$.
At each iteration, the current best algorithm is $a^* = \argmin_{a \in \mathcal{A}} J(a)$,
where $J(a) = \frac{1}{m} \sum_{i=1}^{m} \text{Cost}(a(x_i))$ is the average training cost.
ATLAS is designed to address two challenges:
\begin{inparaenum}
    \item the risk that fitness pressure concentrates search around a small number of early designs, and
    \item inefficient resource allocation, which under-refines promising but currently weaker regions.
\end{inparaenum}
To address these challenges, ATLAS organizes the archive in embedding space,
clusters it into operational semantic regions, and allocates search effort
through a three-layer strategy for local refinement, representative-level maturation, and cross-cluster synthesis.

\begin{figure*}[t]
    \centering
    \includegraphics[width=0.95\textwidth]{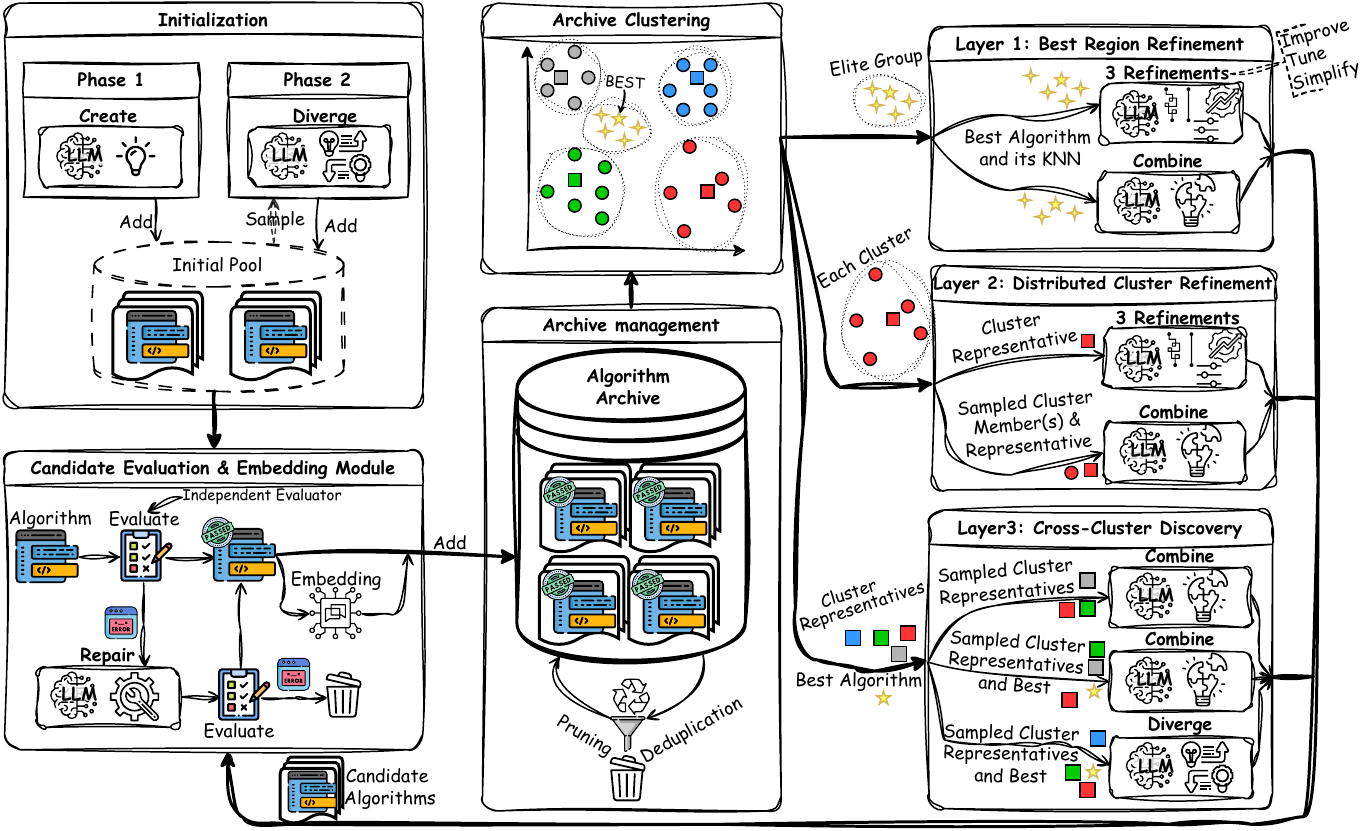}
    \caption{\emph{Overview of ATLAS.}
    \textbf{(Top-Left)} \textit{Initialization} generates diverse candidates via Create and \textsc{Diverge} operators. 
    \textbf{(Bottom-Left)} \textit{Candidate Processing} evaluates, repairs, and embeds generated algorithms. 
    \textbf{(Bottom-Center)} \textit{Archive Management} performs deduplication and capacity-constrained pruning. 
    \textbf{(Top-Center)} \textit{Archive Clustering} organizes algorithms by similarity under the selected embedding
    into the elite region $\mathcal{G}^*$ (gold stars) and semantic clusters (colored circles);
    cluster representatives are shown as squares.
    \textbf{(Right)} \textit{Three-Layer Search} allocates resources hierarchically: 
    Layer~1 refines the best algorithm ($a^*$) and its neighborhood; 
    Layer~2 refines cluster representatives in isolation; 
    Layer~3 performs cross-cluster discovery. 
    Candidates cycle through processing and archive integration
    to enable systematic refinement across multiple represented regions.
    }
 \label{fig:atlas_overview}
\end{figure*}

As illustrated in Figure~\ref{fig:atlas_overview}, ATLAS iterates over the following cycle:
evaluate and insert valid algorithms into $\mathcal{A}$, organize and cluster the archive in embedding space,
and generate new candidates through three-layer search.
{
Algorithm~\ref{alg:atlas_highlevel} summarizes this orchestration.
Appendix~\ref{app:ATLASalgorithms} provides the connected module-level procedures,
detailed in Algorithms~\ref{alg:atlas_main}--\ref{alg:atlas_gen}.
}

\begin{algorithm2e}[!t]

\fontsize{7.5}{8.4}\selectfont
\DontPrintSemicolon
\SetKwComment{tcp}{// }{}
\SetKwFunction{Create}{Create}
\SetKwFunction{Improve}{Improve}
\SetKwFunction{Tune}{Tune}
\SetKwFunction{Simplify}{Simplify}
\SetKwFunction{Combine}{Combine}
\SetKwFunction{Diverge}{Diverge}
\SetKwFunction{Repair}{Repair}
\SetKwFunction{ProcessCandidate}{ProcessCandidate}
\SetKwProg{Proc}{Procedure}{:}{}

\caption{High-Level ATLAS Pseudocode}
\label{alg:atlas_highlevel}
\KwIn{Problem specification and I/O interface; LLM; synthesis budget}
\KwOut{Training-best algorithm $a^*$, final archive $\mathcal{A}$, and cluster representatives $\mathcal{T}$}

\Proc{\ProcessCandidate{$c$}}{
    Execute $c$ on every training instance under the resource limits\;
    Verify its output and feasibility, and independently recompute its objectives\;
    \If{$c$ fails execution, interface, resource, or feasibility checks}{
        Regenerate $c$ with \Repair using the aggregated failure evidence\;
        Fully re-evaluate the repaired algorithm on all training instances\;
    }
    \eIf{$c$ succeeds on every training instance}{
        Mark $c$ valid and retain its training objective and lineage\;
    }{
        Discard $c$\;
    }
}

$\mathcal{A}\leftarrow\emptyset$\;

\tcp{Two-phase initialization}
Use \Create to synthesize algorithms; call \ProcessCandidate{$c$}\;
Iteratively apply \Diverge to references sampled from the growing valid archive;
call \ProcessCandidate{$c$}\;
Embed the retained algorithms and remove embedding-near duplicates\;

\While{the synthesis budget is not exhausted}{
    \tcp{Reorganize the evolving archive}
    $a^*\leftarrow$ the training-best member of $\mathcal{A}$\;
    $\mathcal{G}^*\leftarrow a^*$ and its embedding-nearest archive members\;
    Cluster the full archive in embedding space\;
    Exclude the cluster containing $a^*$ from the non-elite region set\;
    Remove other $\mathcal{G}^*$ members, discard empty regions, and select
    each region's training-best surviving representative\;
    Freeze $\mathcal{G}^*$, the non-elite regions, and their representatives for this iteration\;

    \tcp{Layer 1: intensify the current best region}
    Schedule repeated \Improve, \Tune, and \Simplify proposals from $a^*$\;
    Schedule probabilistically selected refinements of the other members of $\mathcal{G}^*$\;
    Schedule local hybrids by applying \Combine to $a^*$ and its neighbors\;

    \tcp{Layer 2: mature each represented non-elite region}
    \ForEach{non-elite region and its representative}{
        Schedule probabilistically selected \Improve, \Tune, and \Simplify proposals from the representative\;
        Schedule a within-region \Combine using the representative and other members of that region\;
    }

    \tcp{Layer 3: search across regions}
    Schedule \Combine between representatives from different regions to form hybrids\;
    Schedule quality-anchored hybrids by applying \Combine to $a^*$ and
    non-elite representatives\;
    Schedule \Diverge from $a^*$ and non-elite representatives to seek
    approaches separated from the referenced regions\;

    $\mathcal{V}\leftarrow\emptyset$\;
    \ForEach{scheduled operator call while budget remains}{
        Ask the LLM operator to synthesize $c$ from the complete sources of its references\;
        Call \ProcessCandidate{$c$} and add it to $\mathcal{V}$ only if valid\;
    }
    Embed all algorithms in $\mathcal{V}$ and set $\mathcal{A}\leftarrow\mathcal{A}\cup\mathcal{V}$\;
    Remove embedding-near redundancy\;
    If capacity is exceeded, prune the most similar pairs first while protecting
    the updated best neighborhood\;
}
After budget exhaustion, set $a^*$ to the training-best member of the finalized archive\;
Cluster the finalized archive and set $\mathcal{T}$ to each cluster's training-best member\;
\KwRet{$a^*,\mathcal{A},\mathcal{T}$}\;
\end{algorithm2e}

The framework comprises five components:
embedding-based similarity (\S\ref{subsec:embeddings}) for reference selection and structure discovery,
a coverage-preserving archive (\S\ref{subsec:archive}) to retain diverse regions,
semantic operators (\S\ref{subsec:operators}) to generate and repair candidate algorithms,
initialization (\S\ref{subsec:initialization}) to establish broad initial coverage, and
three-layer search (\S\ref{subsec:3layerStrategy}) to allocate search effort across regions.

\subsection{Embedding-Based Similarity}
\label{subsec:embeddings}
ATLAS structures the algorithm search space using pretrained embeddings,
providing a semantic coordinate system for reference algorithm selection, region discovery, and archive management.

\subsubsection{Algorithm Representation}
Each algorithm is represented as a tuple $a = (n, d, c)$ comprising:
\begin{inparaenum}
    \item \emph{Name} $n$,
    \item \emph{Description} $d$, and
    \item \emph{Source Code} $c$ (a standalone executable program).
\end{inparaenum}
These components are generated jointly by the LLM during synthesis (\S\ref{subsec:operators}).

\subsubsection{Embedding Strategy}
We represent algorithms using mGTE-large-en-v1.5~\cite{zhang2024mgte}, concatenating the name, description,
and preprocessed source code into a single text input.
This configuration was selected through systematic evaluation of multiple embedding strategies
(Appendix~\ref{app:embedding_ablation}).

\subsubsection{Similarity Infrastructure}
Let $\mathbf{e}_i$ denote the embedding of algorithm $a_i$.
We maintain an incremental distance matrix $\mathbf{D} \in \mathbb{R}^{|\mathcal{A}| \times |\mathcal{A}|}$
with entries $D_{ij} = 1 - \cos(\mathbf{e}_i,\mathbf{e}_j)$, updated when new algorithms enter $\mathcal{A}$.
$\mathbf{D}$ supports:
\begin{inparaenum}
    \item k-nearest-neighbor retrieval for targeted sampling,
    \item clustering to define operational embedding-space regions (\S\ref{subsec:3layerStrategy}), and
    \item similarity-based deduplication and pruning (\S\ref{subsec:archive}).
\end{inparaenum}

\subsection{Coverage-Preserving Archive}
\label{subsec:archive}

ATLAS maintains a bounded archive $\mathcal{A}$ of valid algorithms that serves as a semantic repertoire of the explored
algorithm space.
Each validated candidate is added to $\mathcal{A}$, after which similarity-based management removes redundancy
while preserving coverage of distinct regions in embedding space.
Here, multimodality is operationalized in the selected representation:
retaining competitive algorithms across separate embedding-space regions,
without treating each region as a verified mechanistic family.

\subsubsection{Similarity-Based Deduplication}
Using the distance matrix $\mathbf{D}$ from \S\ref{subsec:embeddings},
ATLAS removes near-duplicate algorithms at each iteration via two tiers (Algorithm~\ref{alg:atlas_dedup}):
\begin{inparaenum}
    \item \emph{strict deduplication} removes one algorithm from each pair where $D_{ij} < 1-\tau_{\text{strict}}$, and
    \item \emph{performance-aware deduplication} removes one algorithm from each pair where $D_{ij} <
      1-\tau_{\text{soft}}$ and their performance difference is below a small absolute tie tolerance $\epsilon$.
\end{inparaenum}
In both tiers, ties are broken uniformly at random.

\subsubsection{Capacity-Constrained Pruning}
If $|\mathcal{A}| > N_{\max}$, ATLAS iteratively removes the lower-performing algorithm from the most similar pair
$(a_i, a_j)$, where $(i,j) = \argmin_{i \neq j} D_{ij}$ (Algorithm~\ref{alg:atlas_prune}).
To prevent loss of refinement anchors, the elite region $\mathcal{G}^* = \{a^*\} \cup \text{k-NN}(a^*, k)$ is excluded.

\subsubsection{Effect}
Unlike selection-driven evolutionary methods that discard alternatives via global fitness ranking,
ATLAS removes algorithms that contribute the least additional coverage (i.e., those redundant under $\mathbf{D}$).
This similarity-first management retains representation across multiple embedding-space regions even
when their candidates are not currently optimal, while avoiding unbounded archive growth.

\subsection{Semantic Search Operators}
\label{subsec:operators}
ATLAS uses LLM-driven semantic operators, implemented as prompts, to create and transform complete
algorithms.
We organize the suite by the number of reference algorithms supplied to the LLM and by whether an
operator is triggered by evaluation failure.%
\footnote{In standard evolutionary computation (EC) terminology, each reference algorithm is a \emph{parent}.
Zero-, single-, and multi-reference operators therefore correspond to zero-, single-, and multi-parent operators,
respectively.
We retain \emph{reference} to emphasize that the complete algorithm source is supplied to the LLM.}
\textsc{Create} receives no input algorithm (zero-reference) and generates initial candidates directly
from the problem specification and I/O requirements.
\textsc{Improve}, \textsc{Tune}, and \textsc{Simplify} refine one archived reference, whereas
\textsc{Combine} and \textsc{Diverge} synthesize a new candidate from multiple archived references.
\textsc{Repair} instead receives a rejected candidate together with classified evaluation evidence and
regenerates it for full re-evaluation.

Creation, variation, and crossover operators are common in LLM-based algorithm
design~\cite{liu2024evolution}; ATLAS adapts these roles to full-algorithm synthesis.
Each operator prompt combines its search objective with the problem specification (the objective,
constraints, and solution semantics) and the external entry-point and I/O requirements needed to produce an executable
algorithm.
Permitted dependencies and the structured response format are implementation requirements rather than
a prescribed algorithmic scaffold.
Every operator requests one structured response containing the complete Python source code, an
algorithm name, and a self-contained description.
When an operator requires archived references, ATLAS selects them according to the current
initialization or search-layer policy, such as uniform archive sampling, current-best selection,
embedding-neighborhood selection, or cluster representatives.
The exact executable system message, operator instructions, problem-specific prompt components,
repair guidance, and response requirements are provided in the public source-code repository.%
\footnote{\nolinkurl{https://github.com/Danial-Yazdani/ATLAS}}

\subsubsection{Zero-Reference Generation}
\textbf{\textsc{Create}} generates complete algorithms from the problem
specification and I/O requirements without a reference algorithm, and is used only for initialization.
Its prompt also provides numerical example instances paired with valid solutions.
These solution-grounding examples clarify the required output structure
and constraint satisfaction without exposing implementation logic or favoring any algorithmic paradigm.

\subsubsection{Single-Reference Refinement}
Given one reference algorithm, these operators refine it while preserving its paradigm:
\begin{inparaenum}
    \item \textbf{\textsc{Improve}}:
      improves logic and efficiency without changing the fundamental approach;
    \item \textbf{\textsc{Tune}}: adjusts numerical parameters
      while keeping structure fixed;
    \item \textbf{\textsc{Simplify}}: reduces redundant code and complexity,
      acting as parsimony pressure against code bloat.
\end{inparaenum}

\subsubsection{Multi-Reference Synthesis}
Given multiple references, these operators synthesize new algorithms:
\begin{inparaenum}
    \item \textbf{\textsc{Combine}}:
      integrates complementary ideas from the references; nearby references provide a local context,
      while embedding-distant references provide cross-region context for potential hybridization;
    \item \textbf{\textsc{Diverge}}:
      generates algorithms explicitly different from the references, enabling large jumps across regions.
\end{inparaenum}

\subsubsection{Repair}
In component synthesis, the surrounding code generally retains responsibility for solution construction
and constraint handling, so the generated component need not produce and validate a complete feasible solution.
Full-algorithm synthesis transfers this responsibility to the candidate.
Its output may therefore violate capacity, time-window, permutation, uniqueness,
or completeness requirements even when the code executes without an exception.
Full-algorithm candidates also expose entry-point, output-format, runtime, and memory failures.

ATLAS records an infeasible or malformed returned solution as an evaluation failure
and rejects the candidate rather than ranking it using a finite penalty.
The external evaluator checks feasibility and independently recomputes the objective, but it never completes, modifies,
or makes a returned solution feasible. A failure on any evaluated instance invalidates the candidate.
For a candidate that reaches evaluation but fails, \textsc{Repair} routes the
aggregate failure evidence among six prompt modes: memory limit, partial failure, timeout, constraint violation,
runtime error, and framework or interface violation.
The prompt asks the LLM to regenerate the complete algorithm source with targeted corrections,
after which the candidate is fully re-evaluated.
Appendix~\ref{app:repair_operator} details the deterministic failure routing, identified
failure classes, correction objectives, and full re-evaluation used by \textsc{Repair}.

\subsection{Initialization}
\label{subsec:initialization}

ATLAS uses a two-phase bootstrap to establish diverse initial coverage (Algorithm~\ref{alg:atlas_init}).
\begin{inparaenum}
\item \textbf{Bootstrap generation:} We sample candidates with \textsc{Create} (\S\ref{subsec:operators}) at temperature
  $T{=}1.0$; despite stochasticity, the outputs often cluster around well-known textbook approaches.
\item \textbf{Reference-conditioned diversification:} To expand coverage,
  we iteratively apply \textsc{Diverge} (\S\ref{subsec:operators}), sampling references uniformly from the current
  archive and requesting approaches that differ from the referenced strategies.
\end{inparaenum}
This dual-phase initialization substantially improves initial coverage compared to sampling solely
with \textsc{Create} (Appendix~\ref{app:init_ablation}).

\subsection{Three-Layer Search Strategy}
\label{subsec:3layerStrategy}

ATLAS allocates operator applications across three complementary layers (Algorithm~\ref{alg:atlas_gen}):
concentrated refinement around the current best, distributed refinement across other represented regions,
and cross-region synthesis using components and interactions from multiple inputs.
This is implemented by clustering the archive in embedding space and operating either locally (within a region) or
globally (across regions).

\subsubsection{Archive Clustering}
All clustering is based on the distance matrix $\mathbf{D}$ (\S\ref{subsec:embeddings}).
At each iteration, the elite region $\mathcal{G}^{*}$ is formed from the current best algorithm $a^*$
and its $k$-nearest neighbors, and Layer~1 operates on this region.
We then apply k-medoids with automatic $K$ selection to the full archive (details in Appendix~\ref{app:clustering}),
yielding operational semantic regions for search.\footnote{Clusters are regions induced by the selected embedding;
mechanistic family identity is not inferred from clustering alone.}
The cluster containing $a^*$ is removed in full, and any other members of $\mathcal{G}^{*}$ are removed from the
remaining clusters.
Empty clusters are discarded, and a representative removed by this process is replaced by the best surviving member
according to the training objective.
Let $\Omega_t$ denote the resulting non-elite clusters and $R_t=|\Omega_t|$.
Layer~2 operates only on $\Omega_t$.
Layer~3 draws its non-elite representatives from $\Omega_t$, while its quality-anchored \textsc{Combine}
and \textsc{Diverge} streams additionally include $a^*$ as a reference.

\subsubsection{Layer Roles}
\textbf{Layer~1 (elite-region refinement)} focuses on $a^*$ and its neighborhood
through repeated single-reference refinement and local \textsc{Combine}.
\textbf{Layer~2 (within-region maturation)} applies single-reference refinement to each non-elite representative and,
when enough members remain, local \textsc{Combine} within its surviving cluster.
This gives a representative from a currently inferior region a dedicated proposal opportunity without requiring
global-fitness parent selection.
The search-layer ablation evaluates this Layer~2 contribution conditional on Layer~3
(Appendix~\ref{app:layer_ablation}).
\textbf{Layer~3 (cross-region discovery)} applies multi-reference operators to representatives
through three synthesis streams with different reference contexts:
\begin{inparaenum}
\item \textsc{Combine} between representatives without an $a^*$ anchor,
\item \textsc{Combine} between $a^*$ and representatives for quality-anchored cross-region synthesis, and
\item \textsc{Diverge} using representatives together with $a^*$ as references (instructions intended to move proposals
  away from the referenced archive regions; see Appendix~\ref{app:note_layer3_strategies}).
\end{inparaenum}
The complete three-layer configuration performs best among the four layer variants;
the scope of the supported comparisons is detailed in Appendix~\ref{app:layer_ablation}.

\subsubsection{Archive-Conditional Allocation}
The operator policy and probabilities are fixed in advance.
Layer~1 uses the configured elite-region and intensification settings,
whereas the nominal work assigned to Layers~2 and~3 scales with the surviving cluster count $R_t$; in particular,
each Layer~3 stream schedules $R_t$ candidates, subject to the remaining global budget.
Because clustering, elite membership, and representatives are recomputed at every iteration,
the scale and reference context of the schedule change with the archive,
and Layer~1 recenters whenever the incumbent best changes.
The detailed operator schedule and settings are given in Algorithm~\ref{alg:atlas_gen}
and Appendix~\ref{app:hyperparameters}.

This structure is related in motivation to island-model search and classic MAP-Elites,
which continue sampling alternatives rather than selecting references solely by global
fitness~\cite{romera2024mathematical,mouret2015illuminating}.
The mechanism is different.
Unlike an island model, ATLAS has no persistent demes connected by migration; unlike classic MAP-Elites,
it has no fixed task-specific behavior-descriptor grid with cell-wise replacement.
Instead, it maintains one global archive and recomputes transient embedding-induced clusters for search allocation.

\section{Experiments}
\label{sec:experiments}

This section first describes the benchmark problems, comparison methods, evaluation protocol,
and implementation settings used to assess ATLAS.
It then reports the main comparisons, generalization beyond the default benchmark settings,
and the training-selected archive representatives.
Finally, component ablations and design analyses examine the principal choices underlying the search.

\subsection{Experimental Setup}
\label{subsec:ExperimentalSetup}

\subsubsection{Benchmark Problems}
We evaluate on four NP-hard combinatorial optimization problems: Capacitated Vehicle Routing Problem
\textbf{(CVRP)}~\cite{dantzig1959truck}, which minimizes travel distance under vehicle capacity constraints;
CVRP with Time Windows \textbf{(CVRPTW)}~\cite{solomon1987algorithms},
which adds customer time-window constraints to CVRP; Flow Shop Scheduling \textbf{(FSS)}~\cite{garey1976complexity},
which minimizes makespan in a fixed machine order; and Quadratic Assignment Problem
\textbf{(QAP)}~\cite{koopmans1957assignment}, which minimizes flow-distance assignment cost between facilities
and locations.
Together, they cover routing, scheduling, and assignment, spanning different constraint structures and objective types.
Problem formulations and instance-generation details are provided in Appendices~\ref{app:problem_definitions}
and~\ref{app:instance_generation}.
The exact problem descriptions and I/O requirements presented to the LLM are provided in the public repository.

\subsubsection{Baselines}
We compare against three groups of baselines (implementation details in Appendix~\ref{app:baselines}):
\paragraph{Human-designed, domain-specific methods}
For CVRP and CVRPTW, we use PyVRP~\cite{wouda2024pyvrp,pyvrp0133software}, Google OR-Tools~\cite{ortools},
and VROOM~\cite{vroom,vroom116software}.
For FSS, we use NEH~\cite{nawaz1983heuristic}, Taillard-accelerated Iterated Greedy with
idle-time tie-breaking (IG-TB)~\cite{ruiz2007simple,taillard1990some,fernandez2014insertion},
and Iterative Beam Search~\cite{libralesso2022iterative}.
For QAP, we use Robust Tabu Search (RoTS)~\cite{taillard1991robust}, Simulated Annealing~\cite{connolly1990improved},
Breakout Local Search (BLS)~\cite{benlic2013breakout}, and Memetic Search (BMA)~\cite{benlic2015memetic}.
Runtime-sensitive methods are evaluated under the corresponding benchmark-setting-specific per-instance runtime caps;
implementation provenance and configurations are provided in Appendix~\ref{app:classical_baselines}.

\paragraph{Component-based LLM methods}
We include \textsc{ReEvo}~\cite{ye2024reevo}, \textsc{EoH}~\cite{liu2024evolution}, and MCTS-AHD~\cite{zheng2025monte}.
Because their official repositories target different problems and use non-uniform scaffolds, prompts, and evaluators,
we evaluate them under a shared greedy-construction scaffold and common benchmark interface based on
LLM4AD~\cite{liu2024llm4ad}, while aligning method-specific search settings to the official repositories.
All LLM-based methods use the same LLM backend and matched token-budget-based termination criteria;
see Appendix~\ref{app:llm_baselines}.

\paragraph{Full-synthesis baseline (\textsc{EoH-Full})}
We construct \textsc{EoH-Full} by adapting \textsc{EoH}'s evolutionary framework to evolve full algorithms
using ATLAS's problem- and interface-aware operators, repair, and evaluation framework
while retaining \textsc{EoH}'s fitness-driven selection.
Comparing component-based \textsc{EoH} with \textsc{EoH-Full} evaluates the combined practical effect of moving from
a fixed component scaffold to scaffold-free full-algorithm synthesis with the machinery needed to support it.
Comparing \textsc{EoH-Full} with ATLAS then isolates the contribution of ATLAS's archive-based embedding-guided
quality-diversity search from standard fitness-driven evolution under matched full-synthesis conditions;
see Appendix~\ref{app:eoh_full}.

\subsubsection{Evaluation Protocol}
We evaluate methods using the following protocol:
\paragraph{Benchmark settings and data splits}
We consider four \textit{default} benchmark settings: FSS with 50 jobs and 10 machines, CVRP with 50 customers,
CVRPTW with 50 customers, and QAP with 50 facilities.
For each setting, the reported protocol uses 31 training instances (split seed 2024)
and 31 independently generated test instances (split seed 42).
Search and archive construction access only the training split.
The algorithm selected using training performance is then evaluated once on the test set.
Split-local random-number generators and immutable instance manifests keep both sets disjoint
and reproducible (Appendix~\ref{app:instance_generation}).
    
\paragraph{Independent runs}
LLM-based methods (ATLAS, \textsc{EoH}, \textsc{ReEvo}, \textsc{MCTS-AHD},
and \textsc{EoH-Full}) perform $R=5$ independent synthesis runs with different random seeds,
each yielding one training-selected optimization algorithm; the algorithms may differ across runs.
Five runs per method is set by the substantial LLM API cost of generation and the computational cost of
repeatedly evaluating candidate algorithms throughout each search.
Human-designed, domain-specific baselines are evaluated directly on the test instances under the same
benchmark-setting-specific runtime caps.
    
\paragraph{LLM configuration and budgets}
All LLM-based synthesis methods use \textsc{OpenAI GPT-5-mini} with reasoning effort set to \textit{low}
and temperature $T=1.0$.
For end-to-end ATLAS component ablations and search-configuration sensitivity analyses,
we fix the search budget to $B=500$ evaluated operator executions.
The embedding and initialization analyses instead use the study-specific sample sizes reported in
Appendix~\ref{app:ablations}.
For cross-method comparisons, we match total token consumption per synthesis run,
since per-iteration token usage differs substantially between component synthesis and full-algorithm synthesis.
The resulting benchmark-specific token budgets, calibrated from the average token usage of 500 evaluated operator
executions under full synthesis, are 5M for FSS, 6.5M for CVRP, 7M for CVRPTW, and 6M for QAP.

\paragraph{Full synthesis evaluation}
Each candidate algorithm is executed on each instance to produce a solution.
That solution is then passed to a problem-specific external evaluator that checks output format,
detects constraint violations, and computes the objective value (Appendix~\ref{app:evaluation}).
    
\paragraph{Runtime caps and execution model}
Each benchmark setting uses a common per-instance runtime cap for all compared methods: 210\,s for FSS, 30\,s for CVRP,
120\,s for CVRPTW, and 240\,s for QAP.
All methods are executed under single-core, single-threaded settings whenever applicable,
making these runtime caps more comparable across synthesized and human-designed methods.
For the routing comparisons, performance-critical components of PyVRP
and the OR-Tools routing engine execute as compiled C++ code through Python interfaces,
whereas the algorithms synthesized by the LLM-based methods, including ATLAS, execute as Python programs.
Equal wall-clock limits therefore control the available time but do not remove implementation-language overhead:
for comparable low-level operations, the compiled baselines can generally perform more search iterations or neighborhood
evaluations within the cap.
Thus, this aspect of the runtime protocol favors the human-designed CVRP
and CVRPTW baselines over the LLM-based methods.
The full cap is applied unchanged during training and test execution.
    
\paragraph{Reporting}
At the end of each ATLAS run, the algorithm with the best training performance is selected and evaluated on the 31
test instances.
Test outcomes cannot trigger fallback, re-ranking, repair, archive mutation, or a repeat test evaluation.
Any timeout, memory violation, exception, invalid output, non-finite objective,
or infeasible solution on any test instance causes the entire run to be classified as failed.
The outcome ledger retains the status and failure reason for every failed instance.
This failure rule was not triggered in any reported LLM-based synthesis run.

\paragraph{Reader-facing objective display}
For every problem and benchmark setting, paired tests on the raw per-instance objectives identify the
statistically best-performing group of human-designed, domain-specific methods.
Within that group, we designate the method with the lowest reported mean objective as the reference.
If $\overline{J}_{m,s}$ is method $m$'s reported mean objective in setting $s$
and $\overline{J}_{\mathrm{ref},s}$ is that reference mean, we report the relative mean gap
$100(\overline{J}_{m,s}-\overline{J}_{\mathrm{ref},s})/\overline{J}_{\mathrm{ref},s}$. A value of zero matches the
reference mean, a positive value is worse, and a negative value is better.
This descriptive quantity is a gap to the best evaluated domain-specific reference, not an optimality gap.
The same transformation is used for the archive-representative tables,
where test gaps are computed from the reported representative mean objectives only after the representatives have been
fixed using training information.
The corresponding reference means are reported in Table~\ref{app:classical_reference_means}.

\paragraph{Statistical testing}
For each test instance, we average the raw objective values from the five independent runs of each LLM-based method;
human-designed methods contribute their direct per-instance raw objectives.
This averaging marginalizes over synthesis runs and yields one raw objective value per method and test instance.
Pairwise comparisons between all methods use a two-sided Wilcoxon signed-rank test at significance level $0.05$ on
the resulting 31 pairs of per-instance raw objective values.
Percentage gaps are descriptive only and are not inputs to the test.
In the result tables, boldface identifies the statistically best-performing group across all methods:
the method with the lowest mean and any method not significantly different from it.
Every textual claim that one method achieves better test-set objective performance than another is based on this paired
test (Appendix~\ref{app:eval_protocol_stats}).

\subsubsection{ATLAS Configuration}
Table~\ref{tab:atlas_config} summarizes the key hyperparameters of ATLAS.

\begin{table}[h]

\centering
\caption{\emph{ATLAS configuration.}
Complete details and rationale in Appendix~\ref{app:hyperparameters}.}
\label{tab:atlas_config}
\footnotesize
\begin{tabular}{@{}ll@{\hspace{2em}}ll@{}}
\toprule
\textbf{Parameter} & \textbf{Value} & \textbf{Parameter} & \textbf{Value} \\
\midrule
Archive capacity & $N_{\max}=100$ & Layer 1 size & $|\mathcal{G}^*|=5$ \\
Initialization & 50 algorithms & Clustering & K-Medoids, adaptive $K$ \\
\midrule
\multicolumn{2}{@{}l}{\textit{Multi-ref operators (refs)}} & \multicolumn{2}{@{}l}{\textit{Single-ref operator probs}} \\
\textsc{Combine} & $m_{\text{comb}}=2$ &  \textsc{Improve}, \textsc{Tune} & $p=0.5$ \\
\textsc{Diverge} & $m_{\text{div}}=3$ &  \textsc{Simplify} & $p=0.2$ \\
\midrule
Embedding model & \textsc{mGTE-large-en-v1.5}~\cite{zhang2024mgte} & LLM & \textsc{GPT-5-mini} (T=1.0, LOW) \\
Budget & 500 ops or benchmark-setting-specific tokens & Timeout & benchmark-setting-specific \\
Train/test & $31/31$ & Train/test split seeds & $2024/42$ \\
\bottomrule
\end{tabular}

\end{table}

\subsubsection{ATLAS Implementation}
Implementation details, including pseudocode, hyperparameters, and the evaluation pipeline,
are provided in Appendix~\ref{app:ImplementationDetails}.
The complete executable prompt definitions are provided in the public repository.

\paragraph{Hardware and environment}
All experiments were run on WSL2 (Ubuntu 22.04 LTS) using Python 3.12, hosted on a workstation with an AMD Ryzen 9 7950X (32 threads), 64\,GB physical RAM (30\,GB allocated to WSL2, 8\,GB swap), and an NVIDIA GeForce RTX 5090 (32\,GB VRAM).
Dependency constraints and installation instructions are provided in the \texttt{requirements.txt} file in the ATLAS repository.
All compared methods were executed under single-core, single-threaded settings whenever applicable: each candidate--instance invocation executes within one worker process and one thread, while different instances are evaluated in parallel; restricting human-designed, domain-specific baselines to the same per-instance execution model makes the runtime caps more comparable across methods.

\paragraph{Per-run resource usage and cost}
Under the default ATLAS settings in Table~\ref{tab:atlas_config} (500 operator calls and approximately 5M--7M total tokens per synthesis run, with roughly two-thirds input and one-third output, including reasoning tokens), the estimated OpenAI API cost is approximately \$4.2--\$5.8 per run using \textsc{GPT-5-mini} (LOW) at current standard pricing.
The wall-clock time of a synthesis run depends on the benchmark setting, API latency, and the execution time of the candidate algorithms being evaluated; under the default budget of 500 evaluated operator executions, a typical synthesis run required approximately 10--15 hours in our setup.
In practical settings, this automated synthesis cost is modest relative to the engineering effort required to manually design and tune a competitive problem-specific optimization algorithm.

\subsection{Main Experimental Results}
\label{sec:baseline-comparison}

\subsubsection{Comparison with Baselines}
The paired Wilcoxon analysis shows that ATLAS statistically outperforms every LLM-based baseline
across all four default benchmark settings (Table~\ref{tab:baseline-comparison-gap}).
The component-synthesis implementations of \textsc{EoH}, \textsc{ReEvo},
and \textsc{MCTS-AHD} share the same greedy-construction scaffold and therefore search only within
that controlled artifact interface.
Among the component-synthesis methods, \textsc{EoH} achieves the best results in three of the four default
settings, while \textsc{MCTS-AHD} performs best on FSS.

The full-synthesis baseline \textsc{EoH-Full} substantially improves over the component-based
implementations.
This comparison demonstrates the benefit of the combined change from their shared greedy component interface to ATLAS's
full-algorithm artifact and corresponding problem- and interface-aware operators;
it should not be interpreted as isolating scaffold freedom alone.
Because \textsc{EoH-Full} and ATLAS share the full-synthesis operators, repair, evaluator, LLM, and budgets,
ATLAS's remaining advantage over \textsc{EoH-Full} more directly supports the contribution of its archive-based embedding-guided
quality-diversity search.
Fitness-driven selection used in \textsc{EoH-Full} can concentrate effort around early promising regions,
whereas ATLAS preserves coverage across embedding-separated regions, refines them in parallel,
and performs cross-cluster synthesis.

ATLAS is also competitive with the human-designed, domain-specific baselines.
Its relative mean gaps to the best such reference are 0.347\% on FSS, 0.094\% on CVRP, 0.000\% on CVRPTW,
and 0.460\% on QAP.
On CVRP, the paired Wilcoxon test finds no statistically significant difference between ATLAS and PyVRP,
the selected human-designed reference.
ATLAS also matches the lowest reported CVRPTW mean at the displayed precision
and remains within 0.46\% of the reference mean on all four default settings,
despite synthesizing an optimization algorithm automatically for each problem.
The reported gaps and displayed-precision comparisons are descriptive;
statistical comparisons use paired raw objectives.
Examples of best-found algorithms generated by ATLAS, one for each problem,
are available with their descriptions in the
\href{https://github.com/Danial-Yazdani/ATLAS/tree/main/SAMPLE_SYNTHESIZED_ALGORITHMS}{source-code repository}.

\begin{table}[tp]
\centering
\footnotesize
\caption{\emph{Comparison with baselines on the default benchmark settings using relative mean gaps.}
Values are relative mean gaps (\%) to the statistically selected human-designed reference in each setting;
lower is better.
The reference in each column is the human-designed, domain-specific method with the lowest mean within the
statistically best-performing human-designed group and therefore has gap 0.000\%.
Statistical tests use the paired raw objectives, not these gaps.
Boldface identifies the statistically best-performing group across all methods under the paired Wilcoxon signed-rank
test: the method with the lowest mean and any method not significantly different from it.
Light-gray cell shading identifies the best LLM-based method in each setting.}
\label{tab:baseline-comparison-gap}
\small
\begin{tabular}{llcccc}
\toprule
\textbf{Group} & \textbf{Method} & \textbf{FSS} $(n50m10)$ & \textbf{CVRP} $(n50)$ & \textbf{CVRPTW} $(n50)$ & \textbf{QAP} $(n50)$ \\
\midrule
\multirow{10}{*}{Domain-specific}
& NEH & 2.809\% & -- & -- & -- \\
& IG-TB (Taillard acc.) & \textbf{0.000\%} & -- & -- & -- \\
& Iterative Beam Search & 1.787\% & -- & -- & -- \\
& OR-Tools (GLS) & -- & 1.430\% & 0.121\% & -- \\
& PyVRP (ILS) & -- & \textbf{0.000\%} & \textbf{0.000\%} & -- \\
& VROOM & -- & \textbf{0.095\%} & \textbf{0.060\%} & -- \\
& Robust Tabu Search & -- & -- & -- & 0.455\% \\
& BLS & -- & -- & -- & \textbf{0.015\%} \\
& BMA & -- & -- & -- & \textbf{0.000\%} \\
& Simulated Annealing & -- & -- & -- & 1.405\% \\
\midrule
\multirow{5}{*}{LLM-based}
& \textsc{EoH} & 3.909\% & 21.106\% & 20.711\% & 1.520\% \\
& \textsc{ReEvo} & 3.912\% & 29.071\% & 35.161\% & 1.986\% \\
& \textsc{MCTS-AHD} & 2.837\% & 24.393\% & 21.505\% & 2.012\% \\
& \textsc{EoH-Full} & 0.782\% & 3.815\% & 4.520\% & 1.324\% \\
& ATLAS (ours) & \cellcolor{gray!20}0.347\% & \cellcolor{gray!20}\textbf{0.094\%} & \cellcolor{gray!20}\textbf{0.000\%} & \cellcolor{gray!20}0.460\% \\
\bottomrule
\end{tabular}%
\end{table}

\begin{table}[tp]
\centering
\footnotesize
\caption{\emph{Generalization beyond the default benchmark settings using relative mean gaps
(re-run from scratch).}
Values are relative mean gaps (\%) to the statistically selected human-designed reference for every problem size.
Lower is better.
Boldface identifies the statistically best-performing group across all methods under the paired Wilcoxon signed-rank
test: the method with the lowest mean and any method not significantly different from it.
Light-gray cell shading identifies the best LLM-based method in each setting.
}
\label{tab:scaled_results_gap}
\small\begin{tabular}{llcccc}
\toprule
\textbf{Group} & \textbf{Method} & \textbf{FSS} $(n25m5)$ & \textbf{FSS} $(n100m20)$ & \textbf{CVRP} $(n25)$ & \textbf{CVRP} $(n100)$ \\
\midrule
\multirow{6}{*}{Domain-specific}
& NEH & 1.143\% & 4.133\% & -- & -- \\
& IG-TB (Taillard acc.) & \textbf{0.000\%} & \textbf{0.000\%} & -- & -- \\
& Iterative Beam Search & 2.321\% & 2.865\% & -- & -- \\
& OR-Tools (GLS) & -- & -- & \textbf{0.162\%} & 4.930\% \\
& PyVRP (ILS) & -- & -- & \textbf{0.000\%} & \textbf{0.000\%} \\
& VROOM & -- & -- & \textbf{0.000\%} & 0.975\% \\
\midrule
\multirow{3}{*}{LLM-based}
& \textsc{EoH} & 0.310\% & 6.810\% & 23.662\% & 16.371\% \\
& \textsc{EoH-Full} & 0.515\% & 2.529\% & 1.919\% & 5.801\% \\
& ATLAS (ours) & \cellcolor{gray!20}\textbf{0.029\%} & \cellcolor{gray!20}1.651\% & \cellcolor{gray!20}\textbf{0.000\%} & \cellcolor{gray!20}1.820\% \\
\bottomrule
\end{tabular}%
\end{table}

\subsubsection{Generalization Beyond Default Benchmark Settings}
We further evaluate ATLAS on FSS and CVRP at half and double the default problem scale,
with runtime caps scaled accordingly.
For each scaled setting, synthesis is re-run from scratch on newly generated train/test instances at that scale.

ATLAS achieves the best results among the LLM-based methods in all four additional settings.
On FSS, ATLAS has relative mean gaps of 0.029\% and 1.651\% at the smaller and larger scales, respectively.
At $n25m5$, the paired Wilcoxon test finds no statistically significant difference between ATLAS and IG-TB.
On CVRP, it matches the best reported domain-specific mean at $n25$ and has a 1.820\% gap at $n100$.

Interestingly, on the smaller FSS setting, \textsc{EoH} outperforms \textsc{EoH-Full},
suggesting that the shared greedy-construction scaffold used by \textsc{EoH} remains particularly effective in this
easier setting.

Our cross-scale investigation suggests that the performance of the synthesized algorithms becomes
more sensitive to numerical hyperparameter settings as problem scale increases.
This increased sensitivity may contribute to the wider gaps observed at larger scales.
The current \textsc{Tune} operator proposes numerical changes based on the candidate source code.
Each proposal is then assessed through the standard ATLAS evaluation and selection process,
but the operator does not perform dedicated feedback-driven hyperparameter optimization.
Accurately calibrating sensitive hyperparameters normally requires repeated empirical evaluations.
Consequently, proposals based only on source code are unlikely to identify robust settings reliably.

\FloatBarrier
\subsubsection{Training-Selected Archive Representatives}
To illustrate ATLAS's multimodal retention across embedding-space regions and the structures it discovers,
Table~\ref{tab:atlas_representatives} reports the complete selected representative set retained in the final archive of a single synthesis run, the run among the $R{=}5$ synthesis runs whose deployed algorithm achieved the best training objective, for each problem.
Each representative is selected using training information as the training-best member of a final embedding-space cluster,
so representative identities are fixed before test evaluation.
The descriptive test results show that multiple representatives for every problem lie within 1\% of
the selected human-designed reference mean for that problem.
The fixed representatives are subsequently ordered by their test relative mean gaps solely to make their quality and
structural differences interpretable.
The listed names and descriptions were manually checked to ensure that they are broadly consistent with the
corresponding synthesized algorithms.

\begin{table}[H]
\centering
\caption{\emph{Training-selected archive representatives with descriptive test performance.}
All representatives are drawn from the final archive of a single synthesis run: the run, among the $R{=}5$ synthesis runs, whose deployed algorithm achieved the best training objective.
Representative identities are fixed from training objectives as the training-best members of final embedding-space clusters.
Values in parentheses are test-set relative mean gaps (\%) to the statistically selected human-designed,
domain-specific reference; lower is better.
Ordering by these gaps is descriptive only and affects neither selection nor statistical tests.}
\label{tab:atlas_representatives}
\footnotesize
\setlength{\tabcolsep}{4pt}
\begin{tabular}{@{}p{1.2cm}p{16.4cm}@{}}
\toprule
\textbf{Problem} & \textbf{Training-selected representatives, ordered by descriptive test gap} \\
\midrule

\textbf{FSS} $(n50m10)$ &
Priority-Seeded Prefix-Guided LNS with Lightweight Anneal (0.331\%);
Priority-seeded NEH with Guided Neighborhood Search (0.458\%);
Priority-Spectral Hybrid with Tabu-Repair Local Search (0.472\%);
Prefix-Aware Hybrid Genetic with Annealed Polish (0.478\%);
Sampled Pairwise Greedy with Bounded Cache and Adaptive Iterated Refinement (0.527\%);
Hybrid NEH-Regret Genetic Search with Aggressive Complexity Reduction (0.714\%);
Guided Greedy Adaptive Cache with Pheromone and Block Relocation (0.965\%);
Bottleneck-Guided Greedy with ILS and Simulated Annealing (1.239\%) \\
\cmidrule(lr){2-2}

\textbf{CVRP} $(n50)$ &
Hybrid Multi-Start Clustered-Repair Large Neighborhood Search (0.000\%);
Multi-Start Regret Insertion with Focused Large Neighborhood Search (0.095\%);
Spectral-Seeding Regret-LNS with Focused Savings (0.191\%);
Hybrid Demand-Seeding with Localized Destroy-and-Repair (3.622\%);
Clustered Demand-Aware Insertion with Focused LNS (3.622\%);
Adaptive Clustered Chain-Savings Solver (3.908\%) \\
\cmidrule(lr){2-2}

\textbf{CVRPTW} $(n50)$ &
Large-Neighborhood Search with Time-Aware Regret Repair (0.000\%);
Ruin-and-Recreate Vehicle Routing with Time Windows (0.000\%);
Hybrid Label-Constructive and Ruin-Repair Routing (0.000\%);
Hybrid Adaptive Ruin-and-Repair with Temporal Beam Seeding (0.060\%);
Hybrid Guided Repair with Selective Savings Merge (0.302\%);
Permutation-Guided Large Neighborhood Search (2.171\%);
Radial and Time-Priority Insertion with Neighbor Guidance (3.317\%) \\
\cmidrule(lr){2-2}

\textbf{QAP} $(n50)$ &
Hierarchical Spectral-Memetic Solver with Block Neighborhoods and Cycle Diversification (0.440\%);
Soft-Probabilistic Beam with Block Reassignment and Tabu Polishing (0.461\%);
Adaptive Spectral-Pheromone Memetic Solver (0.496\%);
Spectral-Guided Adaptive Large-Neighborhood Search with Local Intensification (0.525\%);
Spectral-Softassign Hybrid with Mixed Local Search (0.813\%) \\
\bottomrule
\end{tabular}
\end{table}
\FloatBarrier

\paragraph{Mechanism diversity and hybridization}

Code inspection of the displayed algorithms in Table~\ref{tab:atlas_representatives} identifies differences in core construction or search components together
with some shared refinement mechanisms.
For example, in CVRP and CVRPTW, several representatives differ in their primary construction or global-search logic, including regret
insertion, savings-based construction, clustering, ruin-and-recreate, and label- or beam-inspired seeding, while often
converging on strong repair or large-neighborhood refinement components.
In FSS, the representatives span different core search patterns, including NEH-style construction, greedy or pairwise
search, genetic search, tabu-guided local search, and large-neighborhood refinement.
In QAP, the representatives likewise cover different core search engines, including memetic search,
pheromone-guided search, beam-style search, tabu-guided local improvement, and large-neighborhood refinement.

This pattern of structural diversity, component reuse, and hybridization is consistent with ATLAS's design:
the archive preserves representatives from multiple
embedding-space regions, while Layer~3 can draw components and their interactions from different input algorithms and
reorganize them into new hybrids.
The displayed repertoire therefore demonstrates that scaffold-free synthesis can retain several competitive,
structurally different, multi-component algorithms rather than only one best design.
It does not by itself establish archive-wide functional validity of the embedding clusters or per-instance portfolio
complementarity.

\subsection{Component Ablations and Design Analyses}
\label{subsec:ablations}

We report the search-layer ablation and summarize the principal design analyses here; complete setups,
results, and discussion for both the component ablations and design analyses are provided in
Appendix~\ref{app:ablations}.

As a controlled comparative experiment rather than a component ablation,
Table~\ref{tab:baseline-comparison-gap} shows that \textsc{EoH-Full}'s full-algorithm artifact and problem-
and interface-aware operators improve over the shared-scaffold component implementations.
ATLAS then improves over \textsc{EoH-Full} under shared full-algorithm operators, repair, evaluator, LLM, and budgets,
providing evidence for the framework-level value of its archive-based semantic quality-diversity search organization
beyond the shared full-synthesis machinery, without isolating every internal search mechanism.

The layer ablation in Table~\ref{tab:layer-ablation-gap} shows
that the complete three-layer configuration performs best among the four layer configurations.
Comparing Variant~A with Variant~C isolates the addition of Layer~1 to Layers~2+3
and shows the value of intensive refinement in the elite region.
Comparing Variant~C with Variant~D isolates the addition of Layer~2 when Layer~3 remains active
and supports representative-level refinement.
Variant~D (Layer~3 only) also outperforms Variant~B (Layer~1 only), showing
that the Layer~3-only configuration is stronger than the Layer~1-only configuration in these settings.
Full details are provided in Appendix~\ref{app:layer_ablation}.

\begin{table}[t]
\centering
\caption{\emph{Relative mean gaps for the layer ablation.}
Percent gap to the statistically selected human-designed reference in each setting
(PyVRP for CVRP and IG-TB for FSS);
lower is better.
Each gap is computed from the configuration's mean raw objective and the corresponding reference mean.
Boldface identifies the statistically best-performing group among the ablation configurations from paired tests on raw
per-instance objectives; the displayed gaps are descriptive and are not test inputs.
Each run is represented by its training-selected algorithm; the test set is used once for evaluation without fallback,
and failed runs are retained.
Detailed setup and analysis are provided in Appendix~\ref{app:layer_ablation}.}
\label{tab:layer-ablation-gap}
\footnotesize
{
\begin{tabular}{@{}lcc@{}}
\toprule
\textbf{Configuration} &
\shortstack{\textbf{CVRP} $(n50)$\\gap (\%)$\downarrow$} &
\shortstack{\textbf{FSS} $(n50m10)$\\gap (\%)$\downarrow$} \\
\midrule
Variant A: All layers   & \textbf{0.094} & \textbf{0.347} \\
Variant B: Layer~1 only & 3.620 & 0.814 \\
Variant C: Layers~2+3   & 0.668 & 0.474 \\
Variant D: Layer~3 only & 2.095 & 0.672 \\
\bottomrule
\end{tabular}}

\end{table}

The appendix reports one further search-guidance ablation and four supporting analyses:
\begin{inparaenum}
\item \textbf{Embedding representation analysis:} \textsc{mGTE-large-en-v1.5} with concatenated name,
  description, and preprocessed code is the strongest single-encoder embedding strategy
  (Appendix~\ref{app:embedding_ablation});
\item \textbf{Initialization-strategy analysis:} dual-phase initialization
  with \textsc{Diverge} yields substantially greater initial embedding-space dispersion than prompt variation alone
  while maintaining high validity (Appendix~\ref{app:init_ablation});
\item \textbf{Semantic search-guidance ablation:} semantic neighbor selection
  and grouping improve search guidance over fitness-based and random alternatives
  (Appendix~\ref{app:clustering_ablation});
\item \textbf{Reasoning-effort sensitivity and cost analysis:} for \textsc{GPT-5-mini},
  Low provides the best quality--cost trade-off, with Medium and High substantially increasing token usage for only
  marginal gains (Appendix~\ref{app:reasoning_ablation});
\item \textbf{Underlying-LLM comparison:} ATLAS remains effective across several LLM backends,
  with stronger models performing better but the differences remaining moderate
  (Appendix~\ref{app:differentLLMs}).
\end{inparaenum}

\section{Conclusion}
\label{sec:conclusion}

We have presented ATLAS, a framework for scaffold-free full-algorithm synthesis in
combinatorial optimization.
The problem specification supplies the objective and constraints, while a minimal external I/O interface defines how
instances are provided and solutions are returned without prescribing internal algorithmic structure.
Each generated algorithm is responsible for satisfying the problem requirements,
and full-algorithm operators leave its internal component inventory and control flow to synthesis.
By organizing complete algorithms in an embedding-guided coverage archive
and allocating search across three complementary layers, ATLAS supports multimodal search, refinement,
and recombination across multiple embedding-space regions.
Across four combinatorial optimization benchmarks, it outperforms several state-of-the-art component-synthesis
methods and the matched full-synthesis baseline, while remaining highly competitive with strong human-designed,
domain-specific methods.
These results support the combination of full-algorithm synthesis under a fixed I/O interface
with embedding-guided quality-diversity search for multimodal retention and hybrid synthesis.

Future work will build directly on our empirical analysis.
Synthesized full algorithms, particularly hybrid designs, can contain several algorithm-specific hyperparameters.
Our inspection suggests that more effective numerical configuration could narrow part of the remaining performance
gap between promising ATLAS-generated algorithms and the strongest human-designed methods.
The current \textsc{Tune} operator proposes changes from source code without conducting dedicated empirical
hyperparameter optimization.
LLaMEA-HPO addresses this challenge by coupling LLM-driven structural generation with in-loop hyperparameter
optimization (HPO), reporting competitive or improved performance with fewer LLM queries~\cite{vanstein2025hpo}.
Integrating training-only HPO into ATLAS is therefore a promising direction for improving the generated algorithms.
A complementary direction concerns how ATLAS distributes search effort across its archive.
ATLAS preserves and refines representatives across embedding-space regions to maintain diverse search opportunities.
A more adaptive resource-allocation mechanism could use accumulated training evidence to concentrate refinement on
promising regions while retaining sufficient coverage for continued exploration and cross-region synthesis.
This could improve search efficiency without sacrificing the diversity supported by the archive.

{\small

\bibliography{bib}
\bibliographystyle{abbrvnat}
}
%%%%%%%%%%%%%%%%%%%%%%%%%%%%%%%%%%%%%%%%%%%%%%%%%%%%%%%%%%%%%%%%%%%%%%%%%%%%%%%
%%%%%%%%%%%%%%%%%%%%%%%%%%%%%%%%%%%%%%%%%%%%%%%%%%%%%%%%%%%%%%%%%%%%%%%%%%%%%%%
% APPENDIX
%%%%%%%%%%%%%%%%%%%%%%%%%%%%%%%%%%%%%%%%%%%%%%%%%%%%%%%%%%%%%%%%%%%%%%%%%%%%%%%
%%%%%%%%%%%%%%%%%%%%%%%%%%%%%%%%%%%%%%%%%%%%%%%%%%%%%%%%%%%%%%%%%%%%%%%%%%%%%%%
\newpage
\appendices
\onecolumn
\makeatletter
\@addtoreset{figure}{section}
\@addtoreset{table}{section}
\@addtoreset{equation}{section}
\@addtoreset{algocf}{section}
\renewcommand{\thefigure}{\thesection.\arabic{figure}}
\renewcommand{\thetable}{\thesection.\arabic{table}}
\renewcommand{\theequation}{\thesection.\arabic{equation}}
\renewcommand{\thealgocf}{\thesection.\arabic{algocf}}
\makeatother

\section*{Appendix Table of Contents} % Optional title for the list
\startcontents[appendices]
\printcontents[appendices]{l}{1}{\setcounter{tocdepth}{2}}

\newpage
\section{Extended Related Work and Comparative Analysis}
\label{app:extended_related_work}
\input{Appendix/RelatedWork}

\section{Implementation Details}
\label{app:ImplementationDetails}

The exact prompts used by ATLAS are maintained with the executable implementation in the public
\href{https://github.com/Danial-Yazdani/ATLAS}{source-code repository}.
Each LLM request combines a system instruction with a user message assembled in
\texttt{framework/prompt\_system.py} from the operator objective, problem specification,
external I/O requirements, structured response requirements, and each selected reference's complete name,
description, and source code when applicable.
The problem-specific components are defined in \texttt{framework/problem\_definitions.py}.
The implementation also records the numerical examples supplied to \textsc{Create}
and the failure-conditioned guidance supplied to \textsc{Repair}.

%------------------------------------------------------------------------------
\subsection{Formal Problem Definitions}
\label{app:problem_definitions}
\input{Appendix/Problems}

%------------------------------------------------------------------------------
\subsection{Benchmark Instance Generation}
\label{app:instance_generation}
\input{Appendix/InstanceGeneration}

%------------------------------------------------------------------------------
\subsection{Evaluation Protocol}
\label{app:evaluation_protocol}
\input{Appendix/EvaluationProtocol}

%------------------------------------------------------------------------------
\subsection{ATLAS: Algorithms and Pseudocode}
\label{app:ATLASalgorithms}
\input{Appendix/ATLASProcedure}

%------------------------------------------------------------------------------
\subsection{ATLAS: Configuration and Hyperparameters}
\label{app:ATLAShyperparameters}
\input{Appendix/Hyperparameters}

\subsection{ATLAS: Candidate Validation, Repair, and Execution Pipeline}
\label{app:evaluation}

Each candidate algorithm generated by the LLM is evaluated through a structured pipeline that assesses executability, constraint feasibility, and solution quality. 

\subsubsection{Sequential LLM Calls, Parallel Instance Evaluation}
LLM synthesis calls are executed sequentially, while evaluation across problem instances is parallelized. Sequential LLM calls simplify rate-limit handling and long-run orchestration of the iterative search loop. Instance evaluations use a CPU-aware process pool whose worker count is the smallest of the instance count, available CPU cores, and the user-specified limit.

\subsubsection{Entry-Point Checking and Worker-Process Execution}
Each generated program is first checked for the required problem-specific entry-point function; programs failing this compatibility check are marked invalid without execution. Programs that pass are executed with normal Python built-ins and import access in separate worker processes. The process boundary, timeouts, and memory limits provide fault and resource containment, but they are not a security sandbox and do not block filesystem or network access. Generated code must therefore be treated as untrusted and evaluated in a disposable, least-privilege virtual machine or container without sensitive files, network access, or inherited credentials.

\subsubsection{Resource Limits and Error Handling}
A per-instance timeout $t_{\text{timeout}}$ is enforced during execution. Forward experiments run under Linux/WSL2 (native Windows execution is refused), where each evaluation worker receives a configured allowance for additional virtual-address-space allocation above its inherited process footprint through \texttt{RLIMIT\_AS}. These controls bound runaway execution and excessive allocation; the memory setting is a worker resource allowance rather than a measurement of an algorithm's peak resident memory.
Timeouts, memory-limit violations, runtime exceptions (e.g., \texttt{IndexError}, \texttt{TypeError}), or invalid outputs mark that instance as failed. For candidates eligible for repair, the operator receives failure counts, evaluator error summaries, and available traceback or resource flags; it does not receive the failing instance data itself.

\subsubsection{Independent Constraint and Objective Validation}
Objective values and constraint satisfaction are evaluated by the framework's authoritative validator. 
The generated algorithm returns only the raw solution representation (e.g., a route set or permutation); the evaluator then independently verifies:
\begin{itemize}[itemsep=1pt]
\item \textbf{Structural validity:} correct output format and dimensions
\item \textbf{Resource constraints:} capacity limits respected (e.g., CVRP)
\item \textbf{Temporal constraints:} time windows satisfied (e.g., CVRPTW)
\item \textbf{Problem-specific validity:} sequencing/order and feasibility conditions required by the target problem (e.g., FSS, QAP)
\end{itemize}

\subsubsection{All-or-Nothing Validity and Repair}
A candidate is considered \emph{valid} only if it succeeds on \emph{all} training instances without timeouts, exceptions, format violations, or constraint violations. If any instance fails, the candidate is tagged as failed and passed once to the \textsc{Repair} operator (\S\ref{subsec:operators}) with diagnostic feedback. The repaired candidate is re-evaluated; if it still fails, it is discarded. Only valid algorithms are added to the archive, and their fitness is computed as the mean objective value across instances.

\input{Appendix/RepairPrompt}

%------------------------------------------------------------------------------
\subsection{Baselines}
\label{app:baselines}
%------------------------------------------------------------------------------
\input{Appendix/Baselines}

\section{Component Ablations and Design Analyses}
\label{app:ablations}

We distinguish component ablations, which disable or replace ATLAS search mechanisms and evaluate final algorithm quality, from supporting design analyses, which study representation quality, initialization behavior, or the sensitivity of ATLAS to its LLM backend. The semantic search-guidance and search-layer studies are component ablations; the embedding, initialization, reasoning-effort, and underlying-LLM studies are design analyses.

\subsection{Embedding Representation Analysis Against Curated Family Labels}
\label{app:embedding_ablation}
\input{Appendix/ablation_embedding}

\subsection{Initialization Strategy Analysis: Validity and Embedding-Space Coverage}
\label{app:init_ablation}
\input{Appendix/ablation_initialization}

%==============================================================================
\subsection{Semantic Search-Guidance Ablation}
\label{app:clustering_ablation}
\input{Appendix/ablation_clustering}
%==============================================================================
\subsection{Search-Layer Ablation}
\label{app:layer_ablation}
\input{Appendix/ablation_layers}
%==============================================================================
\subsection{Reasoning-Effort Sensitivity and Cost Analysis}
\label{app:reasoning_ablation}
\input{Appendix/ablation_reasoning}
%==============================================================================
\subsection{Effect of the Underlying LLM on ATLAS}
\label{app:differentLLMs}

This subsection examines how the underlying LLM affects ATLAS when the framework itself is otherwise kept fixed. 
We compare four models on FSS under the default ATLAS and benchmark settings: \textsc{GPT-5-nano} (low), \textsc{GPT-5-mini} (low), \textsc{GPT-5} (low), and \textsc{Claude Haiku 4.5} with its default setting (that is, without extended thinking). 
The \textsc{GPT-5} family is evaluated at the same low reasoning level to keep the reasoning configuration consistent and better isolate differences due to model capability rather than reasoning budget. 
\textsc{Claude Haiku 4.5} is included to provide a comparison with a model from a different provider. 
We first compare final test performance using one training-locked algorithm per run.
We then inspect the training-selected archive representatives, using test gaps only for descriptive display and ordering.
This descriptive view does not contribute to the headline comparison or its statistical tests.

\subsubsection{Performance Comparison}
Table~\ref{tab:fss_llm_comparison_gap} shows that the underlying LLM affects ATLAS performance, although the stronger models remain relatively close.
\textsc{GPT-5} achieves the best mean performance, with \textsc{GPT-5-mini} close behind, while \textsc{GPT-5-nano} and \textsc{Claude Haiku 4.5} are weaker but still achieve acceptable performance. 
This comparison establishes a difference in final selected-algorithm quality, but does not isolate whether the cause is local code quality, proposal diversity, or longer-term refinement dynamics.

\subsubsection{Impact on Archive Diversity}
Table~\ref{tab:fss_llm_representatives_full} reports a training-selected representative set from the displayed final archive for each LLM. Representative identities are fixed from training objectives, and the test-set gaps are used only to order and interpret the already selected repertoires. The table is not used in the headline LLM comparison or its statistical tests. Within this descriptive view, \textsc{GPT-5-mini} and \textsc{GPT-5} retain strong and structurally varied sets of competitive representatives.
The inspected \textsc{GPT-5-mini} representatives include NEH-style construction, large-neighborhood search, genetic search, spectral hybrids, and tabu-guided refinement; \textsc{GPT-5} also includes backbone- and precedence-guided approaches. The displayed \textsc{Claude Haiku 4.5} representatives are more concentrated around simulated-annealing and memetic variants, while the displayed \textsc{GPT-5-nano} representatives have a wider test-gap spread. These training-selected examples illustrate the repertoire and hybridization capabilities of ATLAS, but do not establish comparative archive-diversity performance or a causal model effect on family discovery.

\begin{table}[t]
\centering
\caption{\emph{Relative mean gaps for the underlying-LLM comparison on FSS.} Percent gap to IG-TB, the strongest human-designed reference for this setting; lower is better. Values are computed from the underlying raw objective means. Boldface mirrors the conclusion of the paired raw-objective analysis and is not based on a test of the displayed gaps.}
\label{tab:fss_llm_comparison_gap}
\footnotesize
{
\begin{tabular}{lc}
\toprule
\textbf{Underlying LLM} & \textbf{FSS} $(n50m10)$ gap (\%)$\downarrow$ \\
\midrule
\textsc{GPT-5-nano} (low)   & 0.557 \\
\textsc{GPT-5-mini} (low)   & 0.347 \\
\textsc{GPT-5} (low)        & \textbf{0.284} \\
\textsc{Claude Haiku 4.5} (default) & 0.540 \\
\bottomrule
\end{tabular}}
\end{table}

\begin{table*}[!t]
\centering
\caption{\emph{Training-selected FSS archive representatives by LLM.} Representative identities are fixed from training objectives as the training-best members of final semantic clusters. Within each LLM, the displayed set is ordered by test-set relative mean gap only for interpretation. Values in parentheses are test-set relative mean gaps (\%) to IG-TB, the strongest human-designed FSS reference; lower is better. This descriptive ordering does not affect representative membership, the training-selected deployable algorithm, the headline LLM comparison, or statistical tests.}
\label{tab:fss_llm_representatives_full}
\footnotesize
\setlength{\tabcolsep}{4pt}
\begin{tabular}{@{}p{1.9cm}p{15.3cm}@{}}
\toprule
\textbf{LLM} & \textbf{Training-selected representatives, ordered by descriptive test gap} \\
\midrule

\textsc{GPT-5-nano (low)} &
Permutation Based Flow Shop Heuristic with Cascaded Insertion and Local Search (0.474\%);
Hybrid Seeded Cascaded Insertion with Global Lookahead (0.501\%);
Fixed Permutation Flow Shop Seeded Local Search (0.527\%);
Diversified Guided Cascaded Beam Search for Permutation Flow Shop (0.753\%);
Hybrid Prefix Guided Insertion with Global Lookahead and Prefix Preserving Local Search (1.333\%);
Genetic Algorithm based on permutation population with crossover and mutation (1.372\%);
Greedy NEH-inspired permutation for flow shop makespan minimization (2.809\%);
Recursive Skyline Block Sequencing (3.319\%);
Hybrid Pheromone Guided Time Slice Sequencing (3.383\%);
Pheromone Guided Permutation Flow Shop Scheduler (3.483\%) \\
\cmidrule(lr){2-2}

\textsc{GPT-5-mini (low)} &
Priority-Seeded Prefix-Guided LNS with Lightweight Anneal (0.331\%);
Priority-seeded NEH with Guided Neighborhood Search (0.458\%);
Priority-Spectral Hybrid with Tabu-Repair Local Search (0.472\%);
Prefix-Aware Hybrid Genetic with Annealed Polish (0.478\%);
Sampled Pairwise Greedy with Bounded Cache and Adaptive Iterated Refinement (0.527\%);
Hybrid NEH-Regret Genetic Search with Aggressive Complexity Reduction (0.714\%);
Guided Greedy Adaptive Cache with Pheromone and Block Relocation (0.965\%);
Bottleneck-Guided Greedy with ILS and Simulated Annealing (1.239\%) \\
\cmidrule(lr){2-2}

\textsc{GPT-5 (low)} &
Pairwise backbone scheduling with regret beam construction and consensus edge relinking (0.277\%);
Precedence corridor scheduling with shockwave rebuild and edge voted relinking (0.280\%);
Vote guided anchor beam scheduling with cached insertion scoring and focused reconstruction (0.304\%);
Multi-sequence NEH flow shop heuristic with CDS, Palmer, and elite reinsertion search (0.391\%);
Consensus rollout scheduling with hot segment repair and block guided relinking (0.454\%);
Pairwise precedence learning with exact insertion and conflict focused segment reconstruction (0.460\%);
Tournament backbone search with window dynamic programming (0.800\%);
Cross-entropy random-key optimization with block permutation polishing (2.746\%) \\
\cmidrule(lr){2-2}

\textsc{Claude Haiku 4.5} &
Hybrid Adaptive Simulated Annealing with Dynamic Critical Path Guidance (0.484\%);
Simulated Annealing with Critical Path Optimization for Flow Shop Scheduling (0.571\%);
Simulated Annealing with Greedy Initialization for Flow Shop Scheduling (0.579\%);
Hybrid Memetic Algorithm with Adaptive Dual-Strategy Exploration (0.809\%);
Hybrid Adaptive Metaheuristic with Critical Path Learning (1.016\%);
Hybrid Initialization and Local Search for Flow Shop Scheduling (1.237\%);
Hybrid Memetic Strategy with Dual-Phase Evolution and Adaptive Intensity Balancing (1.537\%);
Genetic Algorithm with Dual Crossover Operators and Adaptive Multi-Strategy Mutations (1.558\%);
Constraint-Based Graph Routing with Bottleneck-Aware Optimization (2.181\%)\\

\bottomrule
\end{tabular}
\end{table*}

\end{document}

%% file: Appendix/RelatedWork.tex
This section contextualizes ATLAS within the broader landscape of LLM-based automated algorithm design from two complementary perspectives.
First, we position ATLAS within recent taxonomic frameworks proposed in comprehensive surveys (Appendix~\ref{app:surveys}), highlighting its formulation of scaffold-free full-algorithm synthesis under a fixed I/O interface.
Second, we provide a technical overview of recent LLM-based methods (Appendix~\ref{app:sota_comparison}) and concentrate direct comparisons on designed-multiplicity and full-synthesis approaches, for which the relationship to ATLAS requires closer examination.

\subsection{Positioning ATLAS in Existing Taxonomies}
\label{app:surveys}
Recent systematic reviews have formalized the landscape of LLM-based optimization, providing useful taxonomic foundations for contextualizing ATLAS. 
We position ATLAS within three complementary frameworks, each offering a different perspective on automated algorithm design.

\subsubsection{LLM-as-Designer (Role-Based View)} \citet{liu2024systematic} categorize LLMs for Algorithm Design (LLM4AD) into four paradigms based on the LLM's role: LLM-as-Optimizer, LLM-as-Predictor, LLM-as-Extractor, and LLM-as-Designer (LLMaD). 
ATLAS belongs to LLMaD, where the model generates algorithmic code rather than directly proposing solutions. 
Critically, \citet{liu2024systematic} observe that most LLMaD methods struggle with ``synthesizing complete, state-of-the-art algorithms'' and instead target ``specific components (e.g., heuristics, reward functions, code snippets) rather than complete algorithms.'' 
ATLAS addresses this limitation in combinatorial optimization by synthesizing complete algorithms from a problem specification under a fixed I/O interface.

\subsubsection{High-Level Algorithm Generation (Intervention-Level View)} Complementing this role-based taxonomy, \citet{zhang2025systematic} organize the field by separating \textit{Optimization Modeling} from \textit{Optimization Solving}, with the latter further divided into three paradigms: 
\begin{inparaenum}
    \item LLMs as stand-alone optimizers,
    \item low-level LLM-assisted methods (embedded components), and
    \item high-level LLM-assisted methods (algorithm selection and generation).
\end{inparaenum}
ATLAS operates at the high-level generation tier. 
While algorithm selection methods ``choose the most suitable algorithm from a portfolio for each problem instance''~\cite{zhang2025systematic}, ATLAS performs \textit{open-ended synthesis}, maintaining a search repertoire of generated algorithms across multiple embedding regions rather than selecting from a fixed library.

\subsubsection{Generative MetaBBO (Mechanism $\times$ Task View)} At a finer level of granularity, \citet{ma2025toward} unify automated design under \textit{Meta-Black-Box Optimization (MetaBBO)}, categorizing methods along two dimensions: learning mechanism (Reinforcement Learning, Supervised Learning, Neuroevolution, or In-Context Learning) and meta-level task (Algorithm Selection, Configuration, Solution Manipulation, or Generation). 
ATLAS is a MetaBBO method for Algorithm Generation via In-Context Learning. 
The key distinction is that while many MetaBBO approaches perform \textit{parametric} meta-optimization, such as tuning hyperparameters or selecting operators within fixed templates, ATLAS performs \textit{structural} meta-design by synthesizing the complete algorithmic logic under a fixed external I/O interface.

\subsubsection*{Synthesis: Scaffold-Free Full-Algorithm Synthesis under a Fixed I/O Interface} Across these three taxonomic lenses, ATLAS lies at the intersection of LLMaD (role), high-level generation (intervention), and structural MetaBBO (mechanism). Appendix~\ref{app:sota_comparison} reviews selected recent methods through these taxonomic and technical dimensions.

\subsection{Technical Overview of Selected Recent Methods}
\label{app:sota_comparison}

Appendix~\ref{app:surveys} positioned ATLAS within the LLM-as-Designer paradigm for high-level algorithm generation. 
The methods reviewed here differ in synthesis scope (component-based vs.\ full-algorithm), internal structural prior, search organization, and the rule by which candidates persist.
Table~\ref{tab:atlas_comparison_summary} summarizes these distinctions, and the following subsections provide technical descriptions of the selected methods.

\begin{itemize}[leftmargin=*, itemsep=2pt]
\item \textbf{Single-Algorithm Component Methods} (Appendix~\ref{app:component_single}) iteratively optimize a designated component within a predefined or hierarchical scaffold.
\item \textbf{Designed-Multiplicity Component Methods} (Appendix~\ref{app:component_multi}) explicitly aim to produce multiple heuristics, prescribing diversity through multi-objective formulations or instance-coverage criteria.
\item \textbf{Full-Synthesis Methods} (Appendix~\ref{app:full_synthesis}) generate and optimize complete algorithms rather than individual components or functions.
\end{itemize}

\textbf{ATLAS Distinction:} ATLAS combines full-algorithm synthesis from a problem specification under a minimal I/O interface with no prescribed internal decomposition, then applies embedding-distance-aware archive maintenance to preserve coverage across search regions.
The following subsections describe representative methods in each category.
Direct comparisons with ATLAS are concentrated on designed-multiplicity and full-synthesis approaches, where differences in output purpose, synthesis scope, and structural assumptions require closer examination.

\begin{table}[H]
\centering
\small
\caption{\emph{Comparative analysis of LLM-based algorithm synthesis methods.}
\textbf{Scope} distinguishes component optimization from full-algorithm synthesis.
\textbf{Organizing space} indicates where diversity/selection pressure is applied.
\textbf{Preservation logic} states the operational rule by which candidates persist in the search state; these mechanisms are reported descriptively without imposing a hierarchy among them.}
\label{tab:atlas_comparison_summary}
\resizebox{\textwidth}{!}{%
\begin{threeparttable}
\setlength{\tabcolsep}{4pt}
\renewcommand{\arraystretch}{1.2}
\begin{tabular}{@{}l p{2cm} p{3.5cm} p{3.5cm} p{5cm}@{}}
\toprule
\textbf{Method} & \textbf{Scope} & \textbf{Primary Mechanism} & \textbf{Organizing Space} & \textbf{Preservation Logic} \\
\midrule
\multicolumn{5}{c}{\textit{Component-Based Single-Algorithm Synthesis (Appendix~\ref{app:component_single})}} \\
\midrule
\rowcolor{gray!12}
\textsc{FunSearch}~\cite{romera2024mathematical} & Component & Islands + best-shot prompting & Output signatures (eval scores) & Cluster-biased sampling + island resets \\
\textsc{EoH}~\cite{liu2024evolution} & Component & Operator-based evolution & Fitness (scalar performance) & Fitness-based selection; no explicit diversity objective \\
\rowcolor{gray!12}
\textsc{ReEvo}~\cite{ye2024reevo} & Component & Verbal gradients + reflection & Fitness + meta-objective constraints & Successful candidates under meta-objective constraints \\
\textsc{HSEvo}~\cite{dat2025hsevo} & Component & Harmony Search + reflection & Embeddings (diversity indices) + fitness & Embedding diversity indices (SWDI/CDI) modulate search \\
\rowcolor{gray!12}
\textsc{MCTS-AHD}~\cite{zheng2025monte} & Component & MCTS with LLM actions & Genealogy (tree) + UCT scores & Tree retention with UCT-based revisiting \\
\textsc{CALM}~\cite{huang2025calm} & Component & RL fine-tuning (GRPO) + pool evolution & Fitness rewards (policy updates) & Pool with stagnation-triggered collapse; diversity not explicit \\
\rowcolor{gray!12}
\textsc{HeurAgenix}~\cite{yang2025heuragenix} & Component\tnote{1} & Contrastive heuristic evolution + online selection & Fixed state--operator interface + heuristic pool & Interchangeable heuristic pool under a predefined $H:\mathcal{Z}\!\rightarrow\!\mathcal{O}$ interface\\
\textsc{MTHS}~\cite{liu2026mths} & Hierarchical Component\tnote{2} & Two-level multi-task evolution + cross-task transfer & Task-score vectors + task-specific program populations & Task champions + Pareto fronts; per-task best-program retention \\
\midrule
\multicolumn{5}{c}{\textit{Component-Based Designed-Multiplicity Synthesis (Appendix~\ref{app:component_multi})}} \\
\midrule
\rowcolor{gray!12}
\textsc{MEoH}~\cite{yao2025multi} & Component & NSGA-II + dominance-based prompting & Objective space (quality vs. runtime) & Non-dominated sorting + crowding distance \\
\textsc{EoH-S}~\cite{liu2025eoh} & Component & Portfolio optimization (CPI) & Instance-wise performance vectors & Greedy marginal CPI + instance-space niche coverage \\
\midrule
\multicolumn{5}{c}{\textit{Full-Synthesis (Appendix~\ref{app:full_synthesis})}} \\
\midrule
\rowcolor{gray!12}
\textsc{LLaMEA}~\cite{vanstein2025llamea} & Full Metaheuristic\tnote{3} & $(1{+}1)/(1{,}1)$ full-algorithm evolution & Scalar BBOB performance~\cite{finck2009bbob} + execution feedback & Best/latest-parent retention \\
\textsc{AlphaEvolve}~\cite{novikov2025alphaevolve} & Scaffolded Full Algorithm\tnote{4} & Patch-based evolution over annotated blocks & Program variants (MAP-Elites-inspired archive) & Archive retention with evolutionary sampling \\
\rowcolor{gray!12}
\textsc{A2DEPT}~\cite{chen2026a2dept} & Structured Full Algorithm\tnote{5} & Program-lineage tree + hierarchical edits & Genealogy, scalar fitness, and operator feedback & SA acceptance + diversity-aware history/partner sampling \\
\textbf{ATLAS (Ours)} & \textbf{Scaffold-Free Full Algorithm\tnote{6}} & \textbf{Full-algorithm operators + semantic quality-diversity archiving} & \textbf{Embedding space (semantic structure) + fitness} & \textbf{Embedding-distance-aware pruning preserves coverage across archive regions} \\
\bottomrule
\end{tabular}%

\begin{tablenotes}[flushleft]
\footnotesize
\item[1] \textsc{HeurAgenix} evolves heuristics within a fixed hyper-heuristic scaffold with predefined state-operator interface.
\item[2] \textsc{MTHS} instantiates a template-structured, task-agnostic metaheuristic as complete task-specific programs, then identifies and evolves one key function per task; complete programs support execution and transfer, but task-specific refinement is component-level.
\item[3] The published \textsc{LLaMEA} instantiation generates complete optimizer classes under a fixed callable interface for box-constrained continuous BBOB~\cite{finck2009bbob}.
\item[4] {\footnotesize
\textsc{AlphaEvolve} evolves programs by applying patches within user-marked evolution blocks; code outside blocks constrains interfaces and execution flow.}
\item[5] \textsc{A2DEPT} represents a complete algorithm as a fixed preface plus a role-partitioned function registry; its macro operator can rewrite the entry-point workflow and introduce helper functions.
\item[6] ATLAS receives the objective, constraints, and valid-solution requirements as the problem specification; its algorithm-facing interface fixes the callable entry point and instance/solution I/O formats without prescribing the algorithm's internal decomposition or control flow.
\end{tablenotes}
\end{threeparttable}}
\end{table}

\subsubsection{Component-Based Single-Algorithm Synthesis} 
\label{app:component_single}

This category encompasses methods whose iterative search targets a designated heuristic component within a predefined algorithmic framework or structured hierarchy.
Prominent examples include \textsc{FunSearch}~\cite{romera2024mathematical}, \textsc{EoH}~\cite{liu2024evolution}, \textsc{ReEvo}~\cite{ye2024reevo}, \textsc{MCTS-AHD}~\cite{zheng2025monte}, \textsc{CALM}~\cite{huang2025calm}, \textsc{HeurAgenix}~\cite{yang2025heuragenix}, and \textsc{MTHS}~\cite{liu2026mths}.
While these approaches use different search mechanisms, they retain a prescribed scaffold or component-level refinement target, such as a priority function, scoring rule, operator, or identified key function.

\paragraph{\textsc{FunSearch}} 
\textsc{FunSearch}~\cite{romera2024mathematical} is an evaluator-guided program discovery framework that uses a pretrained code LLM to iteratively improve a target function within a user-provided program. 
The user provides an \texttt{evaluate} routine (the scoring oracle) together with an initial implementation; in practice, \textsc{FunSearch} is most naturally applied when the program serves as a \emph{scaffold} and the search is focused on a critical function (e.g., a priority rule in a greedy algorithm), allowing the LLM to optimize the key design choice within a fixed algorithmic structure. 
\textit{Best-shot prompting} samples $k$ programs from a population, inserts them as versioned functions (e.g., \texttt{priority\_v0}, \texttt{priority\_v1}), and appends an empty \texttt{priority\_v2} header for the LLM to complete, encouraging reuse and recombination of ideas across sampled candidates.

To sustain exploration, \textsc{FunSearch} maintains a programs database evolved with an \emph{islands model}, in which sub-populations evolve independently and are periodically reset by discarding the worst-performing half of islands and reseeding them from the best programs of surviving islands. 
Within each island, candidates are clustered by a \emph{functional signature}, defined as the tuple of scores over the evaluation inputs, and sampling proceeds by first choosing a signature cluster (biased toward higher score) and then selecting shorter programs within that cluster. 
This yields a form of behavioral diversity, but the induced partition remains tied to observed outputs on a finite evaluation set.

\paragraph{Evolution of Heuristics (\textsc{EoH})}
\textsc{EoH}~\cite{liu2024evolution} introduces an LLM-guided evolutionary framework for heuristic design in which each candidate is represented by both a concise natural-language \emph{thought}, describing the core algorithmic idea, and an executable code implementation (typically a Python function) that realizes it. 
Candidates are evaluated by executing their code on training instances to obtain a scalar fitness, and new candidates are generated through five LLM-driven operators. 
Three operators refine individual heuristics through different mutation-style transformations, while two exploration operators either combine information from multiple parents or generate more novel variants. 
Population updates are then driven by fitness-based selection, such as tournament or rank-based selection, which retains stronger candidates and discards weaker ones.

\paragraph{Reflective Evolution (\textsc{ReEvo})} 
\textsc{ReEvo}~\cite{ye2024reevo} shifts evolutionary search from purely stochastic mutation toward reasoning-guided optimization by introducing \textit{verbal gradients}, namely natural-language critiques that explain why one heuristic outperforms another. 
The system employs dual LLM roles: a \textit{generator} produces candidate heuristics as executable code, and a \textit{reflector} performs comparative analysis through short-term reflection (pairwise comparison guiding crossover) and long-term reflection (persistent memory of design patterns guiding mutation). 
To maintain exploration, \textsc{ReEvo} selects parent pairs randomly from successful candidates with an explicit constraint to avoid identical meta-objective values, encouraging behavioral diversity and reducing premature convergence toward near-duplicate heuristics.
\textsc{ReEvo}'s reflection mechanism provides explicit directional guidance for code improvement: the LLM receives textual critiques of why particular design choices succeed or fail, which can accelerate local iterative refinement within a fixed scaffold.

\paragraph{Harmony Search Evolution (\textsc{HSEvo})}
\textsc{HSEvo}~\cite{dat2025hsevo} addresses the exploration-exploitation trade-off in LLM-based heuristic synthesis by treating diversity as an explicit optimization concern. 
Motivated by the empirical observation that earlier methods may preserve higher diversity at the expense of objective performance (e.g., \textsc{EoH}) or achieve stronger objective performance with lower diversity (e.g., \textsc{ReEvo}, \textsc{FunSearch}), \textsc{HSEvo} proposes two embedding-based diversity measures: the Shannon-Wiener Diversity Index (SWDI) and Cumulative Diversity Index (CDI). 
It adapts Harmony Search by maintaining a Harmony Memory updated via \textit{Memory Consideration} (reuse of existing patterns), \textit{Pitch Adjustment} (LLM-guided local variation), and \textit{Random Selection} (novel proposals). 
\textsc{HSEvo} further augments the reflection-based pipeline with \textit{Flash Reflection}, a batch ranking and distillation procedure, and an HS-based individual tuning stage that extracts tunable parameters (e.g., thresholds, weights) from top individuals, optimizes them via Harmony Search, and reinserts the improved heuristics into the population.

\paragraph{Monte Carlo Tree Search for Automatic Heuristic Design (MCTS-AHD)} 
\textsc{MCTS-AHD}~\cite{zheng2025monte} replaces population-based evolution with Monte Carlo Tree Search to avoid prematurely discarding temporarily underperforming heuristics. 
Motivated by the concern that population-based retention rules may eliminate candidates that do not immediately outperform the current worst individual, thereby blocking worse-before-better search trajectories, MCTS-AHD preserves LLM-generated heuristic functions as nodes in an MCTS tree, each associated with executable code and a linguistic description, and iteratively applies selection, expansion, simulation, and backpropagation. 
The core premise is that weak individuals may still serve as useful stepping stones toward higher-quality solutions. 
Expansion employs LLM-driven actions analogous to evolutionary operators (initialization, mutation, crossover), together with a distinctive \textit{Tree-Path Reasoning} action in which the LLM analyzes the full evolutionary lineage from the root to the current leaf and synthesizes insights from that history of design modifications. 
To reduce code-description mismatch, the method uses \textit{thought-alignment}, generating descriptions after code generation rather than before. 
To balance exploration and convergence, it employs \textit{Progressive Widening} to control the branching factor dynamically and an \textit{Exploration-Decay} mechanism that linearly reduces the exploration weight over time.

\paragraph{Co-evolution of Algorithms and Language Model (\textsc{CALM})} 
\textsc{CALM}~\cite{huang2025calm} extends LLM-based automatic heuristic design beyond prompt-level evolution by jointly evolving the heuristic pool and the LLM via reinforcement learning. 
Specifically, CALM integrates \emph{verbal guidance} (operator-driven evolution over a pool of heuristics) with \emph{numerical guidance} by fine-tuning the LLM using Group Relative Policy Optimization (GRPO)~\cite{shao2024deepseekmath} from heuristic-performance rewards. 
The framework maintains a pool of heuristics, each storing an idea, executable code, and performance; at each round it samples an evolutionary operator (injection, replacement, crossover, simplification, or initialization), generates multiple candidate responses, evaluates them, and uses the resulting prompt-response-performance tuples to update the LLM. 
For most operators, parent heuristics are sampled by performance rank with probability inversely proportional to rank (and heuristics beyond the pool cutoff receive zero probability), and CALM includes a stagnation-triggered \emph{collapse mechanism} that resets the pool by discarding all but the original seed and the current best heuristic to re-initiate exploration.

\paragraph{\textsc{HeurAgenix}} 
\textsc{HeurAgenix}~\cite{yang2025heuragenix} proposes an LLM-based hyper-heuristic framework with two coupled stages: 
\begin{inparaenum}
    \item a contrastive, data-driven heuristic evolution phase that discovers reusable evolution strategies from contrastive solution trajectories, and
    \item an adaptive problem-solving phase that selects among the evolved heuristic pool online, using either a frontier LLM or a lightweight fine-tuned selector model. 
\end{inparaenum}
The framework is positioned as evolving and deploying heuristics without relying on an external framework, while enabling instance-level adaptation through dynamic heuristic switching.

Despite this end-to-end framing, \textsc{HeurAgenix} remains component-based from the perspective of synthesis scope because the generated artifacts are constrained to a fixed hyper-heuristic scaffold: 
a heuristic is explicitly defined as a function mapping a structured \emph{problem state} to an allowable \emph{operation} ($H:\mathcal{Z}\!\rightarrow\!\mathcal{O}$), which is then executed by a predefined transition function ($T:\mathcal{Z}\!\times\!\mathcal{O}\!\rightarrow\!\mathcal{Z}$). 
All heuristics in the pool must satisfy a common callable interface and produce operators from a predefined operator family to support runtime switching.

\paragraph{Multi-Task Hierarchical Search (\textsc{MTHS})}
\textsc{MTHS}~\cite{liu2026mths} addresses cross-task automated heuristic design through an explicitly scaffolded two-level representation.
Each high-level individual contains a task-agnostic metaheuristic expressed through a structured template, with components such as initialization, the main search loop, and post-processing.
High-level evolution revises this structured metaheuristic description rather than an unrestricted task-specific source artifact.
For each task, an LLM converts that representation into a complete executable implementation under a supplied program template, identifies one performance-critical key function, and conducts low-level evolution by replacing that function while preserving its interface and the surrounding program.

A cross-task transfer stage supplies a successful source-task program together with the target task's template to generate an adapted target program.
At the high level, each individual is evaluated by a vector of task-specific scores; population management retains task champions and then applies Pareto-front survival, while the low-level search retains the best task-specific program associated with each metaheuristic.
Complete programs are therefore generated for evaluation and transfer, but the task-specific evolutionary refinement remains component-level and the high-level representation prescribes an algorithmic structure.
We consequently classify \textsc{MTHS} as scaffolded hierarchical synthesis with key-function refinement, rather than full-algorithm synthesis.

\subsubsection{Component-Based Multi-Algorithm Synthesis} 
\label{app:component_multi}

\paragraph{Multi-objective Evolution of Heuristics (\textsc{MEoH})} 
\textsc{MEoH}~\cite{yao2025multi} reformulates automatic heuristic design as a multi-objective optimization problem to address the practical trade-off between solution quality and computational complexity. 
Arguing that standard methods (e.g., \textsc{EoH}) may neglect runtime efficiency in pursuit of marginal performance gains, \textsc{MEoH} embeds an LLM within an NSGA-II-based evolutionary framework~\cite{deb2002fast} to approximate the Pareto front of heuristic designs. 
The framework employs a specialized \textit{dominance-based prompting} strategy: the LLM is presented with parent heuristics together with their objective vectors (e.g., optimality gap and runtime) and is instructed to generate offspring that improve the trade-off between the competing objectives. 
By applying non-dominated sorting and crowding-distance selection, \textsc{MEoH} maintains a diverse population of heuristics, ultimately producing a spectrum ranging from fast, efficient variants to more computationally intensive, high-performance alternatives.

\paragraph{Evolution of Heuristic Set (\textsc{EoH-S})} 
Observing that a single heuristic may generalize poorly across instance distributions and scales, \textsc{EoH-S}~\cite{liu2025eoh} reformulates LLM-based design as Automated Heuristic Set Design: learning a small portfolio that minimizes the Complementary Performance Index (CPI), defined as the average across instances of the best-performing heuristic in the set. 
Leveraging CPI's monotone-supermodular structure, \textsc{EoH-S} introduces Complementary Population Management, which greedily selects heuristics by marginal CPI gains, together with a diversity-aware memetic search that chooses parents with maximal Manhattan distance in instance-wise performance space so as to synthesize heuristics covering different niches.

\textsc{MEoH} and \textsc{EoH-S} both target multiplicity as an explicit output objective: respectively, a Pareto front over quality and runtime, or a complementary portfolio over instance regimes. ATLAS does not optimize either of those output criteria. Instead, it explicitly prescribes embedding-space coverage as a search-state preservation rule. The identities and algorithmic mechanisms of the retained candidates emerge during synthesis, while similarity-based pruning, neighborhood retrieval, and cross-region selection determine how that discovered repertoire is maintained and revisited.

\subsubsection{Full-Synthesis} 
\label{app:full_synthesis}

Full-synthesis approaches remain relatively underexplored and instantiate the setting with
different domains, external interfaces, and internal structural priors.
\textsc{LLaMEA}~\cite{vanstein2025llamea} evolves complete metaheuristics under a fixed continuous
black-box optimizer interface; \textsc{AlphaEvolve}~\cite{novikov2025alphaevolve} edits marked
regions of user-provided programs; and the concurrent \textsc{A2DEPT}~\cite{chen2026a2dept} evolves
complete algorithms through an explicit function-level representation and hierarchical edit
operators. ATLAS combines a problem-specific combinatorial I/O interface with internally
undecomposed full-algorithm operators and embedding-space coverage preservation.

\paragraph{\textsc{LLaMEA}}
\textsc{LLaMEA}~\cite{vanstein2025llamea} is a generate--evaluate--refine framework whose
published instantiation synthesizes complete Python metaheuristics for box-constrained continuous
black-box optimization. Its outer search uses either elitist $(1{+}1)$ selection, which returns the
best-so-far algorithm to the LLM, or non-elitist $(1{,}1)$ selection, which returns the latest
algorithm. In both cases the refinement prompt contains the selected algorithm's complete code,
aggregate performance, variability, and execution errors, and asks the LLM to refine or redesign it.
The prompt also lists the names and mean performance scores of previously generated algorithms,
providing a compact summary of the explored search history; however, only the selected algorithm is
supplied in full as the parent for refinement. The LLM therefore chooses the effective edit scale
and can change parameters, operators, and higher-level interactions inside the complete algorithm.

The framework is generic, but the published study focuses on the BBOB functions~\cite{finck2009bbob} in a homogeneous real-vector search space
with box constraints, a fixed function-evaluation budget, and a prescribed callable optimizer interface.
Its $(1{+}1)$ and $(1{,}1)$ search schemes maintain a single active evolutionary trajectory,
with either the best-so-far or the most recent algorithm serving as the complete parent design.
Although the names and mean scores of earlier algorithms provide lightweight historical context,
their complete designs are not retained as simultaneous parents for continued refinement or recombination.
This organization is simple, but it does not simultaneously preserve and refine multiple competing algorithm families.
That limitation can become important in broad full-algorithm design spaces containing alternative component inventories,
multiple interacting components, and opportunities for hybridization, because a promising but temporarily inferior design
may be discarded before it can be further refined or combined with other designs.
\textsc{LLaMEA} therefore establishes the feasibility of full-algorithm LLM evolution in its demonstrated setting,
while leaving open how population- or archive-based search could improve coverage and refinement in more heterogeneous design spaces.

\paragraph{\textsc{AlphaEvolve}}
\textsc{AlphaEvolve}~\cite{novikov2025alphaevolve} is an evolutionary coding agent for scientific and algorithmic discovery that operates over complete code artifacts, potentially spanning multiple functions or components, using one or more automated evaluators that return scalar feedback signals.
The user provides an initial codebase and explicitly marks editable regions with evolution-block annotations, while code outside those regions remains fixed and provides the surrounding execution structure. 
Candidate updates are generated by an ensemble of \textsc{Gemini} models and are applied as structured code edits before evaluation. 
The public description also highlights support for expensive objectives through staged evaluation cascades, optional LLM-based auxiliary judging, and large-scale parallel evaluation.

A key positioning claim of \textsc{AlphaEvolve} is that it extends beyond single-function program search by enabling coordinated evolution over larger code regions and richer code contexts. 
At the same time, its notion of full synthesis is most accurately understood as \emph{block-wise evolution of a supplied code artifact}: the search space is still determined by the user-provided codebase, the selected editable regions, and the surrounding fixed interfaces and execution flow.
In this sense, \textsc{AlphaEvolve} goes beyond component-level function evolution, but it does not operate in a scaffold-free setting.

\textsc{AlphaEvolve} was initially introduced through a public white paper and is now
offered as a Google Cloud product~\cite{googlecloud2026alphaevolve}; its underlying implementation
remains closed source.

\paragraph{\textsc{A2DEPT}}
\textsc{A2DEPT}~\cite{chen2026a2dept} was developed concurrently with ATLAS and is
therefore treated here as parallel work rather than prior work. It targets the same broad setting of
complete executable algorithm synthesis for combinatorial optimization. Its ``program tree'' is a search genealogy
whose nodes are complete algorithm variants, rather than a syntax tree of components. Within each
node, however, \textsc{A2DEPT} imposes an explicit function-level representation: a fixed preface is
paired with a function registry that an LLM partitions into immutable definitions and mutable
strategies. Micro-tuning edits one mutable strategy while preserving its interface; macro-mutation
rewrites the entry-point workflow and may introduce new helper functions; semantic crossover
combines parent strategies; and call-graph analysis repairs missing dependencies and prunes
unreachable code.

Thus, \textsc{A2DEPT} is not component synthesis, nor does it fix the number of
components: it is full-algorithm synthesis with a strong function-level inductive bias and explicit
edit granularities. In this respect it shares with \textsc{AlphaEvolve} the use of additional
structure to guide full-algorithm editing, although the structures differ: \textsc{AlphaEvolve}
relies on user-marked editable blocks in a supplied codebase, whereas \textsc{A2DEPT} generates
complete initial algorithms and can rewrite their entry-point workflows. ATLAS takes the
complementary design choice of supplying complete algorithm source directly to general operators,
without a mutable/immutable role partition or a prescribed function hierarchy, and delegates
identification and coordinated revision of internal mechanisms to the LLM.

ATLAS's released illustrative algorithms provide qualitative evidence that this
undecomposed full-algorithm representation can yield coordinated multi-function algorithms: for
example, the CVRP algorithm connects alternative construction procedures, savings merging, regret
repair, relocation, 2-opt, large-neighborhood search, and simulated-annealing acceptance in one
executable workflow. This demonstrates that explicit component labels are not required to realize
multi-component designs in the studied setting; it does not establish that explicit structural
guidance cannot improve sample efficiency or reliability elsewhere. Similarly, ATLAS could add a
prompt policy that focuses an edit on a selected function or on cross-function interactions without
changing its archive, evaluator, or three-layer search, but the current work does not evaluate that
specialized policy. No public \textsc{A2DEPT} implementation was available at the time of writing,
so we restrict this comparison to the methodological level and do not include an empirical comparison.

%% file: Appendix/Problems.tex
This section provides formal mathematical formulations for the four benchmark problems. 

%------------------------------------------------------------------------------
\subsubsection{Permutation Flow Shop Scheduling (FSS)}
\label{app:fss_definition}
%------------------------------------------------------------------------------

\textbf{Verbal definition:}
In the permutation flow shop scheduling problem, $n$ jobs must be processed on $m$ machines 
in a fixed sequential order. All jobs visit machines in the same sequence: Machine 1 
$\rightarrow$ Machine 2 $\rightarrow$ \ldots $\rightarrow$ Machine $m$. The goal is to find 
a job permutation that minimizes the makespan (total completion time).

\textbf{Notation:}
Let $\mathcal{J} = \{1, \ldots, n\}$ be the set of jobs and $\mathcal{M} = \{1, \ldots, m\}$ 
be the set of machines in processing order. Let $p_{j,k} \geq 0$ denote the processing time 
of job $j \in \mathcal{J}$ on machine $k \in \mathcal{M}$.

\textbf{Decision variable:}
A permutation $\pi = (\pi_1, \pi_2, \ldots, \pi_n)$ of jobs, where $\pi_\ell \in \mathcal{J}$ 
denotes the job scheduled in position $\ell$. This ordering applies to all machines.

\textbf{Completion times:}
Let $C_{\ell,k}$ denote the completion time of the job in position $\ell$ on machine $k$. 
With boundary conditions $C_{0,k} = 0$ for all $k$ and $C_{\ell,0} = 0$ for all $\ell$, 
completion times are computed recursively:
\begin{equation}
C_{\ell,k} = \max(C_{\ell-1,k}, C_{\ell,k-1}) + p_{\pi_\ell, k}
\quad \text{for } \ell = 1, \ldots, n, \; k = 1, \ldots, m.
\label{eq:fss_completion}
\end{equation}

\textbf{Objective function:}
Minimize the makespan:
\begin{equation}
\min_{\pi} \; C_{\max}(\pi) = C_{n,m}.
\label{eq:fss_objective}
\end{equation}

\textbf{Constraints:}
\begin{itemize}[itemsep=1pt]
\item \textbf{Permutation:} $\pi$ is a valid permutation of $\mathcal{J}$.
\item \textbf{Machine sequence:} All jobs visit machines in order $1 \to 2 \to \cdots \to m$.
\item \textbf{Machine capacity:} Each machine processes at most one job at a time (implicitly enforced by the sequential structure of $\pi$ and Eq.~\eqref{eq:fss_completion}).
\item \textbf{Non-preemption:} Operations cannot be interrupted once started.
\item \textbf{Precedence:} Job $\pi_\ell$ cannot start on machine $k+1$ until it completes on machine $k$ (enforced by Eq.~\eqref{eq:fss_completion}).
\end{itemize}

\textbf{Implementation note:}
In the released prompt implementation, jobs and machines are
0-indexed (0 to $n-1$ and 0 to $m-1$), matching Python conventions. The mathematical 
formulation above uses 1-indexing for notational clarity; the two representations are equivalent.

%------------------------------------------------------------------------------
\subsubsection{Capacitated Vehicle Routing Problem (CVRP)}
\label{app:cvrp_definition}
%------------------------------------------------------------------------------

\textbf{Verbal definition:}
A fleet of homogeneous vehicles with capacity $Q$ must serve $n$ customers from a central 
depot. Each customer has a demand and geographic coordinates. All vehicles start and end 
at the depot. The goal is to construct routes that visit all customers while respecting 
vehicle capacity constraints and minimizing total distance traveled.

\textbf{Notation:}
Let $V = \{1, \ldots, n\}$ be the set of customers and node 0 denote the depot. 
Let $p_i = (x_i, y_i) \in \mathbb{R}^2$ be the coordinates of node $i \in \{0\} \cup V$. 
Define Euclidean distance:
\begin{equation}
c_{ij} = \|p_i - p_j\|_2 = \sqrt{(x_i - x_j)^2 + (y_i - y_j)^2}.
\label{eq:euclidean_distance}
\end{equation}
Each customer $i \in V$ has demand $d_i > 0$, and vehicle capacity is $Q$.

\textbf{Decision variables:}
A solution is a set of routes $\mathcal{R} = \{r^{(1)}, \ldots, r^{(K)}\}$ where each 
route $r^{(k)} = (v_1^{(k)}, \ldots, v_{m_k}^{(k)})$ is an ordered sequence of distinct 
customers with $v_\ell^{(k)} \in V$. The number of routes $K$ is not fixed a priori 
(unlimited fleet).

\textbf{Objective function:}
The cost of route $r = (v_1, \ldots, v_m)$ is:
\begin{equation}
\text{cost}(r) = c_{0,v_1} + \sum_{\ell=1}^{m-1} c_{v_\ell, v_{\ell+1}} + c_{v_m, 0}.
\label{eq:cvrp_route_cost}
\end{equation}
Minimize total distance:
\begin{equation}
\min_{\mathcal{R}} \sum_{r \in \mathcal{R}} \text{cost}(r).
\label{eq:cvrp_objective}
\end{equation}

\textbf{Constraints:}
For a route $r$, let $\mathrm{cust}(r)$ denote the set of customers appearing in $r$.
\begin{align}
&\textbf{(Coverage)} && \bigcup_{r \in \mathcal{R}} \mathrm{cust}(r) = V, \quad 
\mathrm{cust}(r) \cap \mathrm{cust}(r') = \emptyset \; \forall r \neq r' \label{eq:cvrp_coverage}\\
&\textbf{(Capacity)} && \sum_{v \in \mathrm{cust}(r)} d_v \leq Q \quad \forall r \in \mathcal{R} \label{eq:cvrp_capacity}
\end{align}

\textbf{Implementation note:}
In the released prompt implementation, customers are 0-indexed
(0 to $n-1$), and the depot is not included in route representations; it is implicit 
that all routes start and end at the depot. This formulation instead uses customer indices 
$1$ to $n$ together with an explicit depot node 0 for mathematical clarity.

%------------------------------------------------------------------------------
\subsubsection{CVRP with Time Windows (CVRPTW)}
\label{app:cvrptw_definition}
%------------------------------------------------------------------------------

\textbf{Verbal definition:}
CVRPTW extends CVRP with temporal constraints. Each customer has a time window 
$[e_i, l_i]$ during which service must begin and a service time $s_i$. The depot 
has operating hours $[e_0, l_0]$. Vehicles travel at unit speed (travel time equals 
Euclidean distance). Vehicles may wait if arriving early, but starting service after 
a time window closes renders the route infeasible. All routes must return to the 
depot before it closes.

\textbf{Notation:}
In addition to CVRP notation, each customer $i \in V$ has service time $s_i \geq 0$ 
and time window $[e_i, l_i]$. The depot (indexed as node 0) has time window 
$[e_0, l_0]$ and zero service time ($s_0 = 0$), so depot feasibility is checked 
on return arrival time. Travel time equals distance: $t_{ij} = c_{ij}$.

\textbf{Decision variables:}
Same as CVRP: a set of routes $\mathcal{R} = \{r^{(1)}, \ldots, r^{(K)}\}$, 
where each route $r^{(k)} = (v_1^{(k)}, \ldots, v_{m_k}^{(k)})$ is an ordered 
sequence of customer nodes $v_\ell^{(k)} \in V = \{1, \ldots, n\}$.

\textbf{Timing dynamics:}
For route $r = (v_1, \ldots, v_m)$, let $\tau_i$ denote the arrival time at node $i$ 
and $\sigma_i$ denote the service start time. Vehicles depart the depot at time 
$\tau_0^{\text{dep}} = 0$ (since $e_0 = 0$ in our instances). The dynamics are:
\begin{align}
\tau_{v_1} &= \tau_0^{\text{dep}} + t_{0,v_1}, \quad 
\sigma_{v_1} = \max(\tau_{v_1}, e_{v_1}) \label{eq:cvrptw_first}\\
\tau_{v_\ell} &= \sigma_{v_{\ell-1}} + s_{v_{\ell-1}} + t_{v_{\ell-1}, v_\ell}, \quad 
\sigma_{v_\ell} = \max(\tau_{v_\ell}, e_{v_\ell}) \; \text{for } \ell = 2, \ldots, m \label{eq:cvrptw_recursion}
\end{align}
Depot return time:
\begin{equation}
\tau_0^{\text{return}} = \sigma_{v_m} + s_{v_m} + t_{v_m, 0}.
\label{eq:cvrptw_return}
\end{equation}

\textbf{Objective function:}
Minimize total distance (same as Eq.~\eqref{eq:cvrp_objective}):
\begin{equation}
\min_{\mathcal{R}} \sum_{r \in \mathcal{R}} \text{cost}(r).
\label{eq:cvrptw_objective}
\end{equation}

\textbf{Constraints:}
For a route $r$, let $\mathrm{cust}(r)$ denote the set of customers appearing in $r$.
\begin{align}
&\textbf{(Coverage)} && \text{Same as Eq.~\eqref{eq:cvrp_coverage}} \nonumber\\
&\textbf{(Capacity)} && \text{Same as Eq.~\eqref{eq:cvrp_capacity}} \nonumber\\
&\textbf{(Customer time windows)} && e_i \leq \sigma_i \leq l_i \quad \forall r \in \mathcal{R}, \; \forall i \in \mathrm{cust}(r) \label{eq:cvrptw_windows}\\
&\textbf{(Depot return)} && \tau_0^{\text{return}} \leq l_0 \quad \forall r \in \mathcal{R} \label{eq:cvrptw_depot}
\end{align}

\textbf{Important notes:}
\begin{itemize}[itemsep=1pt]
\item Time windows are \textbf{inclusive}: $[e_i, l_i]$ means service start at exactly $e_i$ or $l_i$ is acceptable.
\item The time-window constraint applies to \textbf{service start} $\sigma_i$. Service may extend beyond $l_i$ as long as $\sigma_i \leq l_i$.
\item Waiting is allowed: if $\tau_i < e_i$, then service starts at $\sigma_i = e_i$.
\item Infeasibility occurs when $\sigma_i > l_i$.
\item \textbf{Index mapping:} Here the depot is node 0 and customers are nodes $\{1, \ldots, n\}$.
In the released prompt implementation, customers are indexed in \texttt{customer\_coords}
as
$\{0, \ldots, n-1\}$ and routes contain only these customer-list indices;
the depot is represented separately via \texttt{depot\_coords} and \texttt{depot\_time\_window}.
\end{itemize}

%------------------------------------------------------------------------------
\subsubsection{Quadratic Assignment Problem (QAP)}
\label{app:qap_definition}
%------------------------------------------------------------------------------

\textbf{Verbal definition:}
Assign $n$ facilities to $n$ locations to minimize total interaction cost. The cost 
depends quadratically on the assignment: if facility $i$ is assigned to location $p$ and 
facility $j$ is assigned to location $q$, their contribution to the total cost is 
$f_{ij} \cdot d_{pq}$, where $f_{ij}$ is the flow between facilities and $d_{pq}$ 
is the distance between locations.

\textbf{Notation:}
Let $\mathcal{F} = \{1, \ldots, n\}$ be the set of facilities and 
$\mathcal{L} = \{1, \ldots, n\}$ be the set of locations. Let $F = [f_{ij}]$ be 
the $n \times n$ flow matrix (symmetric with $f_{ii} = 0$) and $D = [d_{pq}]$ be 
the $n \times n$ distance matrix (symmetric with $d_{pp} = 0$).

\textbf{Decision variable:}
A permutation $\pi: \mathcal{F} \to \mathcal{L}$, where $\pi(i)$ denotes the location 
assigned to facility $i$.

\textbf{Objective function:}
Minimize the total quadratic cost:
\begin{equation}
\min_{\pi} \sum_{i=1}^{n} \sum_{j=1}^{n} f_{ij} \cdot d_{\pi(i), \pi(j)}.
\label{eq:qap_objective}
\end{equation}

Equivalently, exploiting symmetry:
\begin{equation}
\min_{\pi} \sum_{i=1}^{n} \sum_{j=i+1}^{n} 2 \cdot f_{ij} \cdot d_{\pi(i), \pi(j)}.
\label{eq:qap_objective_symmetric}
\end{equation}

\textbf{Constraints:}
\begin{equation}
\pi \text{ is a bijection from } \mathcal{F} \text{ to } \mathcal{L}.
\label{eq:qap_constraint}
\end{equation}

\textbf{Implementation note:}
In the released prompt implementation, facilities and locations are 0-indexed
(0 to $n-1$). The assignment $\pi[i] = p$ means facility $i$ is placed at location $p$. 
The evaluator computes cost by summing over all pairs as in Eq.~\eqref{eq:qap_objective}. 
Since $f_{ii} = 0$, diagonal terms ($i = j$) contribute zero to the sum.

%% file: Appendix/InstanceGeneration.tex
We generate synthetic instances for each problem using benchmark generators based on the \textsc{LLM4AD} framework~\cite{liu2024llm4ad}. 
For CVRP, CVRPTW, and QAP, we follow the corresponding \textsc{LLM4AD} generators. 
For FSS, however, \textsc{LLM4AD} uses COBench predefined instances~\cite{sun2026co}, whereas we use a synthetic generator for FSS for consistency with the other benchmark problems; we therefore replace the FSS benchmark source with the generator described in Appendix~\ref{app:fss_generation}.

For each benchmark setting, the protocol generates $N_{\mathrm{train}}=31$ training
instances and $N_{\mathrm{test}}=31$ independently generated test instances. The corresponding split
seeds are 2024 and 42. The training split is the only split accessible to synthesis and archive
construction; the test split is first accessed after the selected algorithm has been frozen.

\textbf{Interpretation of train--test objective values:}
Differences between mean training and test objective values in our setting should not be
interpreted in the same way as train--test loss gaps in supervised learning.
The splits contain different sampled optimization instances, and any split may have a
higher or lower mean raw objective simply because its instances are intrinsically harder or have
higher optimal objective values.
Accordingly, generalization is assessed by performance on unseen test instances, including
comparison against human-designed, domain-specific baselines on those same test instances, rather
than by the raw gap between mean train and mean test objective values.
Even a method that always finds the global optimum on every instance could exhibit either higher or lower average objective values on the train and test splits if the underlying instances differ in their optimal objective scales.

%------------------------------------------------------------------------------
\subsubsection{CVRP}
\label{app:cvrp_generation}
%------------------------------------------------------------------------------

Each instance contains a depot and $n$ customers. Coordinates are sampled i.i.d.\
uniformly in $[0,1]^2$ for all $n+1$ nodes (depot + customers). Customer demands are
sampled independently as integers in $\{1,\ldots,9\}$, and the depot demand is set to 0.
The vehicle capacity is fixed to $Q=40$. Distances are Euclidean.

\textbf{Parameters (default):}
$n=50$, $Q=40$, coordinate range $[0,1]$, demand range $\{1,\ldots,9\}$.

%------------------------------------------------------------------------------
\subsubsection{CVRPTW}
\label{app:cvrptw_generation}
%------------------------------------------------------------------------------

We sample one depot coordinate and $n$ customer coordinates i.i.d.\ uniformly in $[0,1]^2$.
Customer demands are sampled independently as integers in $\{1,\ldots,9\}$, and capacity is fixed to $Q=40$.
Travel time equals Euclidean distance.

Service times are sampled for customers i.i.d.\ from $\mathcal{U}(0.15,0.20)$, with depot
service time set to 0. The depot time window is fixed to $[0, T_{\max}]$ with $T_{\max}=4.6$.
Customer time windows follow the \textsc{LLM4AD} construction: for each customer $i$, a
window length $L_i \sim \mathcal{U}(0.15,0.20)$ is sampled, and the early time is set as
$e_i = \alpha_i \, d_{0i}$, where $d_{0i}$ is the distance from the depot to customer $i$, and
\[
\alpha_i = 1 + U_i \cdot \left(\frac{T_{\max} - s_i - L_i}{d_{0i}} - 2\right),
\quad U_i \sim \mathcal{U}(0,1),
\]
with $s_i$ being the service time of customer $i$. The late time is $l_i = e_i + L_i$.

\textbf{Parameters (default):}
$n=50$, $Q=40$, coordinate range $[0,1]$, demand range $\{1,\ldots,9\}$,
$T_{\max}=4.6$, service time $\mathcal{U}(0.15,0.20)$, window length $\mathcal{U}(0.15,0.20)$.

%------------------------------------------------------------------------------
\subsubsection{Permutation Flow Shop Scheduling (FSS)}
\label{app:fss_generation}
%------------------------------------------------------------------------------

For each instance, processing times are sampled i.i.d.\ as integers in $\{10,\ldots,99\}$ for each
job--machine pair, then cast to float for evaluation. We consider permutation flow shop:
a solution is a permutation of the $n$ jobs applied across all $m$ machines.

\textbf{Parameters (default):}
$n=50$ jobs, $m=10$ machines, processing times in $\{10,\ldots,99\}$.

%------------------------------------------------------------------------------
\subsubsection{Quadratic Assignment Problem (QAP)}
\label{app:qap_generation}
%------------------------------------------------------------------------------

We generate a symmetric flow matrix $F = [f_{ij}] \in \mathbb{Z}_+^{n\times n}$ and a symmetric distance matrix
$D = [d_{pq}] \in \mathbb{Z}_+^{n\times n}$. To do so, we first sample integer matrices
$\tilde{F}$ and $\tilde{D}$ with entries in $\{1,\ldots,100\}$, then symmetrize them via
elementwise floor division:
\[
F = \left\lfloor \frac{\tilde{F} + \tilde{F}^\top}{2} \right\rfloor,
\qquad
D = \left\lfloor \frac{\tilde{D} + \tilde{D}^\top}{2} \right\rfloor,
\]
and set their diagonals to zero. The objective is computed by summing over all ordered pairs
as in Eq.~\eqref{eq:qap_objective}.

\textbf{Parameters (default):}
$n=50$, matrix entries in $\{1,\ldots,100\}$, symmetric with zero diagonal.

%% file: Appendix/EvaluationProtocol.tex
This subsection specifies the evaluation protocol, including evaluator design, synthesis
budgets, runtime caps, selection of the deployable algorithm, test execution, reporting, and
statistical testing. Unless otherwise stated, experiments use the default benchmark settings defined
in the main experimental setup and the train/test generation procedure in
Appendix~\ref{app:instance_generation}.

\subsubsection{Evaluators}
\label{app:evaluators}

All methods are evaluated on the same benchmark instances for each problem, but the evaluation interface depends on the synthesis setting. 
For component-based LLM baselines (\textsc{EoH}, \textsc{ReEvo}, and \textsc{MCTS-AHD}), evaluation follows the shared scaffolded setup aligned with \textsc{LLM4AD}~\cite{liu2024llm4ad}, where the synthesized artifact is a heuristic component executed inside a fixed external framework. 
In this setting, parts of the solution-construction and constraint-handling logic are
supplied by the scaffold. The underlying constraints remain properties of the problem; the scaffold
supplies algorithmic machinery for handling them.

For full-synthesis methods (\textsc{ATLAS} and \textsc{EoH-Full}), the synthesized
artifact is a complete executable algorithm under the stated I/O interface and execution
environment.
We therefore use evaluators designed for full algorithms, which execute the synthesized program end-to-end on each benchmark instance and verify both objective quality and problem-specific feasibility conditions, including constraint satisfaction and output-format validity. 
\textsc{ATLAS} and \textsc{EoH-Full} use identical evaluators, which are provided in the \textsc{ATLAS} codebase. 
Although the evaluator interfaces differ between component-based and full-synthesis settings, they produce identical objective values for identical feasible solutions; the distinction lies in how candidate solutions are generated and validated, not in the underlying cost computation.

Thus, all methods use the same benchmark instances, but component-based and full-synthesis methods are evaluated through interfaces appropriate to their respective synthesis settings.

%------------------------------------------------------------------------------
\subsubsection{Synthesis Runs and LLM Configuration}
\label{app:eval_protocol_runs}
%------------------------------------------------------------------------------

All LLM-based synthesis methods use \textsc{OpenAI GPT-5-mini} with reasoning effort set to \textit{low} and temperature $T=1.0$.

LLM-based methods (\textsc{ATLAS}, \textsc{EoH}, \textsc{ReEvo}, \textsc{MCTS-AHD}, and
\textsc{EoH-Full}) perform $R=5$ independent synthesis runs with different random seeds, each
yielding one final optimization algorithm; the algorithms may differ across runs.
Since the human-designed, domain-specific baselines do not synthesize algorithms, they are
evaluated once on the test instances.
Each ATLAS synthesis run yields exactly one deployable algorithm, chosen on training data
alone as specified below. Every comparison method likewise contributes one training-selected
algorithm per run, subject to the same test protocol.

%------------------------------------------------------------------------------
\subsubsection{Budgets and Runtime Caps}
\label{app:eval_protocol_budgets}
%------------------------------------------------------------------------------

We use two budget notions in our experiments.

\begin{itemize}[leftmargin=*, itemsep=2pt]
\item \emph{Synthesis budgets for LLM-based methods:}
For end-to-end ATLAS component ablations and search-configuration sensitivity analyses, we
fix the search budget to $B=500$ evaluated operator executions. The embedding and initialization
analyses use their separately reported sample sizes.
For cross-method comparisons with other LLM-based baselines, we instead match total token consumption per synthesis run, since per-iteration token usage differs substantially between component synthesis and full-algorithm synthesis.
The resulting benchmark-specific token budgets are calibrated from the average token usage of 500 evaluated operator executions under full synthesis: 5M tokens for FSS, 6.5M for CVRP, 7M for CVRPTW, and 6M for QAP.

\item \emph{Per-instance runtime caps:}
For each default benchmark setting, we use a common \textit{per-instance} runtime cap across all compared methods: 210\,s for FSS, 30\,s for CVRP, 120\,s for CVRPTW, and 240\,s for QAP.
For LLM-based methods, this cap is enforced during synthesis-time evaluation: exceeding
the cap produces a failed evaluation rather than a problem-infeasibility judgment.
Runtime-sensitive human-designed baselines are evaluated under the same
benchmark-setting-specific caps.
The same full cap is enforced on training and test executions.
\end{itemize}

%------------------------------------------------------------------------------
\subsubsection{Execution Model and Human-Designed Baselines}
\label{app:eval_protocol_execution}
%------------------------------------------------------------------------------

All compared methods are executed under single-core, single-threaded settings whenever applicable. 
Each candidate--instance invocation executes within one worker process and one thread,
while different instances are evaluated in parallel. Restricting human-designed baselines to the
same per-instance execution model makes runtime caps more comparable across methods.
Human-designed baselines are evaluated directly on the test instances. NEH runs
deterministically to completion. OR-Tools is deterministic in our single-threaded configuration and
is evaluated once per test instance. The runtime-sensitive methods including PyVRP, VROOM, the IG variants,
Iterative Beam Search, RoTS, SA, BLS, and BMA are each executed once under the same
benchmark-setting-specific per-instance runtime cap.

%------------------------------------------------------------------------------
\subsubsection{Reporting and Statistical Testing}
\label{app:eval_protocol_stats}
%------------------------------------------------------------------------------

\paragraph{Training-defined ATLAS selection}
At the end of each ATLAS run, the algorithm with the best training performance is selected as the
output and evaluated on the test instances for reporting the results.

\paragraph{Test-set eligibility}
The best selected algorithm on train set is executed on all $N_{\mathrm{test}}=31$ test instances. This
stage is evaluation-only: it cannot invoke \textsc{Repair}, add or remove archive members, change
embeddings or clusters, or otherwise resume search. A run is successful only if all 31 executions
finish within the ordinary problem-specific cap and pass memory, exception, output-format,
finite-objective, and problem-feasibility checks.

\paragraph{Paired statistical comparison}
For each test instance, we average the raw objective values from the five independent runs of each
LLM-based method; human-designed methods contribute their direct per-instance raw objectives.
This averaging marginalizes over synthesis runs and yields one raw objective value per method and test instance.
Pairwise comparisons between all methods use a two-sided Wilcoxon signed-rank test at significance
level $0.05$ on the resulting 31 pairs of per-instance raw objective values.
Within each result table, boldface identifies the statistically best-performing group: the method
with the lowest mean and any method not significantly different from it.
Every textual claim that one method achieves better test-set objective performance than another is
supported by this paired test.

\paragraph{Percentage-gap display}
For each setting, let $\overline J_{m,s}$ be method $m$'s mean objective over the 31 test instances.
Among the statistically best-performing human-designed methods, the one with the lowest mean is
used as the reference with mean $\overline J_{\mathrm{ref},s}$. We report

\[
\operatorname{Gap}_{m,s}=100\,\frac{\overline J_{m,s}-\overline J_{\mathrm{ref},s}}{\overline J_{\mathrm{ref},s}}.
\]

This percentage is a descriptive gap to the selected domain-specific reference, not an optimality
gap. Statistical testing uses the per-instance raw objective values, not the percentage gaps.
Table~\ref{app:classical_reference_means} lists the resulting reference method and mean
$\overline J_{\mathrm{ref},s}$ for every benchmark setting; the implementations and configurations
of these methods are described in Appendix~\ref{app:classical_baselines}.

\begin{table}[htbp]
\centering
\caption{\emph{Raw reference means used for percentage-gap reporting.} For each problem
and size, paired tests on raw per-instance objectives identify the statistically best-performing
group of human-designed, domain-specific methods; the member of that group with the lowest mean
supplies the reported reference. These are empirical reference means on the study instances, not
optima or best-known-solution claims. The CVRP $n25$ methods tie at the reported precision.}
\label{app:classical_reference_means}
\small
{
\begin{tabular}{lll}
\toprule
\textbf{Problem setting} & \textbf{Reference method} & \textbf{Raw mean objective} \\
\midrule
FSS $(n50m10)$ & IG-TB (Taillard acc.)~\cite{ruiz2007simple,taillard1990some,fernandez2014insertion} & 3\,343.00 \\
CVRP $(n50)$ & PyVRP (ILS)~\cite{wouda2024pyvrp,pyvrp0133software} & 10.49 \\
CVRPTW $(n50)$ & PyVRP (ILS)~\cite{wouda2024pyvrp,pyvrp0133software} & 16.58 \\
QAP $(n50)$ & BMA~\cite{benlic2015memetic} & 5\,704\,446 \\
FSS $(n25m5)$ & IG-TB (Taillard acc.)~\cite{ruiz2007simple,taillard1990some,fernandez2014insertion} & 1\,599.84 \\
FSS $(n100m20)$ & IG-TB (Taillard acc.)~\cite{ruiz2007simple,taillard1990some,fernandez2014insertion} & 6\,808.45 \\
CVRP $(n25)$ & PyVRP (ILS)~\cite{wouda2024pyvrp,pyvrp0133software} / VROOM~\cite{vroom,vroom116software} & 6.19 \\
CVRP $(n100)$ & PyVRP (ILS)~\cite{wouda2024pyvrp,pyvrp0133software} & 18.46 \\
\bottomrule
\end{tabular}}
\end{table}

%% file: Appendix/ATLASProcedure.tex
This section provides the detailed procedural modules implementing the high-level ATLAS workflow in
Algorithm~\ref{alg:atlas_highlevel}.
The algorithms below specify the control flow, hyperparameter usage, operator-application strategies,
and archive-management logic.

\subsubsection{Main Execution Loop}
Algorithm~\ref{alg:atlas_main} expands the high-level workflow into the detailed execution loop from initialization
through final archive output.

\begin{algorithm2e}[t]
\small
\DontPrintSemicolon
\SetKwComment{tcp}{// }{}
\SetKwFunction{Init}{InitializeArchive}
\SetKwFunction{Gen}{GenerateCandidates}
\SetKwFunction{Update}{UpdateArchive}
\SetKwFunction{Cluster}{Cluster}
\SetKwFunction{Evaluate}{Evaluate}
\SetKwFunction{Repair}{Repair}
\SetKwFunction{Embed}{ComputeEmbedding}
\SetKwFunction{Combine}{Combine}
\SetKwFunction{Diverge}{Diverge}
\SetKwFunction{Improve}{Improve}
\SetKwFunction{Tune}{Tune}
\SetKwFunction{Simplify}{Simplify}
\SetKwFunction{Create}{Create}
\SetKwFunction{Pruning}{ArchiveSizeManagement}
\SetKwFunction{Dedup}{ArchiveDeduplication}

\caption{ATLAS: Main Execution Loop}
\label{alg:atlas_main}
\KwIn{\textbf{Problem \& Model:} $\mathcal{P}$ (problem), $\mathcal{M}$ (LLM)}
\KwIn{\textbf{Search Budget:} $B$ (iterations)}
\KwIn{\textbf{Archive:} $N_{\max}$ (capacity), $\tau_{\text{strict}}$ and $\tau_{\text{soft}}$ (similarity thresholds), $\epsilon$ (performance tolerance)}
\KwIn{\textbf{Layer~1 Neighborhood:} $k$ (kNN size)}
\KwIn{\textbf{Initialization:} $n_{\text{init}}^{\text{crt}}$ (Phase~1 count), $n_{\text{init}}^{\text{div}}$ (Phase~2 count)}
\KwIn{\textbf{Operator Counts:} $n_{\text{intensify}}$ (refine repeats on $a^*$)}
\KwIn{\textbf{Reference Sizes:} $m_{\text{comb}}$ (\Combine refs), $m_{\text{div}}$ (\Diverge refs)}
\KwIn{\textbf{Operator Probabilities:} $p_{\text{imp}}$, $p_{\text{tune}}$, $p_{\text{simp}}$ (refinement ops in Layers 1~\&~2)}
\KwOut{Best Algorithm $a^*$, Archive $\mathcal{A}$, and Cluster Representatives $\mathcal{T}$}

\BlankLine
\tcp{1. Initialization (See Alg.~\ref{alg:atlas_init})}
$\mathcal{A} \leftarrow \Init(\mathcal{P}, \mathcal{M}, n_{\text{init}}^{\text{crt}}, n_{\text{init}}^{\text{div}}, m_{\text{div}})$ \;
\BlankLine
\tcp{Compute initial distance matrix}
Compute pairwise distance matrix $\mathbf{D}$\;
$\mathcal{A}, \mathbf{D} \leftarrow \Dedup(\mathcal{A}, \mathbf{D}, \tau_{\text{strict}}, \tau_{\text{soft}}, \epsilon)$ \quad \tcp{Deduplicate initialization archive}
\BlankLine
\While(\tcp*[f]{Main Search Loop}){\text{Budget } $B$ \text{ not exhausted}}{
    \BlankLine
    \tcp{2. Layer Configuration from the Maintained Archive}
    $a^* \leftarrow \argmin_{a \in \mathcal{A}} J(a)$ \quad \tcp{where $J(a) = a.J$} 
    $\mathcal{G}^* \leftarrow \{a^*\} \cup \text{kNN}(a^*, \mathcal{A}, k, \mathbf{D})$ \quad \tcp{Elite group (Layer 1)}
    \BlankLine
    $\{C_1, \dots, C_K\} \leftarrow \Cluster(\mathcal{A}, \mathbf{D})$ \quad \tcp{Cluster entire archive}
    $C_{\text{best}} \leftarrow \{C_k : a^* \in C_k\}$ \quad \tcp{Cluster containing $a^*$}
    Remove $\mathcal{G}^*$ from the remaining clusters and discard empty clusters\;
$\Omega \leftarrow \{(C_i, r_i) : C_i \neq C_{\text{best}},\ C_i \neq \emptyset,\ r_i = \argmin_{a
\in C_i} J(a)\}$ \quad \tcp{Re-select representatives after elite removal}
\tcp{3. Three-Layer Candidate Generation (See Alg.~\ref{alg:atlas_gen})}
$\mathcal{C} \leftarrow \Gen(\mathcal{G}^*, \Omega, a^*, m_{\text{comb}}, m_{\text{div}}, n_{\text{intensify}}, p_{\text{imp}}, p_{\text{tune}}, p_{\text{simp}})$ \;
    \BlankLine
    \tcp{4. Candidate Evaluation \& Archive Update}
    \ForEach{$c \in \mathcal{C}$}{
        $c.J, c.\text{error} \leftarrow \Evaluate(c, \mathcal{I}_{\text{train}})$\;
        \If{$c.\text{error} \neq \text{null}$}{
            $c \leftarrow \Repair(c, c.\text{error}, \mathcal{P}, \mathcal{M})$\;
            $c.J, c.\text{error} \leftarrow \Evaluate(c, \mathcal{I}_{\text{train}})$\;
        }
        \If{$c.\text{error} = \text{null}$}{
            $c.\text{embed} \leftarrow \Embed(c)$\;
            $\mathcal{A} \leftarrow \mathcal{A} \cup \{c\}$\;
        }
    }
    \tcp{Update distances for newly added algorithms}
    Update distance matrix $\mathbf{D}$ for new algorithms\;
    $\mathcal{A}, \mathbf{D} \leftarrow \Dedup(\mathcal{A}, \mathbf{D}, \tau_{\text{strict}}, \tau_{\text{soft}}, \epsilon)$ \quad \tcp{See Alg.~\ref{alg:atlas_dedup}}
    $a^* \leftarrow \argmin_{a \in \mathcal{A}} J(a)$\;
    $\mathcal{G}^* \leftarrow \{a^*\} \cup \text{kNN}(a^*, \mathcal{A}, k, \mathbf{D})$ \quad \tcp{Protect the updated best neighborhood}
    \If{$|\mathcal{A}|>N_{\max}$}{
        $\mathcal{A}, \mathbf{D} \leftarrow \Pruning(\mathcal{A}, \mathbf{D}, N_{\max},\mathcal{G}^*)$ \quad \tcp{See Alg.~\ref{alg:atlas_prune}}
    }
}
\BlankLine
\tcp{Final cluster representatives}
$\{C_1, \dots, C_K\} \leftarrow \Cluster(\mathcal{A}, \mathbf{D})$ \quad \tcp{Cluster the final archive}
$\mathcal{T} \leftarrow \{\argmin_{a \in C_k} J(a) : k = 1, \dots, K\}$\;
\Return $a^*, \mathcal{A}, \mathcal{T}$
\end{algorithm2e}

\subsubsection{Initialization}
Algorithm~\ref{alg:atlas_init} implements the two-phase bootstrap that establishes initial archive coverage: diverse generation via \textsc{Create} (Phase 1), followed by paradigm diversification via \textsc{Diverge} (Phase 2).

\begin{algorithm2e}[t]
\small
\DontPrintSemicolon
\SetKwComment{tcp}{// }{}
\SetKwFunction{Create}{Create}
\SetKwFunction{Diverge}{Diverge}
\SetKwFunction{Repair}{Repair}
\SetKwFunction{Evaluate}{Evaluate}
\SetKwFunction{Embed}{ComputeEmbedding}

\caption{Module: Two-Phase Initialization}
\label{alg:atlas_init}

\KwIn{$\mathcal{P}$ (problem), $\mathcal{M}$ (LLM), $n_{\text{init}}^{\text{crt}}$ (Phase~1 count), $n_{\text{init}}^{\text{div}}$ (Phase~2 count), $m_{\text{div}}$ (Phase~2 refs)}

\KwOut{Initial Archive $\mathcal{A}$}

\BlankLine
$\mathcal{A} \leftarrow \emptyset$\;

\BlankLine
\tcp{Phase 1: Bootstrapping via Create}
\For{$i = 1$ \KwTo $n_{\text{init}}^{\text{crt}}$}{
    $a \leftarrow \Create(\mathcal{P}, \mathcal{M})$\;
    $a.J, a.\text{error} \leftarrow \Evaluate(a, \mathcal{I}_{\text{train}})$\;
    \If{$a.\text{error} \neq \text{null}$}{
        $a \leftarrow \Repair(a, a.\text{error}, \mathcal{P}, \mathcal{M})$\;
        $a.J, a.\text{error} \leftarrow \Evaluate(a, \mathcal{I}_{\text{train}})$\;
    }
    \If{$a.\text{error} = \text{null}$}{
        $a.\text{embed} \leftarrow \Embed(a)$\;
        $\mathcal{A} \leftarrow \mathcal{A} \cup \{a\}$\;
    }
}

\BlankLine
\tcp{Phase 2: Paradigm Diversification via \textsc{Diverge}}
\For{$i = 1$ \KwTo $n_{\text{init}}^{\text{div}}$}{
    Randomly sample references $\mathcal{R} \subset \mathcal{A}$ of size $m_{\text{div}}$\;
    $a \leftarrow \Diverge(\mathcal{R}, \mathcal{P}, \mathcal{M})$\;
    $a.J, a.\text{error} \leftarrow \Evaluate(a, \mathcal{I}_{\text{train}})$\;
    \If{$a.\text{error} \neq \text{null}$}{
        $a \leftarrow \Repair(a, a.\text{error}, \mathcal{P}, \mathcal{M})$\;
        $a.J, a.\text{error} \leftarrow \Evaluate(a, \mathcal{I}_{\text{train}})$\;
    }
    \If{$a.\text{error} = \text{null}$}{
        $a.\text{embed} \leftarrow \Embed(a)$\;
        $\mathcal{A} \leftarrow \mathcal{A} \cup \{a\}$\;
    }
}

\BlankLine
\Return $\mathcal{A}$

\end{algorithm2e}

\subsubsection{Archive Maintenance}

Algorithms~\ref{alg:atlas_dedup} and~\ref{alg:atlas_prune} manage archive quality and capacity.

\paragraph{Deduplication (Algorithm~\ref{alg:atlas_dedup})} 
The two-tier mechanism removes semantic duplicates (strict threshold $\tau_{\text{strict}}$) and redundant near-duplicates that offer negligible performance gains (soft threshold $\tau_{\text{soft}}$ with absolute tie tolerance $\epsilon$).

\paragraph{Capacity-Constrained Pruning (Algorithm~\ref{alg:atlas_prune})} When the archive exceeds $N_{\max}$, pruning must remove algorithms to maintain computational feasibility.
Fitness-ranked survival can reduce representation of lower-fitness alternatives. ATLAS
instead prioritizes \emph{embedding-space coverage} through a similarity-first strategy: it
iteratively identifies the most similar pair $(a_i, a_j)$ via distance matrix $\mathbf{D}$ and
removes the worse performer.
This forces the archive to spread across the semantic space rather than concentrating on high-performing regions. 
Note that performance is used only to decide \emph{which member} of the closest pair to remove, not to determine \emph{which pair} to prune.

\textit{Elite Protection to Resolve Conflict:} 
A pure similarity-first strategy would aggressively prune $\mathcal{G}^*$, as this region around $a^*$ is naturally the densest in the archive. 
However, these algorithms have survived the configured deduplication thresholds and serve
as the refinement anchor for Layer~1's intensive search.
Removing them would weaken exploitation capability precisely where it matters most. 
Therefore, Algorithm~\ref{alg:atlas_prune} explicitly protects $\mathcal{G}^*$, maintaining both global diversity and local refinement power under capacity constraints.
Crucially, this protection is \emph{dynamic}: when a superior algorithm is discovered in a different region, $\mathcal{G}^*$ shifts immediately to the new neighborhood, and the previous elite region (now stripped of protection and often highly dense) becomes a primary target for pruning. 
This allows ATLAS to reclaim capacity from regions that are no longer the current best
while securing refinement power around a newly leading region, without unbounded archive growth.

\textbf{Note on Algorithm~\ref{alg:atlas_prune}:}
Algorithm~\ref{alg:atlas_prune} is presented at a high level. In the configurations considered in this paper, the archive size is always substantially larger than the protected elite region $|\mathcal{G}^*|$, so capacity-constrained pruning always has many removable non-elite pairs available. Consequently, the case in which all candidate pairs are protected does not arise under the intended operating regime; handling such degenerate configurations is an implementation detail omitted from the pseudocode.

\begin{algorithm2e}[t]
\small
\DontPrintSemicolon
\SetKwComment{tcp}{// }{}

\caption{Module: Archive Deduplication}
\label{alg:atlas_dedup}

\KwIn{Archive $\mathcal{A}$, distance matrix $\mathbf{D}$, $\tau_{\text{strict}}$ (strict similarity threshold), $\tau_{\text{soft}}$ (soft similarity threshold), $\epsilon$ (performance tolerance)}

\KwOut{Deduplicated Archive $\mathcal{A}$ and updated $\mathbf{D}$}

\BlankLine
\tcp{Two-Tier Deduplication Strategy}
\ForEach{pair $(a_i, a_j) \in \mathcal{A} \times \mathcal{A}$ with $i < j$}{
    \uIf{$D_{ij} < 1-\tau_{\text{strict}}$}{
        \tcp{Strict Tier: Remove near-duplicate regardless of performance}
        \eIf{$a_i.J \neq a_j.J$}{
            Remove $\argmax_{a \in \{a_i, a_j\}} a.J$ from $\mathcal{A}$ \tcp*{Remove worse}
        }{
            Remove random $a \in \{a_i, a_j\}$ from $\mathcal{A}$ \tcp*{Identical performance}
        }
        Remove corresponding row/column from $\mathbf{D}$\;
    }
    \uElseIf{$D_{ij} < 1-\tau_{\text{soft}}$ \textbf{and} $|a_i.J - a_j.J| < \epsilon$}{
        \tcp{Soft Tier: Remove similar with close performance}
        \eIf{$a_i.J \neq a_j.J$}{
            Remove $\argmax_{a \in \{a_i, a_j\}} a.J$ from $\mathcal{A}$ \tcp*{Remove worse}
        }{
            Remove random $a \in \{a_i, a_j\}$ from $\mathcal{A}$ \tcp*{Identical performance}
        }
        Remove corresponding row/column from $\mathbf{D}$\;
    }
}

\BlankLine
\Return $\mathcal{A}, \mathbf{D}$

\end{algorithm2e}

\begin{algorithm2e}[t]
\small
\DontPrintSemicolon
\SetKwComment{tcp}{// }{}

\caption{Module: Archive Size Management}
\label{alg:atlas_prune}

\KwIn{Archive $\mathcal{A}$, distance matrix $\mathbf{D}$, $N_{\max}$ (capacity), $\mathcal{G}^*$ (protected elite region)}

\KwOut{Pruned Archive $\mathcal{A}$ and updated $\mathbf{D}$}

\BlankLine
$\mathcal{E} \leftarrow \emptyset$ \tcp*{Set of skipped protected pairs}

\BlankLine
\tcp{Iterative Capacity-Constrained Pruning with Elite Protection}
\While{$|\mathcal{A}| > N_{\max}$}{

    $(a_i, a_j) \leftarrow \argmin_{\substack{x,y \in \mathcal{A} \\ x \neq y,\ (x,y)\notin \mathcal{E}}} D_{xy}$ \tcp*{Find closest unexamined pair}
    
    \BlankLine
    \uIf{$a_i \in \mathcal{G}^*$ \textbf{and} $a_j \in \mathcal{G}^*$}{
        \tcp{Both protected: skip this pair}
        $\mathcal{E} \leftarrow \mathcal{E} \cup \{(a_i, a_j)\}$\;
        continue to next closest pair\;
    }
    \uElseIf{$a_i \in \mathcal{G}^*$ \textbf{or} $a_j \in \mathcal{G}^*$}{
        \tcp{One protected: remove the unprotected one}
        \eIf{$a_i \in \mathcal{G}^*$}{
            Remove $a_j$ from $\mathcal{A}$\;
        }{
            Remove $a_i$ from $\mathcal{A}$\;
        }
        Remove corresponding row/column from $\mathbf{D}$\;
    }
    \Else{
        \tcp{Neither protected: remove worse performer}
        \eIf{$a_i.J \neq a_j.J$}{
            Remove $\argmax_{a \in \{a_i, a_j\}} a.J$ from $\mathcal{A}$\;
        }{
            Remove random $a \in \{a_i, a_j\}$ from $\mathcal{A}$\;
        }
        Remove corresponding row/column from $\mathbf{D}$\;
    }
}

\BlankLine
\Return $\mathcal{A}, \mathbf{D}$

\end{algorithm2e}

\subsubsection{Three-Layer Search Strategy}
\label{app:threeLayerSearch}

\begin{algorithm2e}[t]
\small
\DontPrintSemicolon
\SetKwComment{tcp}{// }{}
\SetKwFunction{Improve}{Improve}
\SetKwFunction{Tune}{Tune}
\SetKwFunction{Simplify}{Simplify}
\SetKwFunction{Combine}{Combine}
\SetKwFunction{Diverge}{Diverge}

\caption{Module: Three-Layer Candidate Generation}
\label{alg:atlas_gen}

\KwIn{$\mathcal{P}$ (problem), $\mathcal{M}$ (LLM), Layer~1 region $\mathcal{G}^*$, best $a^*$,
semantic clusters $\Omega = \{(C_i, r_i)\}$}
\KwIn{Counts: $n_{\text{intensify}}$}
\KwIn{Reference sizes: $m_{\text{comb}}$, $m_{\text{div}}$}
\KwIn{Probabilities: $p_{\text{imp}}$, $p_{\text{tune}}$, $p_{\text{simp}}$}

\KwOut{Candidate Set $\mathcal{C}$}

\BlankLine
$\mathcal{C} \leftarrow \emptyset$\;

\tcp{Layer 1: Intensive Refinement of Best Algorithm}
\For{$i = 1$ \KwTo $n_{\text{intensify}}$}{
    $\mathcal{C} \leftarrow \mathcal{C} \cup \{\Improve(a^*, \mathcal{P}, \mathcal{M})\}$\;
    $\mathcal{C} \leftarrow \mathcal{C} \cup \{\Tune(a^*, \mathcal{P}, \mathcal{M})\}$\;
    $\mathcal{C} \leftarrow \mathcal{C} \cup \{\Simplify(a^*, \mathcal{P}, \mathcal{M})\}$\;
}

\tcp{Layer 1: Probabilistic Refinement of Neighbors}
\ForEach{$a \in \mathcal{G}^* \setminus \{a^*\}$}{
    \If{$\text{random}() < p_{\text{imp}}$}{
        $\mathcal{C} \leftarrow \mathcal{C} \cup \{\Improve(a, \mathcal{P}, \mathcal{M})\}$\;
    }
    \If{$\text{random}() < p_{\text{tune}}$}{
        $\mathcal{C} \leftarrow \mathcal{C} \cup \{\Tune(a, \mathcal{P}, \mathcal{M})\}$\;
    }
    \If{$\text{random}() < p_{\text{simp}}$}{
        $\mathcal{C} \leftarrow \mathcal{C} \cup \{\Simplify(a, \mathcal{P}, \mathcal{M})\}$\;
    }
}

\tcp{Layer 1: Combination within Best Region}
    \For{$j = 1$ \KwTo $|\mathcal{G}^*|$}{
    Randomly sample $\mathcal{R}' \subset \mathcal{G}^* \setminus \{a^*\}$ of size $m_{\text{comb}} - 1$\;
    $\mathcal{R} \leftarrow \{a^*\} \cup \mathcal{R}'$ \quad \tcp{Ensure $a^*$ included}
    $\mathcal{C} \leftarrow \mathcal{C} \cup \{\Combine(\mathcal{R}, \mathcal{P}, \mathcal{M})\}$\;
}

\tcp{Layer 2: Distributed Regional Refinement}
\ForEach{cluster $(C_k, r_k) \in \Omega$}{
    \If{$\text{random}() < p_{\text{imp}}$}{
        $\mathcal{C} \leftarrow \mathcal{C} \cup \{\Improve(r_k, \mathcal{P}, \mathcal{M})\}$\;
    }
    \If{$\text{random}() < p_{\text{tune}}$}{
        $\mathcal{C} \leftarrow \mathcal{C} \cup \{\Tune(r_k, \mathcal{P}, \mathcal{M})\}$\;
    }
    \If{$\text{random}() < p_{\text{simp}}$}{
        $\mathcal{C} \leftarrow \mathcal{C} \cup \{\Simplify(r_k, \mathcal{P}, \mathcal{M})\}$\;
    }
    Randomly sample $\mathcal{R}' \subset C_k \setminus \{r_k\}$ of size $m_{\text{comb}} - 1$\;
        $\mathcal{R} \leftarrow \{r_k\} \cup \mathcal{R}'$ \quad \tcp{Ensure $r_k$ included}
        $\mathcal{C} \leftarrow \mathcal{C} \cup \{\Combine(\mathcal{R}, \mathcal{P}, \mathcal{M})\}$\;
}

\tcp{Layer 3: Cross-Cluster Operators}
\For{$j = 1$ \KwTo $|\Omega|$}{
    Randomly sample $\mathcal{R} \subset \{\,r_k : (C_k, r_k) \in \Omega\,\}$ of size $m_{\text{comb}}$ from different clusters\;
    $\mathcal{C} \leftarrow \mathcal{C} \cup \{\Combine(\mathcal{R}, \mathcal{P}, \mathcal{M})\}$\;
    
\BlankLine

    Randomly sample $\mathcal{R}' \subset \{\,r_k : (C_k, r_k) \in \Omega\,\}$ of size $m_{\text{comb}} - 1$ from different clusters\;
    $\mathcal{R} \leftarrow \{a^*\} \cup \mathcal{R}'$\; 
    $\mathcal{C} \leftarrow \mathcal{C} \cup \{\Combine(\mathcal{R}, \mathcal{P}, \mathcal{M})\}$\;
    
\BlankLine

    Randomly sample $\mathcal{R}'' \subset \{\,r_k : (C_k, r_k) \in \Omega\,\}$ of size $m_{\text{div}} - 1$\;
    $\mathcal{R} \leftarrow \{a^*\} \cup \mathcal{R}''$ \quad \tcp{Include $a^*$ to stay distinct from the $\mathcal{G}^*$ region}
    $\mathcal{C} \leftarrow \mathcal{C} \cup \{\Diverge(\mathcal{R}, \mathcal{P}, \mathcal{M})\}$\;
}

\BlankLine
\Return $\mathcal{C}$

\end{algorithm2e}

Algorithm~\ref{alg:atlas_gen} specifies the hierarchical search across intensive best-region 
refinement (Layer 1), distributed regional refinement (Layer 2), and cross-cluster discovery (Layer 3).

\paragraph{Note on Intensive Refinement in Layer 1} 
In many evolutionary algorithms, the selection process yields multiple near-clones of the best individual ($a^*$), which are then refined independently through variation operators~\cite{liu2024evolution} in each generation. 
This effectively intensifies refinement on $a^*$.
ATLAS enforces strict deduplication, so $a^*$ exists as a unique instance. 
To compensate, Layer 1 explicitly iterates refinement operators on $a^*$ for $n_{\text{intensify}}$ times.

\paragraph{Note on Layer 3 Operator Strategies}
\label{app:note_layer3_strategies}
Layer 3 applies multi-reference operators to cluster representatives to enable global exploration:

\begin{itemize}
    \item \textbf{Cross-Cluster \textsc{Combine}:} Combines representatives from different
embedding clusters and instructs the LLM to integrate useful mechanisms from them. The operation is
intended to generate hybrids or intermediate designs between represented regions.
    
    \item \textbf{Best-Distant \textsc{Combine}:} Combines the current best algorithm
$a^*$ with representatives from other embedding clusters. Unlike Layer~1's local synthesis with
nearby candidates, this operation anchors the prompt with $a^*$ while supplying more distant
reference context and asking the LLM to integrate complementary ideas.
    
    \item \textbf{\textsc{Diverge} Includes $a^*$:}
    The \textsc{Diverge} operator explicitly includes $a^*$ in the reference set $\mathcal{R}$. 
    This complements Layer~1, which already concentrates refinement effort around the current best region. 
    Including $a^*$ as a reference instructs the LLM to avoid variants too close to that
already sampled region and to propose an alternative separated from the referenced approaches.
\end{itemize}

%% file: Appendix/Hyperparameters.tex
\subsubsection{Embedding Strategy}
\label{app:embedding_strategy}

\textbf{Chosen representation (early fusion + \textsc{mGTE}):}
Based on the embedding representation analysis in Appendix~\ref{app:embedding_ablation},
we represent each algorithm $a$ using a \emph{single} embedding produced by
\textsc{mGTE-large-en-v1.5}~\cite{zhang2024mgte} from the concatenation of its \emph{name} $n$,
\emph{description} $d$, and \emph{preprocessed code} $c$, where comments, docstrings, and blank
lines are removed.
Similarity is computed in this embedding space using cosine similarity under the early-fusion formulation in Eq.~\eqref{eq:concatSimilarity}.

In Appendix~\ref{app:embedding_ablation}, the \emph{early-fusion + \textsc{mGTE}} configuration achieved the highest Class Cohesion Index (Eq.~\eqref{eq:CCI}) and the second-best accuracy.
Compared with the top-accuracy variant, this representation is simpler, more practical, and based on an open-source model, which motivates its use as the default representation in ATLAS.

\textbf{Practical details:}
All embeddings are $\ell_2$-normalized before similarity computation, so cosine similarity is equivalent to the dot product between normalized vectors.
%------------------------------------------------------------------------------
\subsubsection{Clustering Details}
\label{app:clustering}
%------------------------------------------------------------------------------

\textbf{Clustering objective and algorithm:}
We cluster candidate algorithms using \emph{k-medoids} on a precomputed distance matrix $\mathbf{D}$, solved with FasterPAM~\cite{schubert2021fastpam}, an optimized variant of the classical PAM (Partitioning Around Medoids) procedure~\cite{kaufman1990finding}. 
K-medoids is used because it operates directly on arbitrary pairwise distances and yields medoids that are \emph{actual} candidate algorithms in the archive, which is useful for downstream representative-based search.

\textbf{Distance matrix:}
For the archive $\mathcal{A}$, pairwise distances are computed from cosine similarity in the early-fusion embedding space:
\[
D_{ij} = 1 - \text{sim}_{\text{concat}}(a_i,a_j),
\qquad a_i,a_j \in \mathcal{A},
\]
where $\text{sim}_{\text{concat}}(\cdot,\cdot)$ is defined in Eq.~\eqref{eq:concatSimilarity}.

\textbf{Initialization (PAM BUILD):}
Each FasterPAM run is initialized with the PAM BUILD procedure~\cite{kaufman1990finding}, which greedily selects an initial set of medoids before refinement.

\textbf{Multiple restarts:}
For each candidate $K$, we perform $n_{\text{init}} = 10$ independent FasterPAM runs with different random seeds and a per-run iteration budget of $\texttt{max\_iter}=300$, and retain the solution with the lowest k-medoids objective value. 
This reduces sensitivity to initialization and helps avoid poor local minima~\cite{kaufman1990finding}.

\textbf{Automatic $K$ selection (Silhouette):}
We select $K$ automatically by maximizing the average Silhouette score~\cite{rousseeuw1987silhouettes} over a candidate range:
\[
K \in \{K_{\min},\ldots,K_{\max}\}, \qquad
K_{\max} = \max\!\left(K_{\min}, \left\lfloor \sqrt{|\mathcal{A}|} \right\rfloor\right),
\]
where clustering is performed on the full archive $\mathcal{A}$. 
We use $K_{\min}=3$ in ATLAS because the cluster containing the current best algorithm $a^*$ is removed after clustering; thus, at least two clusters must remain to support Layer~3 operations. 
For each candidate $K$, we cluster $\mathcal{A}$ with FasterPAM and compute the Silhouette score using the precomputed distance matrix. 
The selected value $K^*$ is the maximizer, with ties broken in favor of smaller $K$.

\textbf{Singleton-cluster handling:}
After clustering, we only post-process \emph{singleton} clusters ($|C|=1$). 
Each singleton is merged into the nearest non-singleton cluster, where ``nearest'' is defined by the minimum average distance from the singleton to members of the target cluster.

\textbf{Representative selection:}
For each cluster, we select as representative the best-performing algorithm in that cluster, i.e., the member with the lowest objective value. 
This representative serves as the cluster's strongest current exemplar for downstream refinement and cross-cluster synthesis.

\subsubsection{Hyperparameter Configuration}
\label{app:hyperparameters}

This section documents all hyperparameters used in ATLAS together with their design rationale. 
Unless otherwise noted, these values are fixed across all experiments.

\paragraph{Archive management}
\begin{itemize}[leftmargin=*,itemsep=3pt]

\item $\boldsymbol{N_{\max}=100}$\textbf{:}
Maximum archive size. 
ATLAS maintains a pairwise embedding distance matrix and performs clustering over the archive, both of which scale at least quadratically with $|\mathcal{A}|$. 
A capacity of 100 permits coverage across multiple embedding-space regions while keeping
per-iteration overhead manageable.
Under the square-root heuristic used for clustering, this yields at most $K_{\max}\approx 10$ candidate clusters.

\item $\boldsymbol{\tau_{\text{strict}} = 0.95}$\textbf{:}
Strict similarity threshold for deduplication. 
This value was chosen based on an exploratory analysis of aggregated archives from 8 pilot runs on FSS and CVRP (800 algorithms per problem), and then fixed for all experiments. 
Although pairwise cosine-similarity distributions vary across problems, manual inspection showed that pairs with similarity greater than $0.95$ predominantly corresponded to clones, functionally equivalent rewrites, trivial edits, or near-identical parameter variants. 
We therefore use $\tau_{\text{strict}}=0.95$ to remove near-duplicates regardless of performance difference.

\item $\boldsymbol{\tau_{\text{soft}} = 0.90}$\textbf{:}
Soft similarity threshold for performance-aware deduplication. 
Pairs above this threshold are highly similar under the selected representation and often
differ only in local design choices, while still sometimes exhibiting meaningful performance
differences.
We therefore treat similarity greater than $0.90$ as a candidate near-duplicate region and apply the performance tolerance below before pruning.

\item $\boldsymbol{\epsilon = 0.01}$\textbf{:}
Performance tolerance for soft-tier deduplication. 
Unlike strict deduplication, which removes extremely similar pairs regardless of performance difference, soft-tier deduplication applies only to highly similar pairs with nearly identical average training objective values.
Specifically, for pairs with similarity greater than $\tau_{\text{soft}}$, one algorithm is removed only if the absolute difference in their average training objective values is smaller than $\epsilon$. 
We set $\epsilon=0.01$ as a very small tie tolerance so that soft deduplication removes only highly similar algorithms with effectively identical performance. 
In our benchmark settings, performance differences among competitive algorithms are typically much larger than this threshold, so $\epsilon$ functions as a near-equality check rather than a substantive performance margin. 
This absolute tolerance is appropriate for the objective scales considered here, but may require adjustment for problems with substantially smaller-magnitude objective values.
\end{itemize}

\paragraph{Initialization}
\begin{itemize}[leftmargin=*,itemsep=3pt]

\item \textbf{Total: 50 initial algorithms (20 Phase~1, 30 Phase~2):}
We generate 50 initial candidates, corresponding to 50\% of $N_{\max}=100$, to provide sufficient pre-search coverage for clustering and k-NN retrieval.
Before the main loop, deduplication removes near-identical seeds, often from Phase~1, so the retained initial archive size can be smaller than 50.
Under the default budget ($B=500$ operator calls), initialization consumes 10\% of synthesis resources, leaving 90\% for iterative refinement in the three-layer strategy.
This balances initial coverage with efficiency: too small a seed set delays coverage
across embedding-space regions, while too large a seed set spends excessive budget on seeding rather
than targeted search.

\item $\boldsymbol{n_{\text{init}}^{\text{crt}} = 20}$\textbf{:}
Phase~1 uses \textsc{Create} for from-scratch generation.
In practice, zero-shot synthesis often concentrates on canonical strategies, such as standard greedy construction and insertion heuristics.
We therefore use 20 \textsc{Create} calls to obtain a diverse initial seed set without allocating excessive budget to a mode that tends to produce overlapping starting points.

\item $\boldsymbol{n_{\text{init}}^{\text{div}} = 30}$\textbf{:}
Phase~2 is weighted more heavily ($30>20$) to explicitly expand beyond the dominant modes produced by \textsc{Create}.
\textsc{Diverge} is applied iteratively to a growing pool: it initially uses Phase~1 seeds as references to avoid, then repeatedly samples from the current pool, including previously diverged algorithms, to drive further structural diversification.
This bootstraps multiple clusters in embedding space before the main loop and provides diverse anchors for subsequent clustering and three-layer search.
\end{itemize}

\paragraph{Layer 1 configuration.}
\begin{itemize}[leftmargin=*,itemsep=3pt]
\item $\boldsymbol{|\mathcal{G}^*| = 5}$ ($\boldsymbol{a^*}$ \textbf{+ 4 nearest neighbors})\textbf{:}
The elite set size balances local refinement around the best-performing region against the budget reserved for Layers~2--3. 
A small neighborhood provides multiple refinement anchors around $a^*$ for k-NN sampling and local search while keeping Layer~1 overhead bounded. 
Smaller elite sets weaken local coverage, whereas larger sets increase per-iteration cost because more members undergo refinement.

\item $\boldsymbol{n_{\text{intensify}} = 3}$\textbf{:}
Number of intensification rounds applied to $a^*$. 
In Layer~1, neighbors in $\mathcal{G}^* \setminus \{a^*\}$ are refined via single-reference operators sampled according to operator probabilities ($p_{\text{imp}}, p_{\text{tune}}, p_{\text{simp}}$). 
In contrast, $a^*$ is excluded from probabilistic refinement and receives special treatment: in each intensification round, we always apply \textsc{Improve}, \textsc{Tune}, and \textsc{Simplify} to $a^*$, thereby maintaining strong exploitation pressure. 
This compensates for strict deduplication: selection-driven evolutionary methods implicitly generate many near-clones of the current best that each undergo refinement, whereas ATLAS removes such clones to preserve diversity and therefore requires explicit intensification.

\textbf{Sensitivity:} 
The value $n_{\text{intensify}}=3$ was selected based solely on training-set performance.
The training analysis indicated that $n_{\text{intensify}}=2$ under-exploited the current-best region, whereas $n_{\text{intensify}}=4$ provided insufficient additional improvement over $n_{\text{intensify}}=3$ to justify reducing the budget available to Layers~2--3.
For interpretability, Table~\ref{tab:sensitivity_intensify_gap} reports the corresponding test-set relative mean gaps for the candidate settings.
These test results were not used either to select $n_{\text{intensify}}$ or to select the algorithm within any run.

\end{itemize}

\begin{table}[h]
\centering
\caption{\emph{Relative mean gaps for the $n_{\mathrm{intensify}}$ sensitivity
analysis.} Percent gap to the strongest human-designed reference in each setting (PyVRP for CVRP and
IG-TB for FSS); lower is better. Values are computed from the underlying raw objective means. 
The setting $n_{\mathrm{intensify}}=3$ was selected using training-set performance only; the test-set gaps are reported only for interpretability.}
\label{tab:sensitivity_intensify_gap}
\small
{
\begin{tabular}{@{}lccc@{\hspace{1em}}ccc@{}}
\toprule
& \multicolumn{3}{c@{\hspace{1em}}}{\textbf{CVRP}$(n50)$ gap (\%)$\downarrow$} & \multicolumn{3}{c}{\textbf{FSS}$(n50m10)$ gap (\%)$\downarrow$} \\
\cmidrule(lr){2-4}\cmidrule(lr){5-7}
$\boldsymbol{n_{\text{intensify}}}$ & 2 & 3 & 4 & 2 & 3 & 4 \\
\midrule
\textbf{Relative mean gap (\%)} & 0.572 & \textbf{0.094} & \textbf{0.095} & 0.401 & \textbf{0.347} & \textbf{0.351} \\
\bottomrule
\end{tabular}}
\end{table}

\paragraph{Clustering}

ATLAS clusters the archive using k-medoids (FasterPAM) on the precomputed embedding distance matrix, with $K$ selected automatically by maximizing the average Silhouette score over $[K_{\min}, K_{\max}]$. 
Each cluster representative is selected as its best-performing member, i.e., the algorithm with the lowest objective value. 
Complete clustering details, including initialization, restarts, and singleton handling, are provided in Appendix~\ref{app:clustering}.

\begin{itemize}[leftmargin=*,itemsep=3pt]
\item $\boldsymbol{K_{\min}=3}$ \textbf{(structural minimum):}
Minimum cluster count required to sustain the three-layer architecture. 
Since the cluster containing $a^*$ is excluded from Layer~2 and Layer~3 selection, at least $K_{\min}-1=2$ non-elite clusters must remain available for Layer~2 maturation and Layer~3 cross-cluster search. 
This ensures that Layer~3 can draw representatives from multiple clusters for
\textsc{Combine} and provides alternative represented regions for \textsc{Diverge} across
iterations.

\item $\boldsymbol{K_{\max}=\max(3,\lfloor\sqrt{|\mathcal{A}|}\rfloor)}$ \textbf{(granularity bound):}
Upper bound for automatic $K$ selection, following the square-root heuristic to balance granularity with stability. 
For $|\mathcal{A}|=100$, this yields $K_{\max}=10$. 
This bound discourages Silhouette optimization from fragmenting the embedding space into
very small clusters and keeps representative-based operations supplied with nontrivial groups.
The $\max(3,\cdot)$ term ensures $K_{\max}\ge K_{\min}$ when the archive is small.
\end{itemize}

\paragraph{Multi-reference operator reference sizes}

These operator-level settings are used consistently wherever \textsc{Combine} and \textsc{Diverge} are applied.

\begin{itemize}[leftmargin=*,itemsep=3pt]
\item $\boldsymbol{m_{\text{comb}} = 2}$ \textbf{(recombination depth):}
Number of reference algorithms for \textsc{Combine}. 
We set $m_{\text{comb}}=2$ because each ATLAS \textsc{Combine} prompt includes complete
source programs, so reference count directly trades integration depth against context length.
This choice is motivated by three considerations:
\begin{inparaenum}
\item \textbf{Code complexity:} ATLAS combines complete executable programs, often already
containing hybrid logic from earlier iterations, so additional references increase the amount of
source that must be integrated.
\item \textbf{Integration depth:} with two references, the LLM more reliably performs substantive integration of both algorithms; with $m_{\text{comb}}\geq 3$, preliminary runs often produced outputs dominated by one reference with only shallow incorporation of the others.
\item \textbf{Token efficiency:} two full algorithms provide sufficient recombination context while keeping prompt length manageable.
\end{inparaenum}

\item $\boldsymbol{m_{\text{div}} = 3}$ \textbf{(negative signal sufficiency):}
Number of reference algorithms for \textsc{Diverge}. 
Unlike \textsc{Combine}, \textsc{Diverge} uses references primarily as strategies to avoid rather than as inputs for deep integration, so it benefits from slightly broader context. 
We therefore use $m_{\text{div}}=3$: three references provide a broader negative context
and are intended to encourage outputs separated from the referenced archive regions while keeping
token cost manageable.
Larger values provided little additional benefit in preliminary runs while increasing prompt length.
\end{itemize}

\paragraph{Operator probabilities (Layers~1 \& 2)}

Single-reference refinement operators are applied stochastically to Layer~1 neighbors (excluding $a^*$, which receives deterministic intensification) and Layer~2 cluster representatives. 
These probabilities control both the refinement mix and budget allocation: each operator invocation produces a candidate that must be validated and evaluated, so increasing these probabilities increases refinement pressure but consumes more of the fixed synthesis budget $B$, thereby reducing the number of outer-loop iterations and the budget available for multi-reference search and cross-cluster exploration.

\begin{itemize}[leftmargin=*,itemsep=3pt]

\item $\boldsymbol{p_{\text{imp}} = 0.5,\; p_{\text{tune}} = 0.5}$ 
\textbf{(balanced refinement mix):}
Probabilities for \textsc{Improve} (structural logic refinement) and \textsc{Tune} (parameter-level adjustment). 
We set them equal to allocate comparable pressure to structural refinement and parameter adjustment, which play complementary roles in full-algorithm synthesis. 
Setting both to 0.5 provides strong refinement pressure in Layers~1 and~2 while preserving sufficient budget for outer-loop progress and multi-reference exploration. 
Substantially larger values over-allocate budget to local refinement, whereas substantially smaller values weaken refinement coverage.

\item $\boldsymbol{p_{\text{simp}} = 0.2}$ \textbf{(parsimony under budget constraints):}
Probability for \textsc{Simplify} (code reduction). 
\textsc{Simplify} acts as a regularizer against code bloat during iterative synthesis, but it is not the primary performance-improvement operator. 
We therefore apply it less frequently than \textsc{Improve} and \textsc{Tune}: a value of 0.2 provides periodic parsimony pressure without diverting too much budget away from performance-driven refinement and broader exploration. 
Higher values can over-emphasize simplification and remove useful logic prematurely, whereas much lower values weaken bloat control.
\end{itemize}

\subsubsection{LLM Configuration}
\begin{itemize}[leftmargin=*,itemsep=3pt]
\item \textbf{Model: \textsc{GPT-5-mini} (reasoning effort: LOW):}
We use \textsc{GPT-5-mini} with low reasoning effort because it is capable enough for full-algorithm synthesis while remaining practical in cost and latency for iterative search over hundreds of operator calls.

\item $\boldsymbol{T=1.0}$ \textbf{(temperature):}
Sampling temperature used for all synthesis operators.
We use $T=1.0$ to balance generation diversity and stability during search.
The same value is fixed across all compared LLM-based synthesis methods in our
experiments.
\end{itemize}

\subsubsection{Embedding configuration}
Both the embedding model and input representation were selected based on the
representation analysis in Appendix~\ref{app:embedding_ablation}.
\begin{itemize}[leftmargin=*,itemsep=3pt]
\item \textbf{Model: \textsc{mGTE-large-en-v1.5}~\cite{zhang2024mgte}:}
See Appendix~\ref{app:embedding_strategy} for details.

\item \textbf{Input: early fusion (name + description + preprocessed code):}
Preprocessing details are provided in Appendix~\ref{app:embedding_strategy}.
\end{itemize}

\subsubsection{Search Budget}
\begin{itemize}[leftmargin=*,itemsep=3pt]
\item $\boldsymbol{B = 500}$ \textbf{evaluated operator executions:}
Default ATLAS search budget used for end-to-end component ablations and
search-configuration analyses.
This budget is sufficient to demonstrate effective search behavior while keeping overall runtime and API cost tractable.

\item \textbf{Token-matched budgets for cross-method comparisons:}
For comparisons against baselines, we match total token consumption per synthesis run rather than the number of evaluated operator executions, because token usage differs substantially between component synthesis and full-algorithm synthesis. 
These budgets are calibrated from the average token usage of 500 evaluated operator executions under full synthesis, yielding problem-specific budgets of 5M tokens for FSS, 6.5M for CVRP, 7M for CVRPTW, and 6M for QAP.
\end{itemize}

\paragraph{Evaluation}
\begin{itemize}[leftmargin=*,itemsep=3pt]
\item \textbf{Per-instance runtime caps (problem dependent):}
Maximum execution time allowed for evaluating an algorithm on a single instance. 
These are upper bounds rather than typical runtimes: many synthesized and baseline algorithms terminate well before the cap. 
Full-algorithm synthesis explores algorithms with substantially different computational profiles, ranging from fast constructive methods to iterative metaheuristics, making benchmark-setting-specific runtime caps necessary to accommodate this diversity while maintaining fair comparison.
The caps serve two purposes: 
\begin{inparaenum}
\item  \textbf{Safety:} they terminate pathological implementations (e.g., infinite loops or excessively slow code) and keep the overall experiments tractable; and 
\item  \textbf{comparison consistency:} they ensure that all methods within a benchmark setting are evaluated under the same computational budget. 
\end{inparaenum}

The runtime caps are benchmark-setting-specific and correspond to the default problem settings used in this work (Appendix~\ref{app:instance_generation}): 210\,s for FSS, 30\,s for CVRP, 120\,s for CVRPTW, and 240\,s for QAP.
These values are applied uniformly to all compared methods within each benchmark setting.
These full caps are applied unchanged during training and test execution.

\item $\boldsymbol{N_{\mathrm{train}}=31,\;N_{\mathrm{test}}=31}$ \textbf{(split sizes):}
Search uses only the 31 training instances, and the independently generated 31-instance test set provides the primary estimate. 
Split seeds are 2024 and 42, respectively.

\item $\boldsymbol{m_{\text{limit}} = 1}$ \textbf{GB additional worker address space:}
The released evaluator requires Linux/WSL2 and refuses native Windows execution. In each worker, \texttt{RLIMIT\_AS} is set to the inherited virtual-memory footprint plus 1\,GB, thereby limiting approximately 1\,GB of additional virtual-address-space allocation while one instance is executed. This is a resource-containment allowance, not a claim that resident memory is measured exactly; a violation disqualifies the candidate execution.
\end{itemize}

%% file: Appendix/RepairPrompt.tex
{
\subsubsection{Failure-Conditioned Repair}
\label{app:repair_operator}

\textsc{Repair} is a source-level recovery operator for a generated full algorithm that reaches
training evaluation but fails during execution or evaluator validation.
Candidates rejected before evaluation, such as malformed structured LLM responses or source that
does not pass the entry-point check, do not enter this repair path.

\paragraph{Failure evidence and routing}
For each evaluated candidate, ATLAS records the aggregate error message, an available traceback,
timeout and memory flags, and the numbers of successful and failed instances.
Deterministic rules classify this evidence without an additional LLM diagnosis call.
The classes are checked in the priority order shown in Table~\ref{tab:repair_classes};
the first matching class determines the correction guidance.

\begin{table}[H]
\centering
\caption{Failure classes and correction objectives used by \textsc{Repair}.}
\label{tab:repair_classes}
\small
{
\begin{tabular}{@{}p{0.05\linewidth}p{0.18\linewidth}p{0.31\linewidth}p{0.35\linewidth}@{}}
\toprule
\textbf{Order} & \textbf{Class} & \textbf{Identifying evidence} & \textbf{Correction objective} \\
\midrule
1 & Memory limit & Memory-limit flag or \texttt{MemoryError} & Reduce large materialized structures,
unbounded containers, redundant copies, or deep recursion \\
2 & Partial failure & Some training instances succeed and others fail & Preserve the core approach while
handling edge cases and removing instance-dependent failures \\
3 & Timeout & Timeout marker in the aggregate error & Reduce computational complexity,
redundant work, and excessive iteration within the existing approach \\
4 & Constraint violation & A feasibility keyword in the evaluator message
(Table~\ref{tab:constraint_keywords}) & Correct construction or modification logic so returned
solutions satisfy the problem constraints \\
5 & Runtime error & Exception, error, or traceback markers after excluding earlier classes & Locate and
correct the source-level crash and add relevant defensive checks \\
6 & Framework violation & Remaining unmatched validation or interface failure & Correct the entry point,
signature, return type, output structure, or other interface requirement \\
\bottomrule
\end{tabular}}
\end{table}

Constraint violations are separated from runtime errors because the candidate executes successfully
but returns an invalid or infeasible solution.
This branch uses case-insensitive keyword matching over the aggregate evaluator message.
Table~\ref{tab:constraint_keywords} reproduces the keyword groups used by the implementation.

\begin{table}[H]
\centering
\caption{Keyword groups used to identify constraint-violation failures. Matches are case-insensitive.}
\label{tab:constraint_keywords}
\small
{
\begin{tabular}{@{}p{0.18\linewidth}p{0.73\linewidth}@{}}
\toprule
\textbf{Category} & \textbf{Keywords} \\
\midrule
Capacity and resources & exceeds capacity; capacity constraint; over capacity \\
Completeness & not all customers visited; missing jobs; missing items; missing customers \\
Uniqueness & visited multiple times; duplicate; appears multiple times \\
Permutation and sequence & not a valid permutation; job sequence length; invalid permutation; sequence length \\
General validity & constraint violation; invalid solution; feasibility; infeasible; out of range; invalid index;
invalid customer; invalid item \\
\bottomrule
\end{tabular}}
\end{table}

\paragraph{Source regeneration and acceptance}
The selected class conditions the repair request on the observed failure rather than issuing a
generic debugging instruction.
The LLM receives the problem specification and I/O requirements, the failed algorithm's name,
description, and complete source, the aggregate diagnostic evidence, and class-specific correction
objectives.
It is asked to regenerate the complete algorithm source while retaining useful algorithmic ideas
where appropriate.
The exact executable prompt and guidance strings are provided in the public source-code repository.%
\footnote{\nolinkurl{https://github.com/Danial-Yazdani/ATLAS}}

The regenerated source passes the same code and interface checks and is then re-evaluated from
scratch on the complete training set.
It is admitted to the archive only if every training instance succeeds; otherwise, the repair is
recorded as failed and the candidate is discarded under the reported default one-attempt policy.
This all-or-nothing re-evaluation prevents \textsc{Repair} from weakening feasibility or execution
requirements while recovering promising designs that would otherwise be lost.
}

%% file: Appendix/Baselines.tex
We compare ATLAS against human-designed, domain-specific optimization methods and
LLM-based synthesis baselines.
This section summarizes their implementations and configurations.

%------------------------------------------------------------------------------
\subsubsection{Human-Designed Domain-Specific Baseline Implementations}
\label{app:classical_baselines}
%------------------------------------------------------------------------------

The configurations below identify both the published method and the exact implementation
used. We use documented defaults or fixed upstream example configurations and do not tune a method
using test results.
Runtime caps are common upper bounds rather than typical runtimes: many LLM-synthesized
algorithms terminate well before the cap, whereas runtime-sensitive domain-specific baselines can
continue improving until the allotted time is exhausted. Thus, the caps do not disadvantage these
baselines and may be more beneficial to methods designed to exploit the full runtime budget. Unless
otherwise noted, the runtime-sensitive baselines use 30\,s for CVRP, 120\,s for CVRPTW, 210\,s for
FSS, and 240\,s for QAP in the default settings (Appendix~\ref{app:instance_generation}).

\paragraph{CVRP}
\begin{itemize}[leftmargin=*,itemsep=2pt]
\item \textbf{PyVRP~\cite{wouda2024pyvrp,pyvrp0133software}:} PyVRP v0.13.3 with its
default iterated local search configuration, evaluated under the benchmark-setting-specific
per-instance runtime cap.
\item \textbf{Google OR-Tools~\cite{ortools}:} OR-Tools v9.15.6755 using
\path{PATH_CHEAPEST_ARC} as the first-solution strategy and \path{GUIDED_LOCAL_SEARCH} as the
local-search metaheuristic, evaluated under the benchmark-setting-specific per-instance runtime
cap.
\item \textbf{VROOM~\cite{vroom,vroom116software}:} VROOM v1.16.0-dev, commit
\texttt{07be776}, built with \texttt{USE\_ROUTING=false} and supplied an explicit cost matrix so no
external routing service is involved. It runs single-threaded (\texttt{-t 1}), at exploration level
5 (\texttt{-x 5}), and with the benchmark-setting-specific limit supplied through \texttt{-l}.
\end{itemize}

\paragraph{CVRPTW}
\begin{itemize}[leftmargin=*,itemsep=2pt]
\item \textbf{PyVRP~\cite{wouda2024pyvrp,pyvrp0133software}:} PyVRP v0.13.3 with its
default iterated local search configuration for VRPTW, evaluated under the
benchmark-setting-specific per-instance runtime cap.
\item \textbf{Google OR-Tools~\cite{ortools}:} OR-Tools v9.15.6755 using
\path{PATH_CHEAPEST_ARC} as the first-solution strategy and \path{GUIDED_LOCAL_SEARCH} as the
local-search metaheuristic, evaluated under the benchmark-setting-specific per-instance runtime
cap.
\item \textbf{VROOM~\cite{vroom,vroom116software}:} the same VROOM build and execution
configuration used for CVRP, with capacities, service times, and time windows supplied through the
problem input and every returned route independently checked by the common evaluator.
\end{itemize}

\paragraph{FSS}
\begin{itemize}[leftmargin=*,itemsep=2pt]
\item \textbf{NEH~\cite{nawaz1983heuristic}:} NEH heuristic implementation following the original method, run to completion because it is a deterministic constructive heuristic with polynomial-time complexity and terminates quickly on the default FSS benchmark setting.
\item \textbf{IG-TB with Taillard
acceleration~\cite{ruiz2007simple,taillard1990some,fernandez2014insertion}:} the Ruiz--St\"utzle IG
configuration implemented by \texttt{js-aguiar}~\cite{jsaguiarpfsp}, with the
Fern\'andez-Viagas--Framinan idle-time insertion tie-breaking rule and Taillard's $O(km)$ insertion
evaluation, using Python with Cython kernels.
\item \textbf{Iterative Beam Search~\cite{libralesso2022iterative}:} anytime constructive
tree search using the authors' \textsc{CATS-PFSP} implementation (commit \texttt{41edc0e}), with the
bidirectional variant and \texttt{g4} guidance (\texttt{variant=1}, \texttt{guide=3}) fixed from the
upstream example.
\end{itemize}

\paragraph{QAP}
\begin{itemize}[leftmargin=*,itemsep=2pt]
\item \textbf{Robust Tabu Search (RoTS)~\cite{taillard1991robust}:} RoTS implementation following the original method, evaluated under the benchmark-setting-specific per-instance runtime cap.
\item \textbf{Simulated Annealing~\cite{connolly1990improved}:} SA implementation following the original method, evaluated under the benchmark-setting-specific per-instance runtime cap.
\item \textbf{Breakout Local Search (BLS)~\cite{benlic2013breakout}:} iterated local
search alternating steepest-descent improvement with adaptive perturbations, using Benlic and Hao's
released source as integrated in the public ISA-QAP-Algorithms collection~\cite{isaqapalgorithms}.
The integration accepts an external RNG seed and runtime cap and exposes the returned permutation
for independent scoring.
\item \textbf{Memetic Search (BMA)~\cite{benlic2015memetic}:} population-based search
integrating BLS, crossover, pool updating, and adaptive mutation, using the implementation
distributed in the same ISA-QAP-Algorithms collection~\cite{isaqapalgorithms}. The integration
supplies the external seed and cap and retains a self-verified best returned permutation so the
common evaluator can recompute its objective.
\end{itemize}

%------------------------------------------------------------------------------
\subsubsection{Component-Synthesis Baseline Implementations}
\label{app:llm_baselines}
%------------------------------------------------------------------------------

We compare against component-oriented implementations of three LLM-based methods:
\textsc{ReEvo}~\cite{ye2024reevo}, \textsc{EoH}~\cite{liu2024evolution}, and
MCTS-AHD~\cite{zheng2025monte}.
Their official repositories target different problems and use non-uniform scaffolds, prompts, and evaluators, making a controlled end-to-end comparison under their native setups difficult to interpret.\footnote{\textsc{ReEvo}: \url{https://github.com/ai4co/reevo}, \textsc{EoH}: \url{https://github.com/FeiLiu36/EoH}, MCTS-AHD: \url{https://github.com/zz1358m/MCTS-AHD-master}} 
We therefore evaluate all three methods using the LLM4AD platform~\cite{liu2024llm4ad},\footnote{LLM4AD (v1.0.0): \url{https://github.com/Optima-CityU/llm4ad}} which provides a shared scaffold, prompts, and evaluators.

\paragraph{Shared benchmark interface}
All three component-synthesis baselines are evaluated under the same LLM4AD-based component-execution and benchmarking infrastructure:
\begin{itemize}[leftmargin=*,itemsep=2pt]
\item \textbf{Common scaffold:} all methods synthesize heuristic components within the same LLM4AD greedy-construction scaffold.
\item \textbf{Common prompts and interface specification:} all methods use the same LLM4AD problem prompts and interface specification under this scaffold, so differences are not attributable to prompt wording or scaffold design.
\item \textbf{Common evaluators:} all methods use the same LLM4AD-based execution and evaluation pipeline, including timeout handling, feasibility handling, objective computation, and reporting.
\item \textbf{Common benchmark instances:} all methods use the same 31 training instances and 31 test instances for each default benchmark setting, with fixed seeds. For CVRP, CVRPTW, and QAP, we use the LLM4AD benchmark generators, which are also used by ATLAS. 
For FSS, LLM4AD's default Co-Bench setup is replaced with the standard generator used throughout this paper (Appendix~\ref{app:fss_generation}) so that all methods are evaluated on the same benchmark setting.
\end{itemize}

\paragraph{Method-specific preservation and controlled adaptations}
Within this shared interface, we retain the default search mechanisms and parameter settings of the official implementations as closely as possible (e.g., selection and population-management logic), modifying only evaluation-facing components needed for fair comparison:
\begin{itemize}[leftmargin=*,itemsep=2pt]
\item \textbf{Termination criterion:} all methods use the same benchmark-specific token budgets as ATLAS, namely 5M tokens for FSS, 6.5M for CVRP, 7M for CVRPTW, and 6M for QAP.
\item \textbf{Per-instance runtime caps:} all methods use the same benchmark-setting-specific per-instance runtime caps as the rest of the paper, namely 210\,s for FSS, 30\,s for CVRP, 120\,s for CVRPTW, and 240\,s for QAP.
\item \textbf{LLM backend:} all methods use \textsc{GPT-5-mini} with temperature $T=1.0$ and reasoning effort set to \textit{low}.
\end{itemize}

This standardization reduces confounding from scaffold design, prompts, evaluators,
benchmark instances, and stopping criteria when comparing the three search procedures in their
tested component-oriented implementations.

%------------------------------------------------------------------------------
\subsubsection{\textsc{EoH-Full}: Full-Algorithm Synthesis Baseline}
\label{app:eoh_full}
%------------------------------------------------------------------------------

To isolate the contribution of ATLAS's search organization, we construct
\textsc{EoH-Full}, a controlled full-synthesis baseline derived from
\textsc{EoH}~\cite{liu2024evolution}. \textsc{EoH-Full} retains \textsc{EoH}'s fitness-driven
evolutionary search framework while using ATLAS's problem- and interface-aware full-algorithm
setting. By matching synthesis operators, prompts, repair, evaluator, benchmark instances, LLM
configuration, runtime caps, and token budget, this comparison is designed to reduce confounding
from operator expressiveness and evaluation conditions when assessing the archive-based semantic
quality-diversity search.

\paragraph{Formulation}
\textsc{EoH-Full} adapts \textsc{EoH} from component synthesis to full-algorithm synthesis. 
It preserves the evolutionary search structure of the original method, including its fitness-driven selection, population-update logic, population size, initialization scheme, and operator probabilities, but replaces component-level generation with the same full-synthesis operators used in ATLAS:
\textsc{Create}, \textsc{Improve}, \textsc{Tune}, \textsc{Simplify}, \textsc{Combine}, \textsc{Diverge}, and \textsc{Repair}. 
In particular, \textsc{EoH-Full} keeps the original \textsc{EoH} population size of 20 rather than matching ATLAS's archive cap $N_{\max}=100$, since the latter is a bounded semantic archive rather than an active evolutionary population. 
It also uses \textsc{EoH}'s original initialization scheme rather than ATLAS's two-phase initialization.

\paragraph{Key differences from ATLAS}
\begin{itemize}[leftmargin=*,itemsep=2pt]
\item \textbf{Reference selection strategy:} 
\textsc{EoH-Full} selects parent/reference algorithms through fitness-driven evolutionary selection (rank-based proportional selection), whereas ATLAS selects references through semantic organization and three-layer search, using embedding neighborhoods, cluster representatives, and cross-cluster combinations to control where search effort is allocated.

\item \textbf{Population vs.\ semantic archive:} 
\textsc{EoH-Full} maintains a fixed-size active evolutionary population, with offspring inserted and survival determined through fitness-driven selection each generation. 
This setup is primarily geared toward improving the currently strongest-performing lineage. 
In contrast, ATLAS maintains an embedding-organized archive with coverage-preserving
management, enabling it to retain and continue refining competitive algorithms in multiple
embedding-space regions rather than relying only on global fitness rank.

\item \textbf{Diversity dynamics and search outcome:}
\textsc{EoH-Full} uses fitness-driven survival, which may concentrate sampling around
high-performing lineages and is primarily oriented toward identifying a single best final algorithm.
ATLAS instead applies embedding-distance-aware archive management and continues refining multiple
represented regions throughout search. As a result, ATLAS can retain multiple competitive algorithms
from distinct regions of the archive and return the corresponding cluster representatives as a diverse
set of candidate algorithms, in addition to the overall best algorithm.

\end{itemize}

\paragraph{Shared components for controlled comparison}
To ensure a controlled comparison, \textsc{EoH-Full} shares the following components with ATLAS:
\begin{itemize}[leftmargin=*,itemsep=2pt]
\item \textbf{Full-synthesis operator tasks:} \textsc{EoH-Full} uses the same full-synthesis operator responsibilities as ATLAS, namely \textsc{Create}, \textsc{Improve}, \textsc{Tune}, \textsc{Simplify}, \textsc{Combine}, \textsc{Diverge}, and \textsc{Repair}, including the same multi-reference settings ($m_{\text{comb}}=2$ for \textsc{Combine} and $m_{\text{div}}=3$ for \textsc{Diverge}). 
\item \textbf{Repair and evaluation:} the same candidate repair mechanism and execution/evaluation pipeline (Appendix~\ref{app:evaluation}), including the same benchmark-setting-specific per-instance runtime caps
\item \textbf{LLM configuration:} \textsc{GPT-5-mini} with temperature $T=1.0$ and reasoning effort set to \textit{low}
\item \textbf{Benchmark instances:} the same 31 training instances and 31 test instances with identical random seeds
\item \textbf{Termination criterion:} the same benchmark-specific token budgets used for cross-method comparison
\end{itemize}

\paragraph{EoH-specific preserved mechanisms}
Within this shared full-synthesis setting, \textsc{EoH-Full} keeps the method-specific evolutionary mechanisms of the original \textsc{EoH} implementation, including its selection procedure and operator probability schedule. 
The original \textsc{EoH} operator groups map naturally onto the ATLAS full-synthesis operator set: modification operators $\mathbf{M_1,M_2,M_3}$ correspond to \textsc{Improve}, \textsc{Tune}, and \textsc{Simplify}, while exploration operators $\mathbf{E_1,E_2}$ correspond to \textsc{Diverge} and \textsc{Combine}, respectively. 
Thus, the operator responsibilities are the same as in ATLAS. 
\textsc{Repair} has no direct counterpart in the original \textsc{EoH} schedule; however, similar to ATLAS, it is applied only when a generated candidate fails validation. 
This shared use of \textsc{Repair} addresses the additional execution, interface, and
feasibility failure modes exposed by complete-program generation and avoids giving ATLAS a recovery
mechanism unavailable to \textsc{EoH-Full}. Under the shared full-synthesis machinery and evaluation
conditions described above, the comparison is designed to estimate the effect of replacing
\textsc{EoH}'s fitness-driven evolutionary search with ATLAS's semantic quality-diversity search.

%% file: Appendix/ablation_embedding.tex
\subsubsection{Aim and Research Questions}

\paragraph{Aim}
ATLAS uses pretrained embeddings as the operational representation for k-nearest-neighbor
retrieval, clustering, deduplication, and similarity-based pruning. Unlike the fitness- and
tree-statistic baselines in this analysis, this representation directly determines archive
neighborhoods and coverage management.
This study evaluates which embedding strategies are most effective for organizing algorithms in a way that 
supports synthesis search in combinatorial optimization.

\paragraph{Research questions}
We study four design questions:
\begin{enumerate}[label=\textbf{RQ\arabic*:}, leftmargin=*, labelindent=30pt, itemsep=3pt]
    \item Author-labeled family discrimination: To what extent do pretrained
embeddings distinguish the broad algorithm-family labels assigned in the curated benchmark (e.g.,
greedy, local search, and metaheuristic-style methods)?
    \item Modality contribution: Which information source contributes most to embedding quality: textual descriptions, source code, or both?
    \item Fusion strategy: If multiple modalities are beneficial, should they be combined through early fusion (concatenation) or late fusion (weighted similarity)?
    \item Model selection: How do encoder characteristics, such as size, specialization, and availability, affect performance?
\end{enumerate}

These analyses provide empirical justification for the embedding design used in ATLAS.

\subsubsection{Dataset Construction and Family Annotation}
\label{app:dataset}

To evaluate embedding quality under controlled conditions, we constructed a curated benchmark of executable algorithm implementations. 
The dataset comprises 308 algorithms across four combinatorial optimization problems: Flow Shop Scheduling (FSS), Capacitated Vehicle Routing Problem (CVRP), Capacitated Vehicle Routing Problem with Time Windows (CVRPTW), and Quadratic Assignment Problem (QAP). 
This selection spans routing (CVRP, CVRPTW), scheduling (FSS), and assignment (QAP), providing diverse algorithmic structure across the problem set.

\paragraph{Algorithm Generation}
For each problem, we independently generated a large candidate pool of algorithm implementations using multiple LLM backends and varied prompting strategies with high-temperature sampling to induce implementation diversity.
Each candidate was then validated for syntactic correctness and functional execution on problem-specific benchmark instances, yielding more than 2,000 validated algorithms across the four problems.
From this validated pool, we selected a balanced subset across problems and broad
author-assigned mechanism labels, while preserving implementation diversity within each label and
excluding near-duplicate variants.

\paragraph{Why curated evaluation is needed}
Embedding evaluation in this setting requires two properties:
\begin{inparaenum}
    \item \textit{inter-label separation}: algorithms should cover different
author-assigned broad families, and
    \item \textit{intra-label diversity}: algorithms assigned to the same family should
differ in implementation choices rather than only superficial code style.
\end{inparaenum}
A curated construction process is therefore useful because naively collecting
implementations from public sources would make it difficult to control either the breadth of the
author-assigned labels or the degree of meaningful implementation variation within each label.
Our curation procedure was designed to provide broad label coverage and non-trivial
implementation diversity for embedding evaluation. Family labels were assigned internally by the
authors from the code, generated description, and recognized search logic.

\subsubsection{Candidate Models and Selection Rationale}
\label{app:models}

We evaluate 10 pretrained embedding models chosen to cover a diverse range of training focuses, model scales, access modes, and context capacities. 
The candidate set includes both general-purpose embedding models and code-specialized encoders, allowing us to examine how these characteristics affect algorithm representation quality in our setting. Table~\ref{tab:embedding_models_specs} summarizes the model specifications.

\begin{table}[t]
\centering
\caption{\emph{Embedding Model Specifications.} Models are grouped into general-purpose encoders and code-specialized encoders.}
\label{tab:embedding_models_specs}
\small
\begin{tabular}{@{}llccc@{}}
\toprule
\textbf{Model} & \textbf{Access} & \textbf{Params} & \textbf{Dim.} & \textbf{Context} \\
\midrule
\multicolumn{5}{l}{\textit{General-Purpose Models}} \\
\quad \textsc{text-embedding-3-large (OpenAI)}~\cite{openai2024embeddings} & Proprietary & -- & 3072 & 8K \\
\quad \textsc{text-embedding-3-small (OpenAI)}~\cite{openai2024embeddings} & Proprietary & -- & 1536 & 8K \\
\quad \textsc{Qwen3-Embedding-0.6B}~\cite{qwen3embedding} & Open & 0.6B & 1024 & 32K \\
\quad \textsc{Qwen3-Embedding-4B}~\cite{qwen3embedding} & Open & 4.0B & 2560 & 32K \\
\quad \textsc{mGTE-large-en-v1.5}~\cite{zhang2024mgte} & Open & 434M & 1024 & 8K \\
\midrule
\multicolumn{5}{l}{\textit{Code-Specialized Models}} \\
\quad \textsc{Nomic-Embed-Code}~\cite{suresh2025cornstack} & Open & 7.0B & 768 & 32K \\
\quad \textsc{CodeRankEmbed}~\cite{suresh2025cornstack} & Open & 137M & 768 & 8K \\
\quad \textsc{codebert-base}~\cite{feng2020codebert} & Open & 125M & 768 & 512 \\
\quad \textsc{GraphCodeBERT-base}~\cite{guo2021graphcodebert} & Open & 125M & 768 & 512 \\
\quad \textsc{CodeT5-small}~\cite{wang2021codet5} & Open & 60M & 512 & 512 \\
\bottomrule
\end{tabular}
\end{table}

\paragraph{Selection Rationale}
The selected models span a broad range of characteristics that may affect embedding behavior, including training focus, scale, access mode, and context capacity. 
This diversity allows us to compare general-purpose and code-specialized encoders, assess whether larger models consistently yield stronger representations, and examine the effect of limited versus long input contexts when representing full synthesized algorithms.

Including both proprietary and open-source models also lets us benchmark against strong commercial embeddings while identifying configurations that remain reproducible with publicly available encoders. 
In particular, the variation in context capacity is important because some source-code inputs exceed the limits of shorter-context models, making this a relevant factor in the design space.

\subsubsection{Code Preprocessing}
\label{app:code_preprocess}

All source code is preprocessed before embedding to reduce implementation noise while retaining the core program structure.
The preprocessing pipeline applies the following transformations:

\begin{enumerate}[leftmargin=*, itemsep=2pt]
    \item \textbf{Docstring Removal:} Triple-quoted strings (\texttt{"""..."""} and \texttt{```...'''}) are removed, as they contain natural language documentation that often overlaps with the algorithm description.
    \item \textbf{Comment Removal:} Single-line comments (starting with \texttt{\#}) are removed while preserving \texttt{\#} characters within string literals.
    \item \textbf{Empty Line Removal:} Blank lines are removed as formatting artifacts without semantic content.
    \item \textbf{Whitespace Normalization:} Multiple consecutive spaces are converted to single spaces while preserving indentation structure.
    \item \textbf{Type Hint Preservation:} Type annotations are retained because they encode structural interface information rather than natural-language description, complementing rather than duplicating the algorithm description field.
\end{enumerate}

This pipeline achieves an average 23\% character reduction across the corpus. 
For this representation analysis, preprocessing helps reduce the influence of
natural-language artifacts in code inputs, making the comparison between text-based and code-based
representations cleaner.
The same preprocessing is applied throughout both the embedding representation analysis
and the main ATLAS system.

\subsubsection{Embedding Strategies}
\label{app:strategies}

We evaluate four embedding strategies that differ in which algorithm components are represented and how the resulting views are combined. 
Each algorithm is represented as $a = (n, d, c)$, comprising a name $n$, description $d$, and source code $c$. 
All code-based strategies use preprocessed source code as described in Appendix~\ref{app:code_preprocess}.

\paragraph{Text-Only}
This baseline embeds only the textual components, namely the algorithm name and description:
\begin{equation}
\text{sim}_{\text{text}}(a_i, a_j) = \cos(\mathbf{e}(n_i \oplus d_i), \mathbf{e}(n_j \oplus d_j)),
\end{equation}
where $\mathbf{e}(\cdot)$ denotes the embedding model applied to the corresponding input
under the corresponding strategy, and $\oplus$ denotes concatenation. This strategy tests whether
textual descriptions alone provide enough information to distinguish the curated broad-family
labels.

\paragraph{Code-Only}
This strategy embeds only the source code implementation:
\begin{equation}
\text{sim}_{\text{code}}(a_i, a_j) = \cos(\mathbf{e}(c_i), \mathbf{e}(c_j)).
\end{equation}
This strategy tests whether preprocessed source code alone provides enough information to
distinguish the curated broad-family labels.

\textbf{Concatenation (Early Fusion):}
Early fusion concatenates all available components into a single input:
\begin{equation}
\label{eq:concatSimilarity}
\text{sim}_{\text{concat}}(a_i, a_j) = \cos(\mathbf{e}(n_i \oplus d_i \oplus c_i), \mathbf{e}(n_j \oplus d_j \oplus c_j)).
\end{equation}
This strategy tests whether a single embedding model can jointly represent textual intent and implementation structure when both are presented in one sequence.

\paragraph{Dual-View (Late Fusion)}
Late fusion computes separate similarities for text and code, then combines them linearly:
\begin{equation}
\label{eq:similarity_dual}
\begin{aligned}
\text{sim}_{\text{dual}}(a_i, a_j; \alpha)
    &= \alpha \cdot \cos(\mathbf{e}(n_i \oplus d_i), \mathbf{e}(n_j \oplus d_j)) \\
    &\quad + (1-\alpha) \cdot \cos(\mathbf{e}(c_i), \mathbf{e}(c_j)),
\end{aligned}
\end{equation}
where $\alpha \in [0,1]$ controls the relative contribution of the text and code views. We sweep $\alpha \in \{0.0, 0.1, 0.2, \ldots, 1.0\}$ to examine the sensitivity of late fusion to the weighting between the two views. This strategy tests whether separating text and code and combining them at the similarity level is more effective than representing them through a single unified input.

\subsubsection{Evaluation Metrics}
\label{app:metrics}

We employ two complementary metrics to assess agreement with the curated author-assigned
labels: Nearest Neighbor Accuracy (NNA) and Class Cohesion Index (CCI), a silhouette-based score.
Both metrics are computed for each embedding strategy defined in Appendix~\ref{app:strategies}.

\paragraph{Nearest Neighbor Accuracy (NNA)}
For each algorithm $a_i$, we identify its nearest neighbor in the embedding space:
\begin{equation}
a_i^{\text{NN}} = \argmax_{j \neq i} \text{sim}(a_i, a_j),
\end{equation}
where $\text{sim}(\cdot, \cdot)$ denotes the similarity function defined by the strategy
(e.g., $\text{sim}_{\text{text}}$, $\text{sim}_{\text{dual}}$), and $a_i^{\text{NN}}$ is the nearest
algorithm to $a_i$. NNA measures the fraction whose nearest neighbor shares the same author-assigned
family label:
\begin{equation}
\text{NNA} = \frac{1}{N} \sum_{i=1}^{N} \mathbb{1}\left[ y(a_i) = y(a_i^{\text{NN}}) \right],
\end{equation}
where $N$ is the total number of algorithms, $y(a_i)$ denotes the author-assigned family
label of algorithm $a_i$, and $\mathbb{1}[\cdot]$ is the indicator function. This metric measures
local agreement between the embedding and those labels.

\paragraph{Class Cohesion Index (CCI)}
To complement nearest-neighbor agreement, we measure continuous separation using a
silhouette-based cohesion score computed from the author-assigned labels. CCI follows the standard
Silhouette Coefficient formulation~\cite{rousseeuw1987silhouettes}, but uses the curated labels
rather than unsupervised clusters. For each algorithm $a_i$ assigned to family $C_k$, we compute:
\begin{equation}
\text{cohesion}_i = \frac{d_{\text{between}}^i - d_{\text{within}}^i}{\max(d_{\text{within}}^i, d_{\text{between}}^i)},
\end{equation}
where:
\begin{itemize}[leftmargin=*,labelindent=20pt,itemsep=1pt]
\item $d_{\text{within}}^i = \frac{1}{|C_k|-1} \sum_{j \in C_k, j \neq i} d(a_i, a_j)$ is the mean distance to other algorithms in the same family (within-family distance).
\item $d_{\text{between}}^i = \min_{C_\ell \neq C_k} \frac{1}{|C_\ell|} \sum_{j \in C_\ell} d(a_i, a_j)$ is the mean distance to the nearest different family (between-family distance).
\end{itemize}
Here, $d(a_i, a_j) = 1 - \text{sim}(a_i, a_j)$ converts similarity to distance, and $C_k$ denotes the author-assigned family containing $a_i$.
 The overall Class Cohesion Index is the mean across all $N$ algorithms:
\begin{equation}
\label{eq:CCI}
\text{CCI} = \frac{1}{N} \sum_{i=1}^{N} \text{cohesion}_i.
\end{equation}
Values range from $-1$ to $+1$, with larger values indicating tighter grouping and clearer
separation with respect to the curated labels.

\subsubsection{Results}
\label{app:results}

\begin{figure}[!ht]
  \centering
  \begin{subfigure}{0.90\linewidth}
    \centering
    \includegraphics[width=\linewidth,height=0.4\textheight,keepaspectratio]{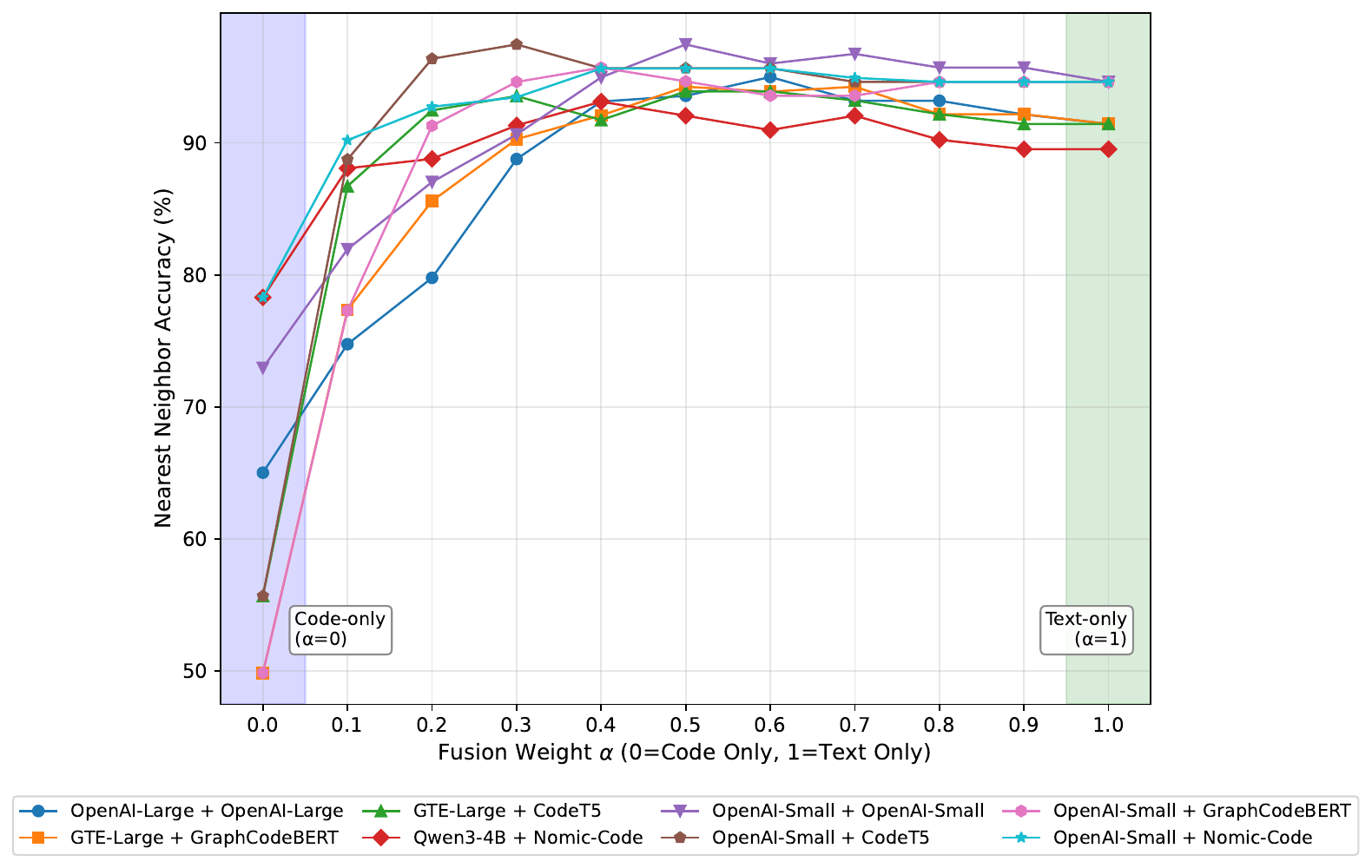}
    \caption{NNA vs. Weighting ($\alpha$)}
    \label{fig:nna_sweep}
  \end{subfigure}

  \vspace{4pt}

  \begin{subfigure}{0.90\linewidth}
    \centering
    \includegraphics[width=\linewidth,height=0.4\textheight,keepaspectratio]{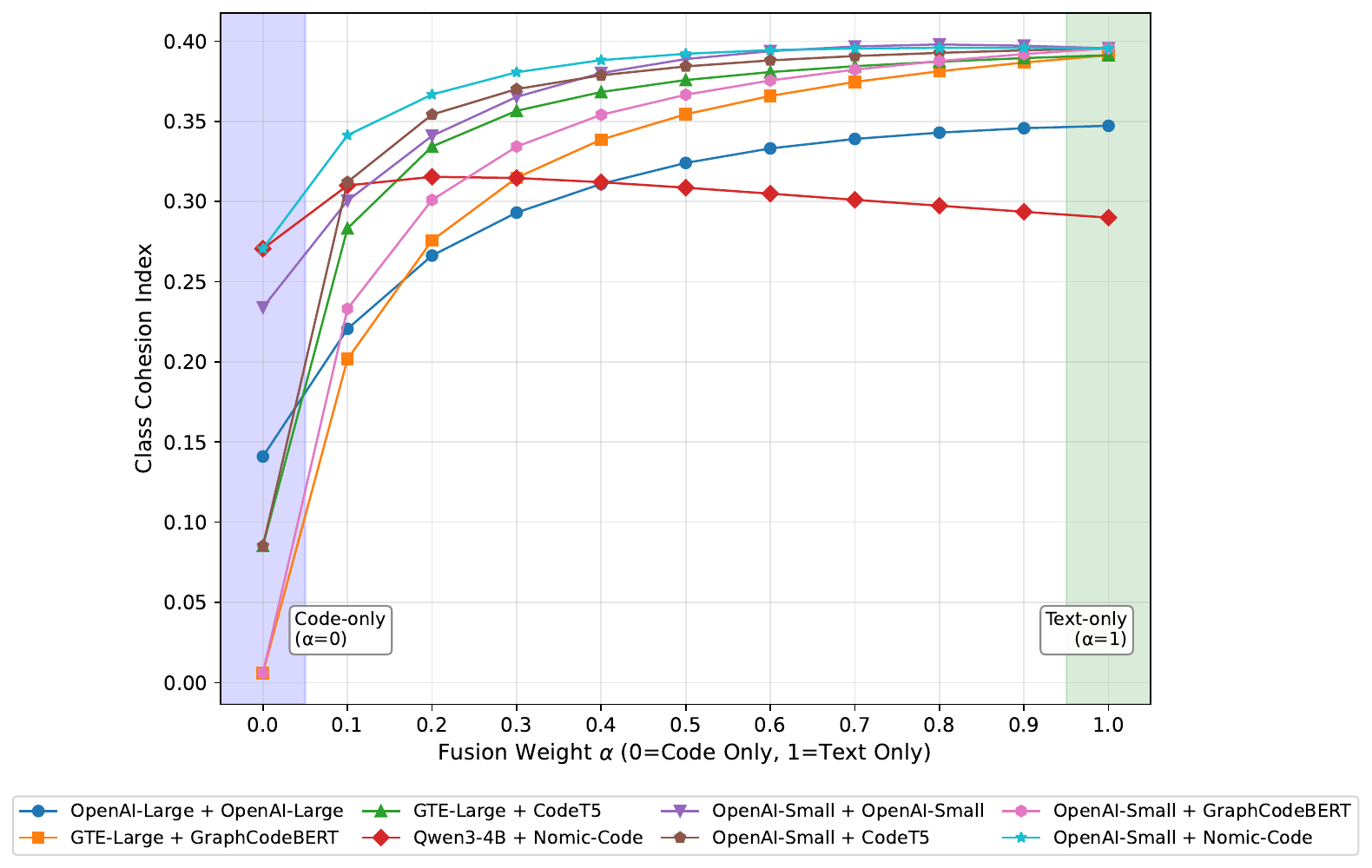}
    \caption{CCI vs. Weighting ($\alpha$)}
    \label{fig:cci_sweep}
  \end{subfigure}
    \caption{\emph{Dual-View (Late Fusion) Sensitivity Analysis.} We sweep the mixing parameter $\alpha$ from 0.0 (code-only) to 1.0 (text-only).}

    \label{fig:dual_view_sweep}
\end{figure}

We evaluated all embedding strategies across the candidate models. 
Table~\ref{tab:single_encoder_results} summarizes the performance of the text-only, code-only, and concatenation strategies. 
Figure~\ref{fig:dual_view_sweep} shows late-fusion performance across fusion weights $\alpha \in [0.0, 1.0]$ for eight representative model pairs.

\begin{table*}[h]
\centering
\caption{\emph{Single-Encoder Strategy Results.} Nearest Neighbor Accuracy (NNA) and Class Cohesion Index (CCI) for text-only, code-only, and concatenation strategies across all models. 
General-purpose models are evaluated on all three strategies, while code-specialized models are evaluated only in the code-only setting. 
\textbf{Bold} indicates the best NNA performance within each strategy. Higher is better for both metrics.}
\label{tab:single_encoder_results}
\footnotesize
\setlength{\tabcolsep}{4.5pt}
\begin{tabular}{@{}lccccccc@{}}
\toprule
& \multicolumn{2}{c}{\textbf{Text-Only}} & \multicolumn{2}{c}{\textbf{Code-Only}} & \multicolumn{2}{c}{\textbf{Concatenation (Early Fusion)}} \\
\cmidrule(lr){2-3} \cmidrule(lr){4-5} \cmidrule(lr){6-7}
\textbf{Model Architecture} & \textbf{NNA} (\%)$\uparrow$ & \textbf{CCI}$\uparrow$ & \textbf{NNA} (\%)$\uparrow$ & \textbf{CCI}$\uparrow$ & \textbf{NNA} (\%)$\uparrow$ & \textbf{CCI}$\uparrow$ \\
\midrule
\multicolumn{7}{l}{\textit{General-Purpose Models}} \\
\quad \textsc{text-embedding-3-small (OpenAI)} & \textbf{94.62} & 0.40 & 73.32 & 0.23 & 91.33 & 0.32 \\
\quad \textsc{text-embedding-3-large (OpenAI)} & 91.44 & 0.35 & 65.38 & 0.14 & 86.62 & 0.25 \\
\quad \textsc{Qwen3-Embedding-0.6B} & 88.77 & 0.29 & 75.81 & 0.22 & 91.80 & 0.31 \\
\quad \textsc{Qwen3-Embedding-4B} & 89.53 & 0.29 & 77.09 & 0.21 & 95.33 & 0.39 \\
\quad \textsc{mGTE-large-en-v1.5} & 91.44 & 0.39 & 72.99 & 0.23 & \textbf{95.38} & 0.44 \\
\midrule
\multicolumn{7}{l}{\textit{Code-Specialized Models}} \\
\quad \textsc{Nomic-Embed-Code} & --- & --- & \textbf{78.29} & 0.27 & --- & --- \\
\quad \textsc{CodeRankEmbed} & --- & --- & 63.87 & 0.21 & --- & --- \\
\quad \textsc{codebert-base} & --- & --- & 54.36 & 0.02 & --- & --- \\
\quad \textsc{GraphCodeBERT-base} & --- & --- & 49.83 & 0.01 & --- & --- \\
\quad \textsc{CodeT5-small} & --- & --- & 55.66 & 0.09 & --- & --- \\
\bottomrule
\end{tabular}
\end{table*}

\subsubsection{Analysis and Interpretation}
\label{app:analysis}

The results in Table~\ref{tab:single_encoder_results} and Figure~\ref{fig:dual_view_sweep} reveal several patterns that inform the embedding configuration used in ATLAS.

\paragraph{Text embeddings agree strongly with the curated broad-family labels}
Text-only strategies achieve 89--95\% NNA across all tested general-purpose models, substantially outperforming code-only approaches (50--78\% NNA). 
This pattern is consistent across diverse model families, with the strongest text-only configuration (\textsc{text-embedding-3-small}, 94.62\%) exceeding the best code-specialized encoder (\textsc{Nomic-Embed-Code}, 78.29\%) by more than 16 percentage points. 
Overall, these results show that the algorithm name and description provide a strong
signal for reproducing the author-assigned broad-family labels in this curated benchmark.

\paragraph{Model scale does not guarantee stronger representations}
Smaller models match or outperform larger ones in several cases. 
Within the OpenAI family, \textsc{text-embedding-3-small} exceeds \textsc{text-embedding-3-large} across all three single-encoder strategies. 
Similarly, \textsc{mGTE-large-en-v1.5} achieves the best concatenation result (95.38\% NNA) while remaining far smaller than \textsc{Qwen3-Embedding-4B}, which performs comparably (95.33\%). 
These results suggest that, for this task, embedding quality depends on more than parameter count alone. 
One possible explanation is that smaller models may produce representations that are less sensitive to superficial lexical variation, although this mechanism is not directly tested here.

\paragraph{The benefit of concatenation is model-dependent}
The effect of early fusion varies substantially across encoders. 
For \textsc{mGTE-large-en-v1.5} and both Qwen models, concatenation improves over text-only performance (mGTE: +3.94pp, Qwen3-4B: +5.80pp, Qwen3-0.6B: +3.03pp). 
In contrast, both OpenAI models degrade under concatenation (OpenAI-Small: -3.29pp, OpenAI-Large: -4.82pp). 
This shows that combining text and code in a single input is not uniformly beneficial; its effectiveness depends on the embedding model.

\paragraph{Code contributes complementary information for compatible encoders}
Although code-only representations are weaker than text-only baselines, source code can still improve representation quality when combined with text in compatible models. 
For example, concatenation improves both NNA and CCI for \textsc{mGTE-large-en-v1.5} and the Qwen models. 
This suggests that code can provide useful disambiguating information beyond the textual description, especially when the encoder can integrate mixed text-code inputs effectively.

\paragraph{Late fusion offers high performance at greater complexity}
Late-fusion strategies can achieve very strong results. 
The strongest configuration, using \textsc{text-embedding-3-small} for both the text and code views, reaches 97.40\% NNA at $\alpha = 0.6$, and performance remains above 96\% over a relatively broad range of $\alpha$ values. 
However, these gains come with added complexity: late fusion requires separate embeddings for text and code, maintains two representation views, and introduces the mixing weight $\alpha$ as an additional hyperparameter. 
This makes late fusion less convenient to deploy than a single-encoder alternative, especially when the performance gap is small.

\subsubsection{Configuration Selection: Balancing Performance and Cohesion}
For the main ATLAS experiments, we select \textbf{\textsc{mGTE-large-en-v1.5} with concatenation}. 
While \textsc{text-embedding-3-small} with text-only achieves competitive NNA (94.62\%), and late-fusion configurations reach higher peak accuracy (97.40\%), \textsc{mGTE-large-en-v1.5} with concatenation provides the most practical balance across multiple criteria:

\begin{itemize}[leftmargin=*, itemsep=2pt]
\item \textbf{Strong Label Agreement:} 95.38\% NNA is the best result among the
single-encoder strategies, with only a 2pp gap to the strongest late-fusion configuration.
\item \textbf{Strong Cohesion Relative to the Curated Labels:} CCI of 0.437 is the highest
among all strategies. Higher cohesion indicates tighter grouping and clearer separation relative to
those labels, which supports the selected representation for neighborhood and clustering operations
on the curated benchmark but does not independently validate evolved archive clusters.
\item \textbf{Architectural Simplicity:} Single-encoder concatenation requires one embedding call per algorithm rather than two, reducing computational overhead and eliminating the need to tune a fusion weight.
\item \textbf{Accessibility:} \textsc{mGTE-large-en-v1.5} is open-source and moderate in size, improving reproducibility without API dependencies or large specialized code models.
\end{itemize}

Although the strongest late-fusion configuration achieves slightly higher peak NNA, the margin over \textsc{mGTE-large-en-v1.5} with concatenation is small. 
Given its superior CCI and strong NNA performance, together with its simplicity and accessibility, \textsc{mGTE-large-en-v1.5} with concatenation is the embedding configuration used in ATLAS.

%% file: Appendix/ablation_initialization.tex
\subsubsection{Aim and Research Question}

\paragraph{Aim}
A key challenge in full algorithm synthesis is to generate a diverse and valid initial archive. 
Prior work in LLM-based code generation has explored prompt variation, including persona-based prompting~\cite{dat2025hsevo}, as a way to induce diversity in generated outputs. 
However, it remains unclear whether such prompt-level variation is sufficient to produce
broad embedding-space coverage in full-algorithm synthesis. This study evaluates whether ATLAS's
dual-phase initialization, which combines zero-reference generation with reference-based divergence,
yields a more dispersed set of valid initial algorithms under the selected embedding metrics than
prompt variation alone.

\paragraph{Research question}
Does ATLAS's dual-phase initialization yield a more diverse set of valid initial algorithms than prompt variation alone?

\subsubsection{Setup}
We evaluate three initialization strategies across four LLM architectures: \textsc{GPT-4.1-Mini}, \textsc{GPT-5 mini} (configured with low reasoning effort), \textsc{Claude Haiku 4.5}, and \textsc{Llama 4 Maverick}. 
The study is conducted on two combinatorial optimization domains: CVRP (50 customers, 31 instances) and FSS (50 jobs, 10 machines, 31 instances). 
These are non-trivial synthesis settings in which generated algorithms may fail due to invalid logic, infeasibility, or execution errors, making them suitable for evaluating both diversity and validity during initialization. 
Evaluating across multiple models and problem types helps reduce the chance that the observations are tied to a single architecture or domain.

Each configuration generates 50 algorithms per trial with temperature $T=1.0$. 
Experiments are repeated across 5 independent trials, and all reported results are averaged across trials. 
The three initialization strategies are:

\begin{enumerate}[leftmargin=*,itemsep=2pt]
    \item \textbf{Baseline (Stochastic Only):} All 50 algorithms are generated via the \textit{Create} operator using a fixed default persona (Table~\ref{tab:personas}). 
    Diversity arises only from sampling stochasticity.
    
    \item \textbf{+Random Persona:} All 50 algorithms are generated via \textit{Create}, with the persona randomly sampled from the full persona library (Table~\ref{tab:personas}) for each generation. 
    This tests whether prompt-level variation can improve the diversity of the initial algorithms.
    
    \item \textbf{+Divergence:} This corresponds to ATLAS's dual-phase initialization strategy. 
    Following the default ATLAS configuration, 20 algorithms are first generated via \textit{Create} using the default persona to form an initial reference pool. 
    Then, 30 additional algorithms are generated incrementally via the \textit{\textsc{Diverge}} operator, also using the default persona. 
    At each step, 3 references are sampled from the current pool and passed to the \textsc{Diverge} operator, which generates a new algorithm intended to be distinct from those references; the resulting valid algorithm is then added back to the pool.
\end{enumerate}

\paragraph{Persona Design}
\label{app:personas}

For this initialization analysis, we constructed a diverse library of prompt personas
spanning several stylistic and design-oriented preferences, including philosophical stance (e.g.,
modernist, traditionalist, exotic), optimization priority (e.g., efficiency, robustness,
scalability), and complexity preference (e.g., simplicity, sophistication).
Table~\ref{tab:personas} lists the full persona library used in the \textbf{+Random Persona}
configuration.

\begin{table}[h]
\centering
\caption{Persona library used in the initialization-strategy analysis. The
\texttt{default} persona is used in ATLAS; the remaining personas are used to evaluate prompt-level
variation during initialization.}
\label{tab:personas}
\small
\begin{tabular}{@{}p{3.5cm}p{10cm}@{}}
\toprule
\textbf{Persona Name} & \textbf{Persona Text} \\
\midrule
\texttt{default} & You are an expert computer scientist specializing in algorithm design for combinatorial optimization problems. \\
\midrule
\multicolumn{2}{l}{\textit{Single-Dimension Personas}} \\
\texttt{simplicity\_focused} & You are an algorithm designer who prioritizes elegant, minimal solutions. \\
\texttt{complexity\_embracing} & You are an algorithm designer who builds sophisticated, feature-rich solutions. \\
\texttt{modernist} & You are an algorithm designer who uses cutting-edge techniques and recent innovations. \\
\texttt{traditionalist} & You are an algorithm designer who relies on proven, standard approaches. \\
\texttt{exotic\_explorer} & You are an algorithm designer who seeks unusual, creative solutions others might miss. \\
\texttt{efficiency\_focused} & You are an algorithm designer who optimizes for speed and resource usage. \\
\texttt{robustness\_oriented} & You are an algorithm designer who prioritizes reliability and error handling. \\
\texttt{scalability\_minded} & You are an algorithm designer who designs for large-scale applications. \\
\midrule
\multicolumn{2}{l}{\textit{Composite Personas}} \\
\texttt{modern\_simple} & You are a modernist algorithm designer who values simplicity and elegance. \\
\texttt{modern\_efficient} & You are a modernist algorithm designer who optimizes for speed and efficiency. \\
\texttt{modern\_robust} & You are a modernist algorithm designer who ensures reliability and robustness. \\
\texttt{traditional\_simple} & You are a traditionalist algorithm designer who values simplicity and proven methods. \\
\texttt{traditional\_robust} & You are a traditionalist algorithm designer who focuses on reliability and robustness. \\
\texttt{traditional\_scalable} & You are a traditionalist algorithm designer who designs for scalability using proven techniques. \\
\texttt{exotic\_efficient} & You are an exotic algorithm designer who prioritizes efficiency through creative approaches. \\
\texttt{exotic\_robust} & You are an exotic algorithm designer who ensures robustness with unusual techniques. \\
\texttt{exotic\_scalable} & You are an exotic algorithm designer who achieves scalability through innovative methods. \\
\texttt{simple\_efficient} & You are an algorithm designer who values both simplicity and computational efficiency. \\
\texttt{simple\_robust} & You are an algorithm designer who combines minimalism with strong reliability. \\
\texttt{complex\_efficient} & You are an algorithm designer who builds sophisticated solutions optimized for performance. \\
\texttt{complex\_robust} & You are an algorithm designer who creates feature-rich, highly reliable solutions. \\
\texttt{complex\_scalable} & You are an algorithm designer who develops comprehensive solutions for large-scale problems. \\
\bottomrule
\end{tabular}
\end{table}

\paragraph{Metrics}
\label{app:diversityMetrics}

We quantify diversity in the embedding space using two complementary metrics. Each algorithm is embedded using \textsc{mGTE-large-en-v1.5} with concatenated name, description, and preprocessed code, following the embedding configuration selected in Appendix~\ref{app:embedding_ablation}. For a set of $N$ algorithms with embeddings $\{\mathbf{e}_i\}_{i=1}^N$, where $\mathbf{e}_i$ denotes the embedding of algorithm $a_i$:

\textbf{Mean Nearest Neighbor Distance (MNND):} Measures local sparsity:
\begin{equation}
\text{MNND} = \frac{1}{N} \sum_{i=1}^{N} \min_{j \neq i} d(\mathbf{e}_i, \mathbf{e}_j)
\end{equation}

\textbf{Mean Pairwise Distance (MPD):} Measures global spread:
\begin{equation}
\text{MPD} = \frac{2}{N(N-1)} \sum_{i=1}^{N} \sum_{j=i+1}^{N} d(\mathbf{e}_i, \mathbf{e}_j)
\end{equation}

where $d(\mathbf{e}_i, \mathbf{e}_j) = 1 - \cos(\mathbf{e}_i, \mathbf{e}_j)$ is the cosine distance between embeddings.
MNND captures whether algorithms are well separated from their nearest neighbors, while MPD captures the overall spread of the set in embedding space.

\subsubsection{Results and Analysis}
Table~\ref{tab:init_results} presents results on both CVRP and FSS. 
Across the tested settings, \textit{+Divergence}, which is the initialization strategy used in ATLAS, consistently yields the most diverse initial algorithm sets. 
In contrast, random persona variation shows no consistent advantage over baseline stochastic sampling.

    \begin{table}[t]
    \centering
    \caption{\emph{Initialization-strategy analysis.} Results are averaged over 5 runs
of 50 algorithms each. Valid denotes the fraction of generated algorithms that pass validation. MNND
and MPD measure local and global diversity in embedding space, respectively; higher is better for
all three metrics. Best results are shown in \textbf{bold}.}
    \label{tab:init_results}
    \small
    \setlength{\tabcolsep}{3pt}
    \begin{tabular}{@{}lccccccccc@{}}
    \toprule
    & \multicolumn{3}{c}{\textbf{Baseline}} & \multicolumn{3}{c}{\textbf{+Random Persona}} & \multicolumn{3}{c}{\textbf{+Divergence (ATLAS)}} \\
    \cmidrule(lr){2-4} \cmidrule(lr){5-7} \cmidrule(lr){8-10}
    \textbf{Model} & Valid$\uparrow$ & MNND$\uparrow$ & MPD$\uparrow$ & Valid$\uparrow$ & MNND$\uparrow$ & MPD$\uparrow$ & Valid$\uparrow$ & MNND$\uparrow$ & MPD$\uparrow$ \\
    \midrule
    \multicolumn{10}{l}{\textbf{CVRP}$(n50)$} \\
    \textsc{GPT-4.1-Mini}     & 0.96 & 0.015 & 0.035 & 0.98 & 0.023 & 0.045 & 0.93 & \textbf{0.058} & \textbf{0.151} \\
    \textsc{GPT-5-mini} (low)       & 0.99 & 0.037 & 0.103 & 0.98 & 0.034 & 0.097 & 0.98 & \textbf{0.062} & \textbf{0.159} \\
    \textsc{Claude Haiku 4.5} & 0.86 & 0.029 & 0.077 & 0.89 & 0.027 & 0.068 & 0.90 & \textbf{0.064} & \textbf{0.147} \\
    \textsc{Llama 4 Maverick} & 0.85 & 0.020 & 0.095 & 0.87 & 0.019 & 0.094 & 0.84 & \textbf{0.039} & \textbf{0.123} \\
    \midrule
    \multicolumn{10}{l}{\textbf{FSS}$(n50m10)$} \\
    \textsc{GPT-4.1-Mini}     & 0.93 & 0.019 & 0.069 & 0.97 & 0.022 & 0.077 & 0.92 & \textbf{0.053} & \textbf{0.147} \\
    \textsc{GPT-5-mini} (low)      & 0.99 & 0.021 & 0.049 & 0.99 & 0.021 & 0.051 & 0.98 & \textbf{0.065} & \textbf{0.160} \\
    \textsc{Claude Haiku 4.5} & 0.96 & 0.030 & 0.078 & 0.98 & 0.030 & 0.081 & 0.93 & \textbf{0.069} & \textbf{0.158} \\
    \textsc{Llama 4 Maverick} & 0.85 & 0.017 & 0.100 & 0.93 & 0.017 & 0.106 & 0.89 & \textbf{0.033} & \textbf{0.142} \\
    \bottomrule
    \end{tabular}
    \end{table}

\begin{enumerate}[leftmargin=*,label=\textbf{(\arabic*)}]
    \item \textbf{Prompt variation provides limited and inconsistent diversity gains.} 
    The \textit{+Random Persona} strategy does not consistently improve diversity. 
    On CVRP, it reduces diversity for three of four models (\textsc{GPT-5-mini} (low): MNND -8\%, \textsc{Claude Haiku 4.5}: MNND -7\%, \textsc{Llama 4 Maverick}: MNND -5\%), while improving it only for \textsc{GPT-4.1-Mini} (MNND +53\%). 
    On FSS, the gains are similarly small, with only modest changes across models (0--16\% MNND and 1--12\% MPD). 
    In an inspected sample set, persona variation often changed surface-level style, such as
variable naming, comment verbosity, or code organization, without consistently changing the
recognized algorithmic approach. Together with the embedding metrics, this suggests that persona
variation alone is a weak mechanism for broadening the initial archive in this setting.

    \item \textbf{\textit{+Divergence} consistently increases diversity.} 
    The \textit{+Divergence} strategy outperforms the other two strategies across all models, both problems, and both diversity metrics (16 out of 16 comparisons). 
    On CVRP, relative improvements range from 68\% to 287\% for MNND and from 29\% to 331\% for MPD. 
    On FSS, they range from 94\% to 210\% for MNND and from 42\% to 227\% for MPD. 
    These results indicate that reference-based divergence is much more effective than stochastic sampling or persona variation at producing well-separated initial algorithms in embedding space. 
    Conditioning generation on multiple existing references is intended to move proposals
away from already represented regions; the reported result establishes greater embedding-space
separation.

    \item \textbf{The diversity gains come with only a modest impact on validity.} 
    The \textit{+Divergence} strategy maintains high validity rates across all models in both domains. 
    On CVRP, validity ranges from 84--98\%, representing decreases of 1--3 percentage points for most models, while \textsc{Claude Haiku 4.5} improves from 86\% to 90\%. 
    On FSS, validity ranges from 89--98\%, with small changes of up to 4 percentage points; most models show slight decreases, while \textsc{Llama 4 Maverick} improves from 85\% to 89\%.
    A slight reduction in validity is expected, since encouraging generation away from existing references can push the model toward less familiar solution patterns, which may increase the chance of invalid outputs. 
    In the ATLAS setting, this trade-off is acceptable because initialization is followed by downstream validation and repair, allowing the method to benefit from broader early coverage while keeping failure rates manageable.
 
\end{enumerate}

\textbf{Implication for initialization design:}
Across both CVRP and FSS, and across all four tested LLMs (\textsc{GPT-4.1-Mini}, \textsc{GPT-5-mini} (low), \textsc{Claude Haiku 4.5}, and \textsc{Llama 4 Maverick}), \textit{+Divergence} consistently produces more diverse initial algorithm sets than either baseline stochastic sampling or persona variation. 
These results support ATLAS's use of reference-based divergence in its dual-phase initialization.

%% file: Appendix/ablation_clustering.tex
\subsubsection{Aim and Research Question}

\paragraph{Aim}
ATLAS organizes the evolving archive using semantic embeddings, grouping similar
candidates into operational regions for retrieval and representative selection.
These clusters guide the three-layer search strategy by defining
\begin{inparaenum}
\item a local elite neighborhood for intensive refinement (Layer~1), and
\item a set of diverse cluster representatives for distributed refinement and exploratory synthesis (Layers~2 and~3).
\end{inparaenum}
This search-guidance ablation replaces both embedding-based neighbor selection and archive
grouping with two alternative guidance strategies that do not rely on semantic embeddings, while
retaining the remaining ATLAS machinery.

\paragraph{Research question} Does using semantic embeddings for Layer~1 neighbor selection
and archive grouping improve search performance relative to fitness-based or random guidance?

\subsubsection{Setup}
We evaluate on two representative combinatorial optimization problems: CVRP (50 customers, 31 instances) and FSS (50 jobs, 10 machines, 31 instances).
For each problem, all variants use the same training and test instance sets and follow the evaluation protocol described in \S\ref{subsec:ExperimentalSetup}. 
Benchmark instances are generated using the default settings in Appendix~\ref{app:instance_generation}. 
Each variant is executed for $R=5$ independent synthesis runs with different random seeds,
under the common budget used in the end-to-end ATLAS search studies: 500 evaluated operator
executions per run.

\paragraph{Controlled Adaptive Number of Clusters}
To ensure a fair comparison and avoid confounding effects from different cluster counts, all variants use the same archive-driven adaptive rule for determining the number of clusters:
\begin{equation}
K_t = \min\Big(\max\big(\big\lfloor |\mathcal{A}_t|/d \big\rfloor,\; K_{\min}\big),\; K_{\max}\Big),
\label{eq:adaptive_k_ablation}
\end{equation}
where $|\mathcal{A}_t|$ is the archive size at iteration $t$, $d$ is a divisor targeting approximately $d$ algorithms per cluster (default $d{=}10$), $K_{\min}{=}3$ ensures that at least two clusters remain for Layers~2 and~3 after removing the cluster containing the best algorithm $a^*$, and $K_{\max}{=}10$ follows the square-root heuristic ($\sqrt{|\mathcal{A}_{\max}|}=10$) to limit cluster granularity. 
Thus, all variants use \emph{the same number of clusters}; they differ only in \emph{how algorithms are assigned to those clusters}.

%------------------------------------------------------------------------------
\paragraph{Unified Architecture with Guidance-Specific Neighbor Selection and Grouping}
To isolate the tested search-guidance strategy, all variants follow the same ATLAS
architecture but differ in how Layer~1 neighbors are selected and how the archive is partitioned
into groups.

At each iteration:
\begin{enumerate}[itemsep=2pt]
\item \textbf{Form the elite set (Layer 1):} Select the best algorithm $a^*$ and its four nearest neighbors according to the variant's similarity metric (five algorithms total).

\item \textbf{Partition the archive:} Partition the archive $\mathcal{A}_t$ into $K_t$ clusters using the variant's grouping method.

\item \textbf{Exclude the best cluster:} Remove the cluster containing $a^*$ to avoid redundant exploration around the current best region.

\item \textbf{Form Layers~2 and~3:} Select the best-performing representative from each remaining cluster.
\end{enumerate}

%------------------------------------------------------------------------------
\textbf{Search-Guidance Strategies:}
%------------------------------------------------------------------------------
\begin{itemize}[itemsep=1pt]

\item \textbf{Variant A: Semantic guidance}
\begin{itemize}[itemsep=1pt]
\item \textit{Elite selection:} k-nearest neighbors in embedding space (cosine similarity over combined text and code embeddings)
\item \textit{Clustering:} K-Medoids applied in embedding space
\end{itemize}

\item \textbf{Variant B: Fitness-based guidance}
\begin{itemize}[itemsep=1pt]
\item \textit{Elite selection:} $a^*$ plus four algorithms with the closest fitness values
\item \textit{Clustering:} Sort algorithms by fitness and partition them into $K_t$ contiguous bands
\end{itemize}

\item \textbf{Variant C: Random guidance}
\begin{itemize}[itemsep=1pt]
\item \textit{Elite selection:} $a^*$ plus four uniformly random algorithms
\item \textit{Clustering:} Randomly shuffle the archive and partition it into $K_t$ equal-sized groups
\end{itemize}

\end{itemize}

All variants share the same archive management mechanisms (embedding-based deduplication and pruning), the same elite set size (five algorithms), the same adaptive cluster count ($K_t$ from Eq.~\eqref{eq:adaptive_k_ablation}), and the same representative selection rule (the best-performing algorithm in each cluster). 
They differ only in the rules used for Layer~1 neighbor selection and archive grouping,
which together guide search across the three layers.

\subsubsection{Results and Analysis}
\label{app:ablation_clustering_results}

Table~\ref{tab:ablation_clustering_gap} reports relative mean test gaps for the algorithm
selected by training performance in each run of each variant.

\begin{table}[htbp]
\centering
\caption{\emph{Relative mean gaps for the semantic search-guidance ablation.} Percent
gap to the strongest human-designed reference in each setting (PyVRP for CVRP and IG-TB for FSS);
lower is better. Values are computed from the underlying raw objective means. Boldface mirrors the
paired raw-objective analysis and is not based on a test of the displayed gaps.}
\label{tab:ablation_clustering_gap}
\small
{
\begin{tabular}{@{}lcc@{}}
\toprule
\textbf{Guidance Strategy} & \textbf{CVRP}$(n50)$ gap (\%)$\downarrow$ & \textbf{FSS}$(n50m10)$ gap (\%)$\downarrow$ \\
\midrule
Variant A: Semantic Guidance      & \textbf{0.570} & \textbf{0.391} \\
Variant B: Fitness-Based Guidance   & 2.601 & 0.715 \\
Variant C: Random Guidance          & 3.143 & 0.884 \\
\bottomrule
\end{tabular}}
\end{table}

\paragraph{Analysis}
Semantic search guidance (Variant A), which uses embeddings for both Layer~1 neighbor selection and archive grouping, 
achieves the best performance on both benchmarks. This combined treatment outperforms the corresponding fitness-based 
and random guidance strategies. It therefore shows that the selected semantic representation provides more effective search 
guidance than the tested alternatives. 
Fitness-based guidance still introduces useful structure by relating algorithms with similar performance and outperforms 
random guidance on both benchmarks. However, algorithms with similar fitness can still differ substantially in their underlying search logic.
Semantic guidance instead relates algorithms using representation-level similarity derived from their descriptions and code.

The combined semantic guidance can benefit the three-layer strategy in several ways. In Layer~1, embedding-based nearest neighbors 
can make local refinement around the current best algorithm more targeted. In Layer~2, each representative is refined with algorithms
 from its own embedding-based group, providing a more coherent local context for representative-level refinement. In Layer~3, semantic 
 grouping supplies representatives separated under the same representation used by the archive. The observed final-quality gain is 
 consistent with this context improving cross-group synthesis.

Table~\ref{tab:ablation_clustering_gap} reports only the performance of the selected best
algorithms. It therefore supports a conclusion about final search performance, but does not measure
repertoire diversity, mechanistic family alignment, or deployment complementarity.

At the same time, the fitness-based and random variants remain reasonably strong because they still retain much of the ATLAS architecture. 
In all variants, Layer~1 continues to refine the current best algorithm, Layer~2 still distributes refinement through group representatives, and Layer~3 still performs cross-group \textsc{Combine} and \textsc{Diverge} operations. 
Moreover, all variants continue to benefit from the same semantically managed archive (embedding-based deduplication and pruning) and diverse initialization. 
Thus, the gains from semantic search guidance should be interpreted as the joint added
value of semantically informed neighbor selection and grouping within the full ATLAS system, rather
than as the sole source of search effectiveness.

Overall, these results support the narrower conclusion that using the selected semantic
representation for neighbor selection and archive grouping improves final search guidance relative
to the tested fitness-based and random strategies.

%% file: Appendix/ablation_layers.tex
\subsubsection{Aim and Research Question}

\paragraph{Aim}
ATLAS uses a three-layer search architecture to balance local refinement, distributed search across multiple archive regions, and cross-cluster synthesis. 
Layer~1 performs intensive refinement around the current best algorithm and its nearest semantic neighbors. 
Layer~2 performs distributed refinement over representatives from multiple embedding
clusters, enabling search across separated archive regions.
Layer~3 performs exploratory synthesis between cluster representatives to generate algorithms that may combine ideas from different regions or move toward previously underexplored ones. 
This ablation compares selected enabled and disabled layer combinations while keeping the
remaining ATLAS components fixed.

\paragraph{Research question}
How does ATLAS performance change across the selected layer configurations, and what is
lost when local or representative-level refinement is removed?

\subsubsection{Setup}
We evaluate on the same two representative problem classes used in the other component
ablations and design analyses: CVRP (50 customers, 31 instances) and FSS (50 jobs, 10 machines, 31
instances).
For each problem, all variants use the same training and test instance sets and follow the evaluation protocol described in \S\ref{subsec:ExperimentalSetup}. 
Each variant is executed for $R=5$ independent synthesis runs with different random seeds,
under the same computational budget of 500 evaluated operator executions per run.
All configurations use the default ATLAS components unless explicitly ablated.

%------------------------------------------------------------------------------
\paragraph{Layer Configurations:}
%------------------------------------------------------------------------------
\begin{itemize}[itemsep=1pt]

\item \textbf{Variant A: All layers active (default)}
\begin{itemize}[itemsep=1pt]
\item \textit{Layer~1:} local intensive refinement around the current best algorithm
\item \textit{Layer~2:} distributed refinement over representatives from multiple clusters
\item \textit{Layer~3:} cross-cluster synthesis and divergence
\end{itemize}

\item \textbf{Variant B: Layer~1 only}
\begin{itemize}[itemsep=1pt]
\item Uses only local intensive refinement around the current best algorithm and its local neighborhood
\item Disables distributed refinement over cluster representatives and cross-cluster synthesis
\end{itemize}

\item \textbf{Variant C: Layers~2+3 only}
\begin{itemize}[itemsep=1pt]
\item Disables local intensive refinement around the current best algorithm
\item Retains distributed refinement over cluster representatives and cross-cluster synthesis
\end{itemize}

\item \textbf{Variant D: Layer~3 only}
\begin{itemize}[itemsep=1pt]
\item Disables both local refinement and representative-level refinement
\item Uses only cross-cluster synthesis and divergence between cluster representatives
\item Emphasizes structured exploration across distinct archive regions, while still grounding synthesis in high-performing representatives
\end{itemize}

\end{itemize}

This ablation tests whether the complete configuration improves over the selected reduced configurations.

%==============================================================================
\subsubsection{Analysis}
\label{app:layer_ablation_results}
%==============================================================================

Table~\ref{tab:layer-ablation-gap} reports the test-set relative mean gap for each ablation configuration.
The full three-layer architecture (Variant~A) achieves the best performance on both
benchmarks among the four configurations, which is consistent with the layers serving complementary
roles.
The second-best results are obtained by Variant~C (Layers~2+3 only), followed by Variant~D (Layer~3 only), while Variant~B (Layer~1 only) performs worst.

The comparison between Variant~A and Variant~C shows that Layer~1 is important: removing intensive local refinement around the current best algorithm leads to a performance drop on both benchmarks. 
This indicates that allocating additional search effort to the neighborhood of the best discovered algorithm improves final solution quality.

However, the lower performance of Variant~B shows that Layer~1 alone is not sufficient. 
Although it benefits from a strong warm start through the dual-phase initialization, which already provides a diverse initial archive (Appendix~\ref{app:init_ablation}), once search is restricted to Layer~1 it can only continue refining around the current best algorithm and its local neighborhood. 
Without Layers~2 and~3, the configuration loses explicit representative-level and
cross-cluster operations. Layer~1 can still modify complete source and may generate internally
hybrid code, but it no longer receives the cross-cluster reference combinations supplied by Layer~3.
The performance result therefore supports broader archive-level search, not a claim that Layer~1 is
incapable of producing any qualitatively new mechanism.

The strong performance of Variant~C shows that ATLAS derives benefit from distributed search over cluster representatives together with cross-cluster synthesis. 
Even without the intensive local refinement of Layer~1, these two layers can still maintain competitive performance by refining multiple promising regions of the archive and generating new candidates through interactions across them. 
This supports the ATLAS design intuition that full-algorithm search benefits from
allocating effort beyond the current-best neighborhood to representatives in multiple
embedding-space regions.

Variant~D (Layer~3 only) provides a direct comparison with Variant~C for estimating the
contribution of Layer~2 when Layer~3 remains active. It is weaker than Variant~C but outperforms
Variant~B (Layer~1 only). The latter comparison shows that the Layer~3-only configuration is
stronger than the Layer~1-only configuration in the tested settings.
At the same time, Layer~3 should not be interpreted as purely exploratory. 
Because its synthesis operators use high-quality cluster representatives, often including the current best algorithm as a reference, Layer~3 also retains an exploitative component. 
It therefore acts as a structured exploration layer that recombines strong representatives
from separated clusters rather than sampling without reference context.

Overall, the results are consistent with a division of labor across layers.
Layer~1 strengthens local refinement around the current best algorithm, Layer~2 supports
distributed refinement across multiple promising archive regions, and Layer~3 enables structured
exploration through recombination and divergence across cluster representatives.
The best results are obtained when all three layers are active, supporting the complete
configuration over the selected reduced alternatives.

%% file: Appendix/ablation_reasoning.tex
\subsubsection{Aim and Research Question}

\paragraph{Aim}
Modern reasoning-capable LLMs such as \textsc{GPT-5-mini} expose a reasoning-effort parameter that controls how much internal reasoning is allocated before generating a response. 
Higher reasoning effort may improve the quality of discovered algorithms by enabling more careful modifications, but it also increases token usage, API cost, and latency per operator call. 
This sensitivity analysis studies whether increasing reasoning effort yields enough
performance benefit to justify its additional computational cost in ATLAS.

\paragraph{Research question}
How does reasoning effort affect the quality of discovered algorithms and the total token cost of search, and which reasoning level provides the best quality--cost trade-off?

\subsubsection{Setup}
We evaluate on the same two representative problem classes used in the component ablations
and other design analyses: CVRP (50 customers, 31 instances) and FSS (50 jobs, 10 machines, 31
instances).
For each problem, all variants use the same training and test instance sets and follow the evaluation protocol described in \S\ref{subsec:ExperimentalSetup}. 
Each variant is executed for $R=5$ independent synthesis runs with different random seeds,
under the same computational budget of 500 evaluated operator executions per run.

All configurations use the default ATLAS architecture and \textsc{GPT-5-mini} as the underlying synthesis model. 
They differ only in the assigned reasoning-effort level: minimal, low (default), medium, and high. 
This controlled sensitivity study examines the effect of reasoning effort on both
synthesis quality and computational cost.

%==============================================================================
\subsubsection{Results and Analysis}
\label{app:reasoning_ablation_results}
%==============================================================================

Table~\ref{tab:reasoning_ablation_gap} reports the relative mean test gaps and average total token usage per run for each reasoning level.

\begin{table}[htbp]
\centering
\caption{\emph{Reasoning-effort sensitivity and cost analysis for \textsc{GPT-5-mini}.}
Test performance is reported as the relative mean gap (\%) to the strongest human-designed reference
in each setting (PyVRP for CVRP and IG-TB for FSS); lower is better. Gaps are computed from the
underlying raw objective means. Token usage is the mean total tokens consumed per run, in thousands,
with standard errors in parentheses; each run uses 500 evaluated operator executions. Boldface
mirrors the paired raw-objective analysis and is not based on a test of the displayed gaps.}
\label{tab:reasoning_ablation_gap}
\small
{
\begin{tabular}{@{}lcccc@{}}
\toprule
\multirow{2}{*}{\textbf{Reasoning effort}} & \multicolumn{2}{c}{\textbf{Relative mean gap (\%)}$\downarrow$} & \multicolumn{2}{c}{\textbf{Total tokens per run (k)}} \\
\cmidrule(lr){2-3} \cmidrule(lr){4-5}
& \textbf{CVRP}$(n50)$ & \textbf{FSS}$(n50m10)$ & \textbf{CVRP}$(n50)$ & \textbf{FSS}$(n50m10)$ \\
\midrule
Minimal       & 1.815 & 0.597 & 6\,146 (135)  & 4\,258 (178) \\
Low (default) & \textbf{0.094} & 0.347 & 6\,335 (144)  & 4\,491 (246) \\
Medium        & \textbf{0.092} & \textbf{0.301} & 8\,126 (161)  & 5\,865 (272) \\
High          & \textbf{0.000} & \textbf{0.292} & 10\,928 (241) & 8\,247 (294) \\
\bottomrule
\end{tabular}}
\end{table}

\paragraph{Analysis}
Increasing reasoning effort from Minimal to Low improves performance on both benchmarks, showing that a modest increase in reasoning effort is beneficial for generating stronger algorithmic modifications. 
Importantly, this improvement comes at only a small increase in token usage: compared with Minimal, Low increases total tokens by approximately 3\% on CVRP and 5\% on FSS. 
Thus, Low provides a favorable improvement in solution quality at a relatively minor additional cost. 
At the same time, Minimal reasoning remains reasonably competitive, but is consistently weaker than the higher-effort settings.

Further increases beyond Low yield only limited additional gains. 
On CVRP, Low, Medium, and High are not statistically significantly different, indicating that additional reasoning budget does not translate into a reliable performance improvement. 
On FSS, Medium and High form the statistically best group and are not significantly different from each other, while Low falls outside this group; the associated improvement is nevertheless small in absolute terms, reducing the mean gap by less than $0.06$ percentage points.
The corresponding increase in token consumption, however, is substantial. 
Compared with Low, Medium increases token usage by approximately 28\% on CVRP and 31\% on FSS, while High increases it by approximately 73\% on CVRP and 84\% on FSS.
These large increases in cost yield only marginal gains in solution quality.

Overall, these results suggest that, for \textsc{GPT-5-mini}, Low reasoning provides the best practical operating point for ATLAS. 
It improves consistently over Minimal at only a modest additional cost, while Medium and High offer only diminishing returns relative to their substantially higher token usage, API cost, and latency. 
We therefore adopt Low reasoning as the default setting.